\documentclass[]{jair}
\let\oldmaketitle\maketitle
\renewcommand{\maketitle}{%
  \oldmaketitle%
  \thispagestyle{fancy}
  \fancyhead[LO,RE]{\textit{As published in JAIR, \url{https://doi.org/10.1613/jair.1.17167}}}}

\setcopyright{cc}
\copyrightyear{2026}
\acmDOI{10.1613/jair.1.17167}

\JAIRAE{Patrik Haslum}
\JAIRTrack{} \acmVolume{85}
\acmArticle{21}
\acmMonth{02}
\acmYear{2026}

\usepackage[utf8]{inputenc}

\RequirePackage[
  datamodel=acmdatamodel,
  style=acmauthoryear,
  backend=biber,
  giveninits=true,
  uniquename=init
  ]{biblatex}

\usepackage{placeins}

\usepackage{soul}
\usepackage{graphicx}
\usepackage{booktabs}
\usepackage{algorithm}
\usepackage{xspace}
\usepackage{siunitx}

\usepackage[noend]{algpseudocode}

\usepackage{tikz}
\usepackage[capitalise]{cleveref}
\usepackage{multirow}

\usepackage{alltt} 
\usepackage{dcolumn} 
\usepackage{longtable}
\usepackage{rotating} 

\usepackage{diagbox} 

\usepackage{listings}

\usepackage{enumitem} 

\usepackage{lineno} 

\usepackage{afterpage} 

 \usepackage{etoolbox}

\newtheorem{definition}{Definition}
\newtheorem{theorem}{Theorem}
\newtheorem{thm}[theorem]{Theorem}

\newtheorem{corollary}[theorem]{Corollary}

\newcommand{\rb}[0]{{r}{b}}
\newcommand{\anysuit}[0]{*}
\newcommand{\samesuit}[0]{$=$}
\def\spidertype{S\;}
\def\randomcard{$\P$\;}
\def\beziquenote{$\ddag$\;}
\def\fannote{$\|$}
\def\canfieldspaces{$\dagger$\;}
\def\fortunesspaces{$\dagger\dagger$\;}
\def\originalnote{$**$\;}
\def\golfnote{$\S$\;}
\def\wholepilemoves{$+$\;}
\def\filledtriangle{$\scalebox{1.33}{$\blacktriangle$}$}

\newcommand{\patienceword}[1]{\textbf{`#1'}}
\newcommand{\gamename}[1]{\textit{#1}}  

\def\thoughtful{\mbox{$[Th.]$}}

\newcommand{\submission}[1]{}

\newcommand*{\dittostraight}{-- '' --} 

\newcommand{\citeS}[1]{\citeauthor{#1}'s}

\newcommand{\tableCite}[1]{}
\newcommand{\tableCiteA}[1]{\citet{#1}}
\newcommand{\tablecite}[2]{\tableCiteA{#2}}
\newcommand{\tableNoteA}[1]{\\ &\multicolumn{3}{r|}{\textit{Note: #1}}}

\newcommand{\tableURL}[1]{\href{https://web.archive.org/web/#1}{#1}}
\renewcommand{\tableURL}[1]{\\ &&\multicolumn{5}{r|}{\href{https://web.archive.org/web/#1}{#1}}}
\renewcommand{\tableURL}[1]{}

\newcommand{\tableURLB}[2]{\\ &&\multicolumn{5}{r|}{\href{https://web.archive.org/web/#1}{#2}}}
\renewcommand{\tableURLB}[2]{}

\newcommand{\tableURLD}[2]{\href{https://web.archive.org/web/#1}{#2}}
\newcommand{\tableURLC}[1]{\tableURLD{#1}{#1}}

\newcommand{\tableGame}[1]{\multicolumn{2}{|l}{#1}}

\newcommand{\sota}[2]{$\mathbf{#1 \pm #2\%}$}
\newcommand{\notsota}[2]{$#1 \pm #2\%$}
\newcommand{\queryable}[2]{$\mathit{#1 \pm #2\%}$}

\newcommand{\done}[1]{}

\apptocmd{\sloppy}{\hbadness 10000\relax}{}{}

\newcommand{\compliant}[0]{compliant\xspace}
\newcommand{\noncompliant}[0]{non-\compliant}

\title[Winnability of Solitaire and Patience Games]{The Winnability of Klondike Solitaire and Many Other Patience Games}

\author[Blake]{Charlie Blake}
\orcid{0009-0006-3374-7241}
\email{thecharlieblake@gmail.com}
\affiliation{\institution{Work undertaken while at University of St Andrews}
  \city{St Andrews}
  \country{United Kingdom}
}

\author[Gent]{Ian Gent}
\authornote{Corresponding Author.}
\orcid{0000-0002-5604-7006}
\email{ian.gent@st-andrews.ac.uk}
\affiliation{\institution{University of St Andrews}
  \city{St Andrews}
  \country{United Kingdom}
}

\begin{document}

\begin{abstract}

Our ignorance of 
the winnability percentage of the solitaire card game `Klondike' has been described as ``one of the embarrassments of applied mathematics''. 
Klondike, the game in the Windows Solitaire program,  is just one of many single-player card games, generically called `patience' or `solitaire' games,  for which players have long wanted to know how likely a particular game is to be winnable. 
A number of different games have been studied empirically in the academic literature and by non-academic enthusiasts. Here we show that a single general purpose 
 Artificial Intelligence   program named `Solvitaire' can be used to determine the winnability percentage of 73 variants of 35 different single-player card games with a 95\% confidence interval of $\pm$ 0.1\% or better. 
For example, we report the winnability of  Klondike as 81.945\% $\pm$ 0.084\% (in the `thoughtful' variant where the player knows the rank and suit of all cards), a 30-fold reduction in confidence interval over the best previous result.  The vast majority of our results are either entirely new or represent significant improvements on previous knowledge.
Solvitaire uses depth-first search and exploits a number of AI techniques including transposition tables, symmetry breaking, dominances, and streamliners.  We give the first correctness proofs of two key dominances for patience games. 

\end{abstract}

\maketitle
\renewcommand{\shortauthors}{Blake \& Gent}

\section{Introduction}

Patience games - single-player card games also known as `solitaire' games\footnote{Herein we use the word `patience' as the traditional word in UK English while `solitaire' is the US usage  \cite{patience-history}.
}  - have been a popular pastime for more than 200 years \cite{patience-history}.  This popularity continues, with Microsoft Windows Solitaire -- just one implementation of one patience game -- being played 100 million times per day in 2020 \cite{solitaire-100million}.
 We compute 
  winnability percentages on random instances of 
many single-deck patience games using a general solver named `Solvitaire'.  Almost all our results are either entirely new or significant improvements on previous knowledge.  Where results were previously known, they were obtained using solvers specific to a particular game or small family of games.  
In contrast, Solvitaire solves a wide variety of  patience games expressible in our flexible rule-description language. 
Based on depth-first backtracking search, it exploits a number of techniques to improve efficiency:
transposition tables \cite{Greenblatt_transposition,smith-caching},  symmetry \cite{gent2006symmetry}, dominances \cite{Chu2015}, and streamliners \cite{streamliner-gomes,wetter2015automatically}.

\gamename{Klondike},\footnote{In the main text of this paper, we distinguish names of games by writing them in italics, e.g. \gamename{Klondike}.} the game in Windows Solitaire, is just one example of hundreds of patience games that exist \cite{parlett}.
Understanding the range of games available requires understanding some key terminology: 
we give a very concise introduction in Section \ref{terminology}.
The probability of winning has always been of interest to players, with advice published as to how likely a given game is to be winnable at least as long ago as 1890 \cite{cavendish}. In this paper, we study 81 variants of about 40 different patience games.
Not knowing the winnability  of just one of these games, \gamename{Klondike}, has been called ``one of the embarrassments of applied mathematics'' \cite{rollout2}.
Only for a very small number of games, e.g. \gamename{FreeCell} \cite{fish_fc_billiard}, has this probability previously been known to a high degree of accuracy.
For games with hidden cards, we follow standard practice in the literature of  considering the `thoughtful' variant \cite{rollout2}, in which the ranks and suits of hidden cards are known to the player at the start of the game. 

We are now able to report the winnability percentage of thoughtful \gamename{Klondike} and  dozens of other games with a 95\% confidence interval  within $\pm 0.1\%$. Remarkably, we achieve this with a solver which can be used for a very wide variety of games and is not highly optimised for any particular one.
Our rule-sets and solver are flexible enough to include famous games such as \gamename{Klondike}, \gamename{Canfield}, \gamename{FreeCell}, \gamename{Spider}, \gamename{Golf}, \gamename{Accordion}, \gamename{Black Hole} and \gamename{King Albert}, all of which are very different from each other. We are not aware of any previous solver which can be used unchanged on any two of these games.

\section{Terminology of Patience Games}
\label{terminology}
\label{description-patience} 

\begin{figure}
            \centering
            \includegraphics[width=0.49\linewidth]{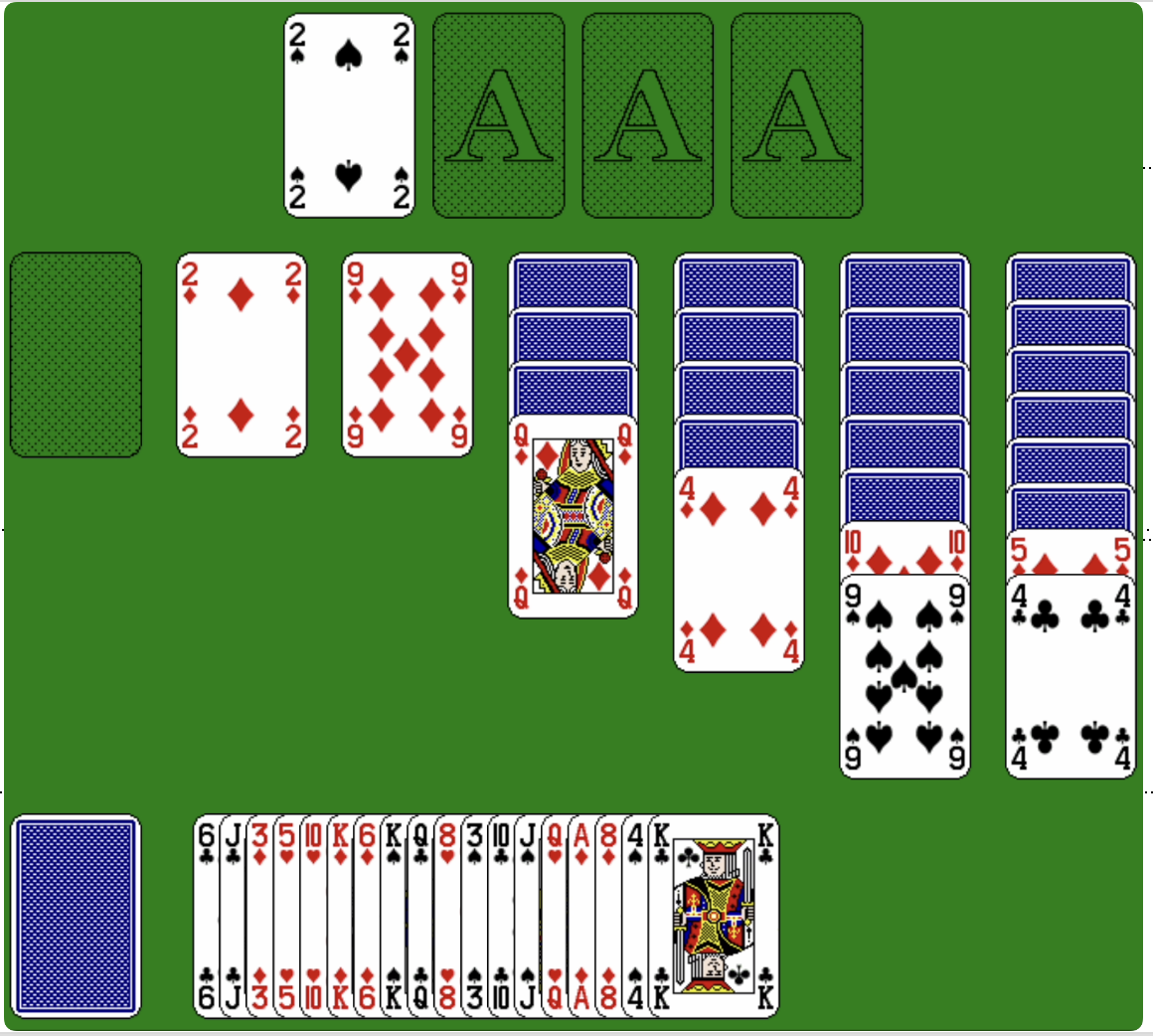}             
            ~\includegraphics[width=0.49\linewidth]
            {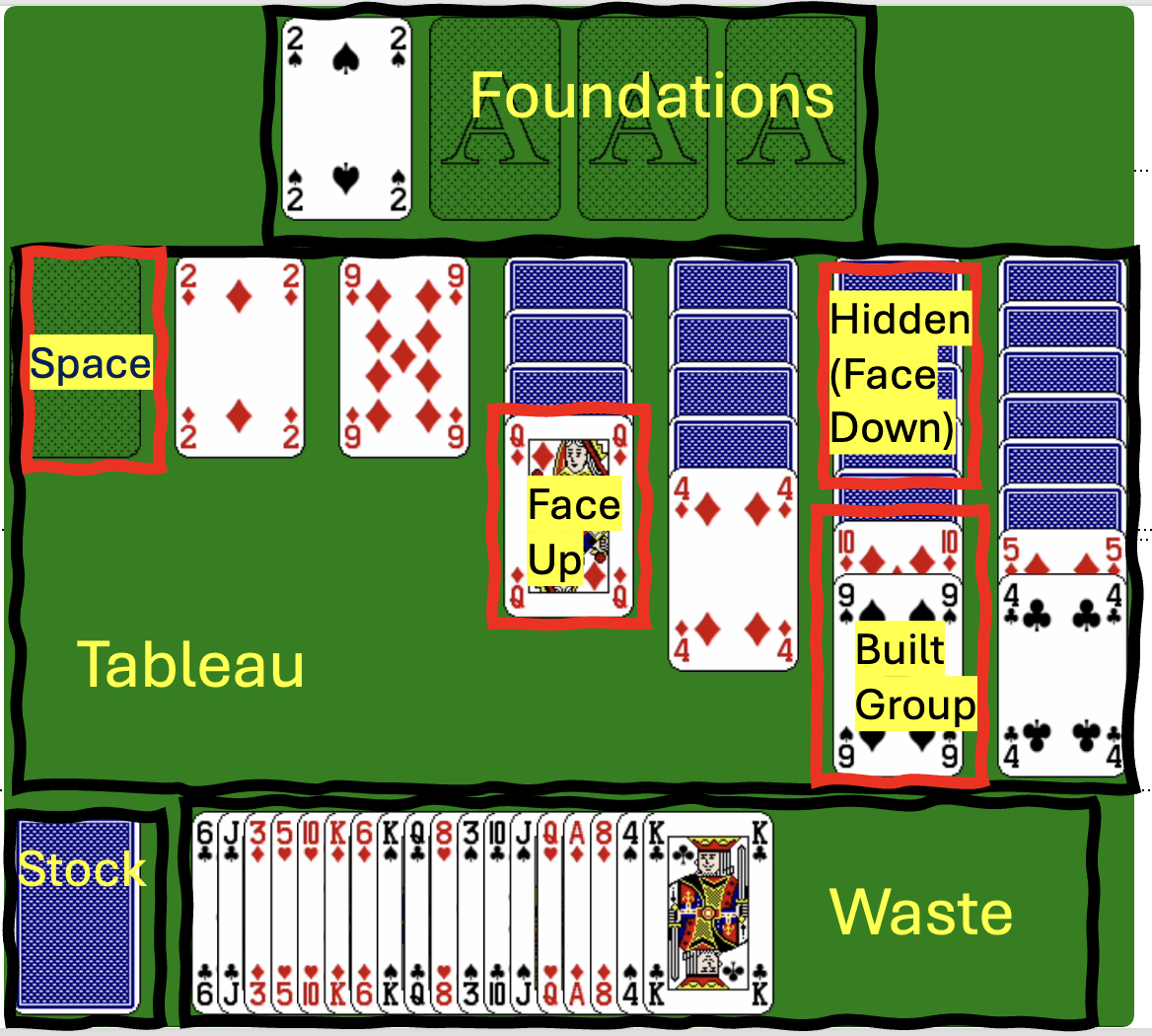}   
     \caption{(Left) Sample layout of the game of \gamename{Klondike} part way through play. (Right) The same layout illustrating some terminology from \Cref{terminology} with general areas of the layout outlined in black, and specific features outlined in red.   }
        \label{fig:layout}
           \Description[Illustration of Klondike with some annotations indicating terminology]{In this illustration of Klondike the following terms are shown in relevant areas of the image: Foundations, Tableau, Waste Stock. Additionally, the following specific features of the layout are described: Space, hidden (also called face down), built group. } 
    \end{figure}

 Giving a general introduction to single-player card games is outwith the scope of this paper.
 Excellent introductions to patience games can easily be found in books \cite{parlett} or online.  A sample layout of \gamename{Klondike} in play is shown in  \Cref{fig:layout} together with an illustration of some relevant terminology. Because terminology of patience games is not always the same in different sources, we briefly define key terms we use in this paper.
 
We use the word \patienceword{game} to refer to a particular set of rules for playing patience. 
A game is played with a number of complete \patienceword{decks} of cards, normally the standard deck with 13 cards of each of 4 suits. The rules of a game specify how the cards are placed before play starts: in the initial position some cards may be \patienceword{hidden} from the player, for example by being placed \patienceword{face-down}.
In this paper we follow previous work in studying \patienceword{thoughtful} variants of a game where the ranks and suits of hidden cards are known. With physical cards, the thoughtful variation is like the player peeking at each hidden card to see what it is.  Electronic implementations with unlimited undos also become thoughtful, because the player can always go back to the start after finding any information they need in the game.
We use the word \patienceword{instance} of a game to refer a particular arrangement of cards for that game, usually after random shuffling.  Most games are won by rearranging cards so as to place them in order on a set of \patienceword{foundations}, typically from A to K within each suit: in some cases the player is given some cards already placed on the foundation as a starter. In some games the goal instead is to move cards into a single \patienceword{hole}, with consecutive cards required to be adjacent in rank but with no regard to suit: games vary whether one is allowed to loop round from K to A and vice versa.  In some games, like \gamename{Spider}, cards are not built to foundations individually but all simultaneously when a complete sequence from A to K in a single suit has been constructed.
An instance of a game is \patienceword{winnable} if there is any legal sequence of moves that leads to the goal predetermined by the rules of the game.
In most games the main area of play is called the \patienceword{tableau}. Cards can often be moved within the tableau: this is called \patienceword{building} one card onto another pile. In the scope of this paper, the card must be one lower in rank than the card it is placed on (with K considered one lower than A if appropriate).  The \patienceword{build policy} determines additional rules: to be built on a card may need to be the same suit as the higher card, or of a suit of the opposite colour, or it may be allowed to be any suit. A sequence of consecutive built cards may be allowed to be moved together as one \patienceword{group}: where allowed this may be with same restriction as the build policy, or sometimes a stricter restriction that the group must all be the same suit. We refer to the number of tableau piles and their sizes at the start of the game as the \patienceword{layout}. 
Typically a face-down card in the tableau is turned \patienceword{face-up} only when the card immediately covering it is moved. 
When a tableau pile becomes empty it is called a \patienceword{space}: some games allow cards to be placed in spaces; sometimes the card placed in the space must be a K and in other games any card is allowed. In some games cards may be \patienceword{worried back}: this means cards can be moved from a foundation to the tableau. 
Some games contain an ordered \patienceword{stock} of cards: often the player  is allowed to \patienceword{draw} a given number of cards at a time. Most often stock cards are moved to a \patienceword{waste} pile, from which the top card can then be played to the tableau, while in others one card is dealt onto each tableau pile. Some games allow \patienceword{redeals}, where the waste pile may be reused to form the stock again.
Some games have a \patienceword{reserve} of cards which can be played onto the tableau or foundations but otherwise are static. \patienceword{Free cells} function like a reserve but cards may be moved from the tableau into free cells as well as in the other direction.

 As an example we can now describe  \gamename{Klondike} as illustrated in \Cref{fig:layout}.  A single standard deck is used and  the goal is to build all cards on foundations in suit from A to K. The game begins with a tableau of 28 cards in a triangular form with piles from 1 to 7 cards, with all but the top card face-down. 
 Face-up cards on the tableau may be built in alternating colour, and built groups may be moved. Face-down cards may not be moved.\footnote{This prohibition means that thoughtful \gamename{Klondike} is subtly different from a variant in which all cards start face-up.}
 Spaces may be filled only by a K. Cards may be worried back from foundations to tableau.  A stock of 24 cards may be drawn in groups of three, and redeals are allowed without limit. This description may be compared with Table~\ref{rules-table}, Appendix \ref{app:rules}. These rules are given to Solvitaire in a JSON format shown in Listing~\ref{rules-klondike}, page~\pageref{rules-klondike}: 
for more details of our rules language, see \Cref{sec:configurable-rule-sets} and Appendix \ref{app:rule-description-language}.

Names of particular patience games are even more confused than the terminology for rules, with different names used for the same game and the same name used for different games. For example, the game we call \gamename{Klondike} in this paper is often just called `Patience' \cite{parlett}  or `Solitaire' which are also names for the general family of single-player card games. Worse than that, \gamename{Klondike} can also be called `Canfield' which is the name we use here for a completely different game. Both games have many other names, for example both being sometimes called `Demon'.\footnote{`Demon' was the name used by Ian Gent's mother for \gamename{Canfield} and by his father  for \gamename{Klondike}.}   Unfortunately this means we sometimes do not know what game is being referred to in historical documents: for example Stanislaw Ulam may have been referring to either game when he wrote that `Canfield Solitaire' motivated his invention of Monte Carlo methods \cite{ulam-solitaire}.
    It is therefore particularly important for us to be clear on the name and rules we use for each game.  
We provide a concise summary of rules we used of most games studied in this paper in \Cref{rules-table}, page~\pageref{rules-table}.  Almost all games we studied can be described in this way, including all  games for which we give the first reported results. 
 The exceptional games that cannot be described in this framework are Accordion \cite{accordion-rules}, its variant with 18 cards (called `Late-Binding Solitaire' by its originator) \cite{accordion-stanford}, and the two variants of Gaps \cite{helmstetter2004searching}: their rules can be found in the papers just cited and are shown in our JSON format in Listings~\ref{rules-accordion} and~\ref{rules-gaps}, page~\pageref{rules-accordion}.
 
\section{History of Solving Patience Games}

\label{history}

The winnability of patience games has interested people for many years, with many books on the topic providing estimates of how often each game can be won.  In some cases, an expert's views were astonishingly accurate: in the nineteenth century \citet{cavendish} said that the game \gamename{Fan} ``with careful play, is slightly against the player'', while we show that it is $48.776\% \pm 0.099\%$ winnable.\footnote{We studied a very minor variant of Cavendish's game, with sixteen piles of three and two piles of two instead of seventeen piles of three and one of one.}  Other stated claims have been very inaccurate:  (British) \gamename{Canister} was described by \citet{parlett} as ``odds in favour'', while Table~\ref{tab:solpercs-new} shows that only slightly more than one in a million games are winnable.  
Distinguished scientists who have taken an interest in the question include Stanislaw Ulam, the inventor of computer-based Monte Carlo Methods,  \cite{ulam-solitaire}, 
Donald Knuth, a Turing Award winner \cite{accordion-stanford}, and Irving Kaplansky, a President of the American Mathematical Society \cite{graham-mackenzie}.   

 There are some patience games where there is no player choice required and the pleasure of the game is the purely mechanical playing out of the game.  We do not pay attention to such games in this paper, but some winnability percentages have been calculated. For example, \gamename{Clock Patience} is provably won exactly $\frac{1}{13}$ of the time \cite{doi:10.1080/0025570X.1981.11976927}. Monte Carlo methods have shown \gamename{Perpetual Motion} to have a winnability of  $8.6692 \pm 0.0017\%$ 
 while superficially minor changes to the rules can increase this 
 to $54.8033 \pm  0.0031\%$ winnability \cite{masten_perpetual,DBLP:journals/corr/abs-0907-1955}.

When Microsoft released one of the early versions of \gamename{FreeCell}, it included 32,000 different instances. It was conjectured that all were winnable, leading to an early example of internet crowdsourcing, the `Internet FreeCell Project' led by Dave Ring in 1994-5 \cite{internet-freecell}.\footnote{Indeed the word `crowdsourcing' itself was not coined until 10 years later \cite{howe-crowdsourcing}.} People shared their solutions online for all instances except one, deal number 11982, which nobody could solve. This is now known to be unwinnable \cite{SolLab}. At a similar time, Don Woods obtained an estimate of 99.999\% winnability for \gamename{FreeCell} from a computer study of a million random instances \cite{SolLab}.\footnote{This is a pleasing example of a case where `99.999\%' is not hyperbole for `almost always' but is the scientifically established value to 5 significant figures.} 

Since then, more computational experiments have given winnability estimates for a variety of games. 
These have been done both inside and outside the academic community. Some games have attracted academic attention, 
including \gamename{Klondike} \cite{klondikemcts,rollout1,rollout2}, \gamename{FreeCell} \cite{paul2016optimal,genetic1,fcnn},  \gamename{Gaps} \cite{helmstetter2004searching}, \gamename{King Albert} \cite{DBLP:journals/corr/Roscoe16} and \gamename{Black Hole}  \cite{gent2007search,smith-caching}.   
For most patience games, however, the best known winnability estimates have been obtained by enthusiasts rather than academic or industrial researchers.  Of games just mentioned,  this includes \gamename{FreeCell} \cite{fish_fc_billiard}, and included \gamename{Klondike} \cite{shootme_klondike} and \gamename{Black Hole} \cite{fish_bhstats} until the current paper.   We are not aware of any academic predecessor to our work which studied a diverse range of patience games, but there have been substantial efforts across a range of games by enthusiasts including
Shlomi Fish, Mark Masten \cite{masten_winrates},  and Jan Wolter \cite{wolter_analysis}, among others.  In summary, the world has owed far more to non-academic than academic research in knowing the winnability of patience games.   We have used ideas from both academic and non-academic researchers. For example, for \gamename{Klondike} and \gamename{Canfield} we made essential use of both the $\textrm{K}^+$ representation of stock from academic research \cite{rollout1} and  the dominance described in \Cref{sec:dominance-tableau} from non-academic research \cite{wolter-code,shootme_klondike}.   Note, however, that this paper is not intended to give a complete survey of either academic or non-academic work on patience games.

We close this brief history with a remarkable echo in our work of the origin of computer-based Monte Carlo methods, which are the method we use to compute winnability estimates throughout this paper.\footnote{Within the wide field of Monte Carlo methods, we are using `Simple Monte Carlo', where large numbers of randomised simulations are run to estimate a parameter \cite{wiki:SimpleMonteCarlo}. We are not, for example, using a method such as Monte Carlo Tree Search \cite{10.1007/978-3-540-75538-8_7}.} 
Monte Carlo methods were actually invented by Stanislaw Ulam with the idea of calculating the winnability of solitaire games, as he recalled:

\begin{quote}
\textit{The first thoughts and attempts I
made to practice [the Monte Carlo
method] were suggested by a question
which occurred to me in 1946 as I was
convalescing from an illness and playing solitaires. The question was what
are the chances that a Canfield
laid out with 52 cards will come out
successfully? After spending a lot of
time trying to estimate them by pure
combinatorial calculations, I wondered
whether a more practical method than
``abstract thinking'' might not be to
lay it out say one hundred times and
simply observe and count the number
of successful plays. This was already
possible to envisage with the beginning
of the new era of fast computers,
and I immediately thought of problems
of neutron diffusion and other questions of mathematical physics, and more
generally how to change processes described by certain differential equations
into an equivalent form interpretable
as a succession of random operations.
}

-- Stanislaw Ulam, unpublished remarks 1983, quoted by \citet{ulam-solitaire}.
\end{quote}

\noindent In this paper, we therefore achieve the idea of Ulam,  giving a very precise estimate of the winnability of the solitaire he was playing using precisely the  Monte Carlo methods he invented to achieve this. As discussed above, we do not know whether Ulam's `Canfield' was the game we call \gamename{Klondike} or \gamename{Canfield}. Whichever it may be, in this paper we have reduced the uncertainty of its winnability by a factor of more than 30 over the previous best estimate and computed a 95\% confidence interval within $\pm 0.1\%$.
 \section{Results Summary}
\label{sec:results}

We have experimented on numerous patience games. Our results fall into three categories: those for games already studied, those for main games which have not been studied before, and finally an extensive investigation into how varying the rules of \gamename{Klondike} affects winnability.  We provide summary of our winnability estimates for each game in the following three subsections.  Details of how these results were obtained occupy the bulk of the rest of this paper.  
Our focus in this paper has been on winnability, rather than accurate measures of time used for benchmarking purposes. However, we provide summary data of time used and nodes searched in the experiments reported in this paper in 
Table~\ref{tab:times}, page ~\pageref{tab:times}. 
Overall, the experiments reported in this paper used about 30 years
of CPU-time.

\subsection{Results for Previously Studied Games}

Solvitaire is able to solve a wide range of previously-researched games, although we have not extended it to be able to search  every game that has already been studied. Results are shown in Table~\ref{tab:solpercs-comparison}, page~\pageref{tab:solpercs-comparison}. In most cases we improve on previous results, and in some very famous games the improvements are dramatic. For example, we have improved the 95\% confidence interval for both \gamename{Klondike} and \gamename{Canfield} by a factor of 30 over the previous best known results. We have also used Solvitaire to identify bugs in previous solvers for those two games: see \Cref{canfield-correction}. 
All results except for \gamename{Gaps (One Deal)},  \gamename{Spider}, and two variants of \gamename{Klondike}, have a 95\% confidence interval  within $\pm 0.1\%$, and this is the first time this has been achieved for ten of these games. 
There are games where Solvitaire is not at good as existing solvers, as we discuss further in \Cref{sec:evaluation}.

\begin{table}
\caption{Comparison with previous work using a consistent methodology for calculating 95\% confidence intervals (CI)  described in \Cref{sec:stats}.   For numbers used for calculation of 95\% CI values,  see
Appendix \ref{app:summary-stats} (for Solvitaire)  
and 
Appendix \ref{app:literature-data} (for data from the literature).
State-of-the-art results for each game are in bold.   
Italics indicate results from the literature open to doubt, see accompanying note.
$\thoughtful$ Thoughtful variant where position of all cards known at start.}
\begin{footnotesize}
\begin{tabular}{| l c c r| }
\hline
    Game \hfill Variant& Solvitaire 95\% C.I.& Best Other 95\% CI & Citation\\ 
\hline

{Accordion \hfill \thoughtful}&
    \notsota{99.99948}{0.00052} & 
    \sota{99.9999936}{0.0000064}
    & \tableCiteA{masten_winrates}   \\ \hline  
{Baker's Game }
    & \sota{75.053}{0.028} & \sota{75.011}{0.028} & \tablecite{Pringle}{pringle_bakers,pringle-pc}
    \tableNoteA{Using a solver by Shlomi Fish} 
    \\ \hline  
{Black Hole} 
    & \sota{86.944}{0.022} & \notsota{86.986}{0.053} 
    & \tablecite{Masten}{masten_winrates} 
    \\ \hline  
{Canfield \hfill \thoughtful}  
    & \sota{71.245}{0.031} & \queryable{71.872}{1.059} &\tablecite{Wolter}{wolter_analysis} 
    \tableNoteA{For discussion of Wolter's code, see \Cref{canfield-correction}} \\  \hline
    
{Eight Off}  & \notsota{99.8805}{0.0022} & \sota{99.8801}{0.0010}
& \tablecite{Masten}{masten_winrates}  \\  \hline

{Fore Cell} & \sota{85.617}{0.024} & \notsota{85.605}{0.385} &  \tablecite{Keller}{SolLab} \tableNoteA{Michael Keller reports results obtained by Danny A. Jones}
\\  
{\dittostraight \hfill Same Suit}& \sota{10.564}{0.020} & \notsota{10.556}{0.061} & \tablecite{Masten}{masten_winrates}
\tableNoteA{Fore Cell (Same Suit) is the same game as Eight Off (4 Depots) }

 \\ 
\hline

{FreeCell}& \hspace{-0.75em} \notsota{99.998881}{0.000207} &
\sota{99.998812}{0.000008} 
& {\tablecite{}{fish_fc_billiard}} 
\\ 
\dittostraight\hfill 0 Cells& \notsota{0.2137}{0.0031} & \sota{0.2173}{0.0012} &
\tableCiteA{fish_freecell_0cell_stats}\\
\dittostraight\hfill 1 Cell~~& \sota{19.348}{0.093} & \notsota{19.519}{0.291}
& \tableCiteA{SolLab}\\
\dittostraight\hfill 2 Cells& \sota{79.544}{0.091} & \notsota{79.468}{0.126}
& \tableCiteA{fish_freecell_2cell_stats}\\
\dittostraight\hfill 3 Cells& \sota{99.3583}{0.0162} & \sota{99.3608}{0.0167 }
& \tableCiteA{SolLab}\\
\dittostraight\hfill 4 Piles& \sota{0.00866}{0.00058} & \notsota{0.02162}{0.01496}
& \dittostraight \\
\dittostraight\hfill 5 Piles& \sota{3.859}{0.040} & \notsota{3.996}{0.248}
& \dittostraight \\
\dittostraight\hfill 6 Piles& \sota{61.421}{0.098} & \notsota{61.719}{0.738}
& \dittostraight \\
\dittostraight\hfill 7 Piles& \sota{98.857}{0.023} & \notsota{98.875}{0.119}
& \dittostraight 
\\

\hline
Gaps \hfill One Deal &
\sota{85.815}{3.717} &
\notsota{89.310}{9.124}&
\tablecite{}{helmstetter2004searching}
\\ 

{\dittostraight \hfill Basic Variant} &
\sota{24.902}{0.028}& \queryable{24.809}{0.847} &

\dittostraight

\tableNoteA{For Basic Variant, raw results unstated in cited paper.}

\\
\hline

{Golf \hfill \thoughtful} & \sota{45.109}{0.032} & 
\notsota{45.077}{0.309} & {\tableCiteA{wolter_analysis}}  \\ \hline

{King Albert} & \sota{68.542}{0.092} & 
\notsota{71.189}{8.678} & 
\tableCiteA{DBLP:journals/corr/Roscoe16,roscoe-pc}
   \\\hline

{Klondike \hfill \thoughtful} & \sota{81.945}{0.084}
&\queryable{84.175}{2.998}
&  \tableCiteA{shootme_klondike} 
 \\
\dittostraight  \hfill Draw 1 & \sota{90.480}{0.116}  & \queryable{92.589}{2.545} &  \dittostraight \\

\dittostraight \hfill Draw 2 & \sota{88.620}{0.135}  & \queryable{91.213}{3.121} &   \dittostraight \\ 
\dittostraight \hfill Draw 4 &  \sota{69.337}{0.098} &  \queryable{71.111}{3.102} &  \dittostraight  \\ 
\dittostraight \hfill Draw 5 & \sota{53.434}{0.099} & \queryable{52.640}{3.139} &  \dittostraight  \\ 
\dittostraight \hfill Draw 6 & \sota{35.854}{0.095} & \queryable{34.559}{2.942}&  \dittostraight  \\
\dittostraight \hfill Draw 7 & \sota{23.779}{0.084}  & \queryable{23.402}{2.618} & \dittostraight  
 \tableNoteA{For discussion of Birrell's code, see \Cref{canfield-correction}}
 \\ \hline
 
{Late-Binding Solitaire} & \sota{47.021}{0.032}
& \notsota{45.418}{3.081}
&{\tableCiteA{accordion-stanford}}\\
\hline
  
{Seahaven Towers} & \notsota{89.332}{0.020}
& \sota{89.319}{0.016}& \tablecite{Masten}{masten_winrates} 
\tableNoteA{Using a solver created by Don Woods}
 \\ \hline

{Simple Simon}& 
\sota{ 97.450}{0.034} & 
\notsota{94.910}{5.090}
&  \tableCiteA{fish_simplesimon} 
\\ \hline 

{Spider \hfill \thoughtful}  
&\notsota{98.487}{1.513}
 & \sota{ 99.9886}{0.0114} 
 &  \tableCiteA{plspider}  
 \tableNoteA{Literature results mainly computer-solved but some human-solved}\\ 

\hline

{Thirty Six \hfill \thoughtful }  &\sota{94.674}{0.100}
& \notsota{94.488}{0.307} 
 & \tableCiteA{wolter_analysis} \\ \hline 
 
{Trigon}& \sota{15.996}{0.023}
& \notsota{16.008}{0.073}& \tablecite{Wolter}{wolter_analysis} \\
\hline

{Worm Hole} & 
\notsota{99.8886}{0.0074} & 
\sota{99.8906}{0.0065} & \tableCiteA{masten_winrates}\\ \hline
 
\end{tabular}

\label{tab:solpercs-comparison}
\end{footnotesize}
\end{table}

\subsection{Results Only Obtained Using Solvitaire}

\begin{table}
\begin{center}
\caption{Solvability percentage:  estimates of 95\% confidence interval for patiences which were obtained for the first time using Solvitaire. $\dagger$ \gamename{Carpet} experiments were performed by \citet{masten_carpet} with results shown in \Cref{tab:literature-data}.  Other experiments performed by us have results shown in \Cref{tab:solvitaireresults}.  $\thoughtful$
Thoughtful variant where position of all cards known at start. 
}\label{tab:solpercs-new}
\begin{tabular}{| l | rcl |}
\hline
Game & \multicolumn{3}{c|}{Confidence Interval}  \\
& \multicolumn{3}{c|}{Percentage Range} \\ 
\hline
Alpha Star& 47.794\%  & $\pm$ & 0.032\% \\ 
American Canister&5.606\% &$\pm$ &0.015\% \\ 
Beleaguered Castle& 68.170\%&$\pm$&0.099\%  \\
British Canister& 0.000129\%&$\pm$&0.000008\%\\ 
Canfield (Whole Pile Moves) \thoughtful  & {67.562\%} &{$\pm$}& {0.034\%} \\
Carpet \thoughtful $\dagger$ &  87.558\% &  $\pm$ &0.021\% \\
\dittostraight ~ (Pre-founded Aces) $\dagger$\thoughtful& 95.186\% &  $\pm$ &0.014\% \\
Delta Star& 34.413\% &$\pm$& 0.030\% \\
East Haven \thoughtful& 82.844\% & $\pm$ & 0.100\%   \\
Fan & 48.776\% & $\pm$ & 0.099\%  \\
Fortune's Favor \thoughtful & 99.9999879\% &$\pm$ & 0.0000022\%  \\
Mrs Mop & 97.992\%& $\pm$ &0.079\%   \\
Northwest Territory \thoughtful& 68.369\% &$\pm$& 0.094\% \\ 
Raglan & 81.226\%&$\pm$ & 0.085\%   \\
Siegecraft & 99.136\% &$\pm$& 0.020\% \\ 
Somerset & 53.725\% &$\pm$&  0.097\%  \\ 
Spanish Patience& 99.863\%&$\pm$& 0.003\%  \\ 
Spiderette \thoughtful & 99.620\% &$\pm$& 0.018\% \\
Streets and Alleys & 51.187\% &$\pm$& 0.186\%  \\  
Stronghold &97.379\% &$\pm$ & 0.042\% \\
Thirty & 67.454\% & $\pm$ & 0.030\% \\
Will O' The Wisp \thoughtful & 99.9240\% &$\pm$ & 0.0027\%  \\ 
 \hline 
 \end{tabular}
 \end{center}
\end{table}

The second class of results is those on
which Solvitaire is responsible for the only good estimate of winnability that we know of.  For new results, we have limited our presentation of results to those for which we can give a very small confidence interval. In Table~\ref{tab:solpercs-new}, 
we give results for 20 games we experimented on ourselves, including some variants that we invented for the purposes of this paper to illustrate the flexibility of our rule language. Most games we give new results for are single-deck games, but we do report a good estimate for the two-deck game \gamename{Mrs Mop}. Additionally, Table~\ref{tab:solpercs-new} shows results for two variants of \gamename{Carpet}, for which the JSON rules were constructed and experiments performed by \citet{masten_carpet}. 
 All but one of the results shown in Table~\ref{tab:solpercs-new} have a 95\% confidence interval within $\pm 0.1\%$: the exception is \gamename{Streets and Alleys}, for which the number of unknown results limited us to  $\pm 0.2\%$. 
 
 One interesting game not included in Table~\ref{tab:solpercs-new} is one we invented  based on \citeauthor{parlett}'s game \gamename{Black Hole}  with the addition of one free cell: we call the game \gamename{Worm Hole}.   Using Solvitaire, we gave the first good estimate of winnability in an earlier version of our paper, but these results have now been improved on by \citet{masten_wormhole}, as shown in \Cref{tab:solpercs-comparison}. Interestingly, those improvements comes from the Masten's use of a game-specific dominance we were not aware of.

Among the games we study is a stricter variant of thoughtful \gamename{Canfield} (invented for this paper) in which moves of partial piles are not allowed:
our results show that about 3.7\% of instances are winnable with the weaker rules but cannot be won with the stronger.

\subsection{Results on Variants of Klondike and Freecell} 
\label{sec:variants}

As well as their general comment on the embarrassment of not knowing the winnability of \gamename{Klondike}, \citet{rollout2} also commented that ``simple questions such as \ldots \textit{ How does this chance depend on
the version I play?} remain beyond mathematical analysis.''  Solvitaire's excellent performance and flexible rule-language  gives an ideal framework to study this question. We studied a number of variants of the rules of \gamename{Klondike} to investigate how winnability of the thoughtful game was affected.  As with our general results, we undertook both replications/improvements and new studies.  

\begin{figure}[b]
\includegraphics[width=3in]{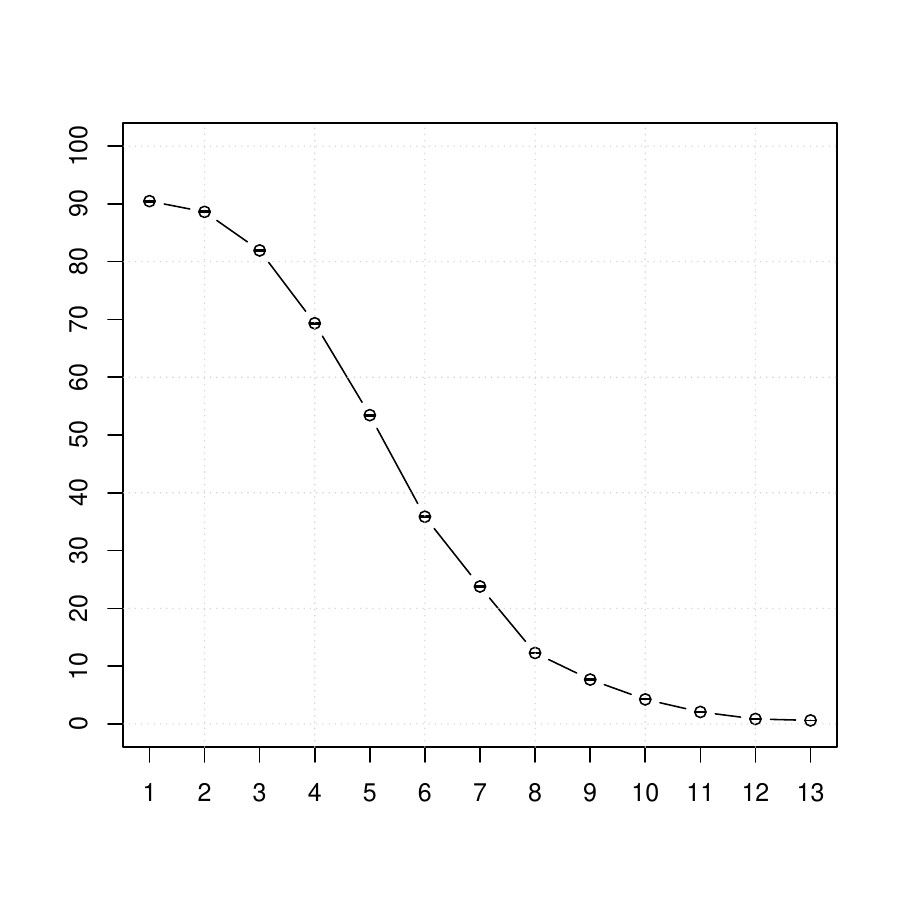} \includegraphics[width=3in]{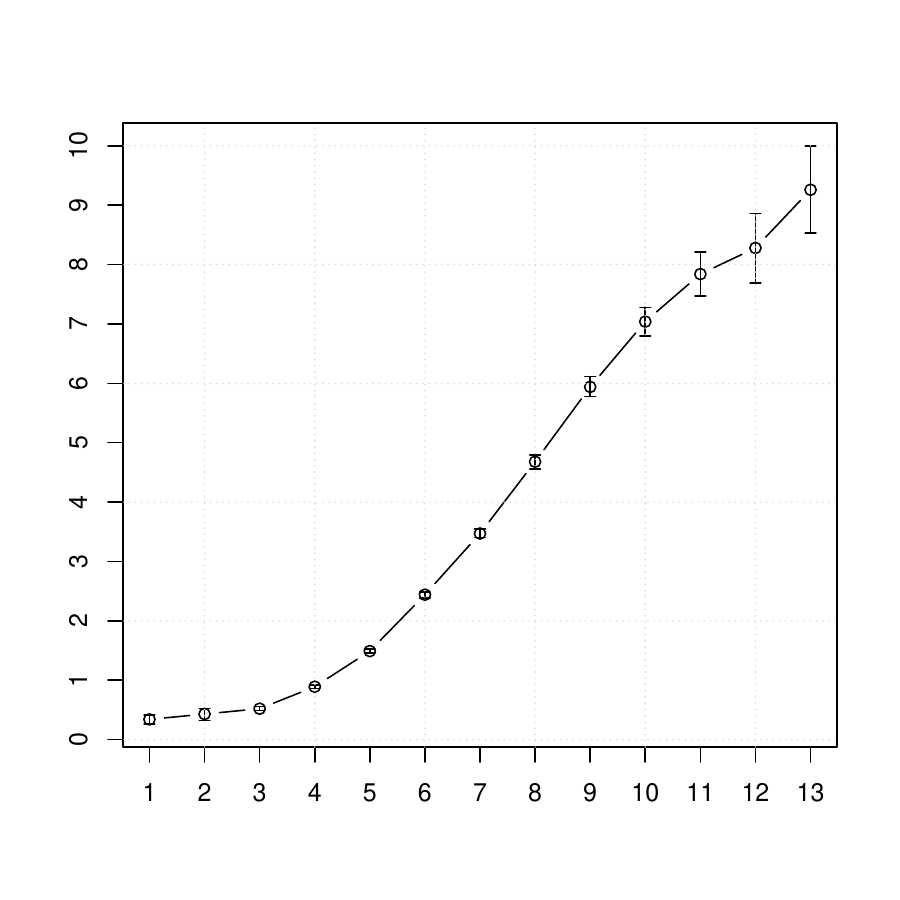}
    \caption{Left: Winnability percentage ($y$-axis) of \gamename{Klondike} with different draw sizes from stock ($x$-axis), but otherwise the same rules as standard  \gamename{Klondike}.  Right: draw size ($x$-axis) against percentage of winnable instances that cannot be won without worrying back at least once ($y$-axis).  In both graphs the horizontal bars show the 95\% confidence interval: in all cases on the left and several on the right, these are significantly narrower than the size of the dot.
}
\label{fig-drawsize}
\label{fig-worryback}
\Description[Graphs showing effect of draw size and worrying back on winnability of Klondike]{Graphs showing effect of draw size and worrying back on winnability of Klondike. These images present data which can be seen textually in Table 3.}
\end{figure}

\begin{table}[ht]
\centering
\caption{Results for \gamename{Klondike} with standard rules except for  varying draw size and whether  worrying back is allowed. The final column shows the percentage of winnable instances where worrying back is necessary: i.e. the instance cannot be won without worrying back at least once. \Cref{sec:stats} describes how the confidence intervals for necessity were calculated.}
\begin{tabular}{|c|rrc|}
  \hline
& \multicolumn{2}{c}{ Worrying Back} &  \\
Klondike& \multicolumn{1}{c}{Allowed} & \multicolumn{1}{c}{Not Allowed} & \\
Draw Size&Winnability (\%) &  Winnability (\%) & Necessity (\%) \\
  \hline
1 &   90.480 $\pm$ 0.116\%
   &  90.204 $\pm$ 0.093\%
   & 0.34 $\pm$ 0.08\%\\ 
  2  & 88.620 $\pm$ 0.135\%
   &  88.289 $\pm$ 0.112\%
   & 0.43 $\pm$ 0.11\%\\ 
  3  & 81.945 $\pm$ 0.084\%
   &  81.524 $\pm$ 0.089\%
   & 0.52 $\pm$ 0.04\%\\ 
  4  & 69.337 $\pm$ 0.098\%
   &  68.723 $\pm$ 0.095\%
   & 0.89 $\pm$ 0.04\%\\ 
  5  & 53.434 $\pm$ 0.099\%
   &  52.638 $\pm$ 0.099\%
   & 1.49 $\pm$ 0.04\%\\ 
  6  & 35.854 $\pm$ 0.095\%
   &  34.982 $\pm$ 0.094\%
   & 2.44 $\pm$ 0.06\% \\ 
  7  & 23.779 $\pm$ 0.084\%
   &  22.952 $\pm$ 0.083\%
   & 3.47 $\pm$ 0.08\%\\ 
  8  & 12.276 $\pm$ 0.065\%
   &  11.703 $\pm$ 0.064\%
   & 4.68 $\pm$ 0.13\% \\ 
  9  & 7.670 $\pm$ 0.053\%
   &  7.214 $\pm$ 0.051\%
   & 5.94 $\pm$ 0.18\% \\ 
  10  & 4.237 $\pm$ 0.040\%
   &  3.939 $\pm$ 0.039\%
   & 7.04 $\pm$ 0.25\% \\ 
  11  & 2.066 $\pm$ 0.029\%
   &  1.904 $\pm$ 0.027\% 
   & 7.84 $\pm$ 0.37\% \\ 
  12  & 0.849 $\pm$ 0.019\%
   &  0.779 $\pm$ 0.018\%
  & 8.28 $\pm$ 0.59\% \\ 
  13  & 0.600 $\pm$ 0.016\%
   &  0.545 $\pm$ 0.015\% 
   & 9.26 $\pm$ 0.74\%\\ 
   \hline
\end{tabular}
\label{tab-worryback}
\end{table}

As a replication, we also experimented on a number of variants of \gamename{FreeCell} that have previously been experimented on, with results in Table~\ref{tab:solpercs-comparison},  page~\pageref{tab:solpercs-comparison}. All results are consistent with previous work, with overlapping estimates of confidence interval.  Several are significant improvements on knowledge.  
Table~\ref{tab:solpercs-comparison} also compares our own results with \citeS{shootme_klondike} reported results for \gamename{Klondike} with varying draw sizes, and our results represent significant improvements.    

For new studies we performed an extensive study of variants of \gamename{Klondike}. An important aspect of this study was to reuse results for one game on related games, as described below in \Cref{sec:relations}: this greatly reduces the time needed to conduct such large sets of experiments. 

Figure \ref{fig-drawsize} and Table \ref{tab-worryback}  show the results on varying numbers of draw size combined with whether or not \patienceword{worrying back} is allowed.  As well as seeing the decline of winnability with increasing draw size, we also see the increasing {\em{necessity}} of worrying back. By `necessity', we mean the percentage of winnable instances that cannot be won without using worrying back at least once.  By  draw size 13, necessity reaches about  9\%. This is to be expected, as the reduced percentage winnability correlates with fewer routes to win, meaning that there are fewer ways to avoid worrying back.

\begin{table}
\caption{Our winnability estimates on variants of \gamename{Klondike} with draw size 3 and worrying back allowed. Rules vary on how cards can be built on in the tableau, and what cards if any may be placed into a space. Standard \gamename{Klondike} is in the central cell.  The entry in italics for Not Allowed/Any Suit is as computed by our protocol but is not a useful confidence interval.}
\centering
\begin{tabular}{|l|rrr|rrr|rrr|}
  \hline
\backslashbox{Spaces\\ Policy}{Build\\ Policy} & \multicolumn{3}{c|}{Any Suit}
  & \multicolumn{3}{c|}{Red-Black} 
 & \multicolumn{3}{c|}{Same Suit} \\ 
  \cline{1-10}

\multirow{1}{*}{Any} 
& \multicolumn{3}{c|}{99.923 $\pm$  0.006\%}  

& \multicolumn{3}{c|}{ 94.959 $\pm$  0.045\%}  

& \multicolumn{3}{c|}{ 40.762 $\pm$  0.097\%}  

\\ \hline

\multirow{1}{*}{King Only} 
& \multicolumn{3}{c|}{ 99.855 $\pm$  0.049\%} & \multicolumn{3}{c|}{81.945 $\pm$  0.084\%} & \multicolumn{3}{c|}{ 6.895 $\pm$  0.050\%} \\ \hline
 \multirow{1}{*}{Not allowed} 
& \multicolumn{3}{c|}{  \textit{51.135 $\pm$  48.759\% } }  

& \multicolumn{3}{c|}{ 2.168 $\pm$  0.121\%}  & \multicolumn{3}{c|}{ 0.178 $\pm$  0.009\%}  

\\
 
 \hline

\end{tabular}

\label{tab-variants}
\end{table}
We also experimented on varying some of the core rules of  \gamename{Klondike}, specifically what is allowed to be placed in spaces and which suits are allowed for building piles.   \Cref{tab-variants} shows the results on \gamename{Klondike} with draw size 3 and nine combinations of rules.   We see that 
 rules can be significantly more effective when combined than individually.  For example, from the most liberal rules (top-left), restricting spaces to kings reduces winnability by only  0.068\% and changing the build policy to red-black reduces winnability by 4.964\%.  However, combining the two restrictions reduces winnability by 17.978\%.

\FloatBarrier

\section{Exhaustive Search using AI Methods in Solvitaire}
\label{sec:exhaustive}
Solvitaire is a depth-first backtracking search solver over the state space of legal card configurations.  For good performance, we needed to improve many aspects of the search procedure from this minimal description, and we use a number of techniques from Artificial Intelligence (AI) to do so. 
We do not claim novelty for these improvements, as  many have  been applied before to patience solvers, singly and in various combinations
\cite{wolter-code,shootme_klondike}, but their use in  combination in a very general patience solver is novel.
In this section we describe relevant aspects of design decisions in Solvitaire and use of AI search techniques. 
After describing our use of depth-first search in \Cref{sec:depth-first}, 
we then describe our use of 
transposition tables in \Cref{sec:transposition},
symmetry in \Cref{sec:symmetry}, 
dominances in \Cref{sec:dominance}, and 
streamliners in \Cref{sec:streamliners}.

The optimisations were included in Solvitaire following informal investigation and  experimentation during the design process.  To give an illustration of their effectiveness in a key game, 
throughout this section we compare how each optimisation affected behaviour in \gamename{Klondike}.  \Cref{table:optimisations}, page~\pageref{table:optimisations}, shows results of Solvitaire with various optimisations enabled or disabled.   While we give a number of  performance indicators, the most important for us is the number of instances that could be correctly resolved within one hour since we wished to avoid instances that cannot be resolved. Experiments for \Cref{table:optimisations} were run on the Cirrus HPC system. CPU Nodes contain 2$\times$Intel Xeon “Broadwell” 18-core cpus, 2.1 Ghz, and 256 GB RAM.   We tested the same 10,000 instances of \gamename{Klondike} in each configuration so performance is directly comparable.

\subsection{Exhaustive Depth-First Search}

\label{sec:depth-first}
A key, early, design decision was to prioritise the ability to determine with certainty whether a given instance of a game is winnable or unwinnable.  A consequence of this was the decision to optimise for efficient \textit{exhaustive} search for unwinnable instances, with less effort devoted to finding solutions quickly.   This led to the choice of depth-first search since it can be implemented extremely efficiently with very little overhead per node searched.   Although transpositions of moves can lead to duplicate search states, we deal with this by using transposition tables, discussed below in \Cref{sec:transposition}.

The core depth-first search process is as follows at each node in search,  starting with the root node being the initial position Solvitaire has been given. 
From the initial position, all possible legal moves are constructed and then one chosen for exploration. This is repeated at each new position.  
If a position is reached where the game has been won, then search is finished.  Alternatively, if no legal moves to a new position are possible, then search backtracks to the last parent of this position and tries an alternative move at that parent.  This will be one of the other possible legal moves at this node previously constructed.
If this process eventually exhausts the possibilities for the starting position, then the instance is proven to be unwinnable. 
Because this process yields complete exhaustive search, if Solvitaire reports that an instance of a game is unwinnable then it has explored every possible way this could be done, so the statement will be correct.  Except for games which are very nearly 100\% winnable, obtaining accurate estimates of winnability requires this kind of certainty.

For efficiency, we use trailing instead of copying to save and restore state in backtracking \cite{trailingVsCopying}.  
Specifically, we  keep a single full copy of the search state which is subject to change when each move is made. At each node in search we store what move is made. Each move is reversible, so when we backtrack the move is reversed to produce the same state as before.

Although not a general AI technique, we use the $\textrm{K}^+$ representation of  stock \cite{rollout1} for patience games with stocks in which there are infinite redeals.  This has proved to be an important optimisation in games  such as \gamename{Klondike} and \gamename{Canfield}.  This replaces the concrete moves which move the stock cards (e.g., three at a time), with calculating which cards can be obtained next using any sequence of individual stock moves.  While it increases the branching rate, 
it also reduces search depth and ensures that each stock move makes concrete progress instead of just moving cards around pointlessly.    
The $\textrm{K}^+$ representation is the only case we implement in Solvitaire where a sequence of several independent moves are combined in a single step to achieve  something useful not possible in one move in the normal rules of the 
game.  
We do not however implement any equivalent of `supermoves' \cite{SolLab} in \gamename{FreeCell} or similar games which do not allow built piles to be moved: a supermove is a sequence of moves which uses spaces to move a built pile from one location to another.  Outside of patience games, in AI Planning the idea of combining moves together as `macro' moves has been used in search \cite{KORF198535,JUNGHANNS2001219,10.5555/1622519.1622534}.  
However, as 
 \citet{JUNGHANNS2001219}  say, 
``special attention must be paid to the side-effects that
macros can have. They might influence the correctness and/or the completeness of the
search.''  Given the extremely obscure bugs that can occur in individual games (see \Cref{sec:canfield-correction}), ensuring correctness of a general macro move sytem would be exceptionally difficult.  Furthermore,  while supermoves can be useful to players, all possibilities have to be considered in exhaustive search, so including them may not be a benefit for unwinnable instances. Nevertheless, exploiting macro/supermoves as streamliners to solve winnable instances more quickly may be valuable in the future.

 A remarkable feature of some games is the extraordinary depths that search can reach while still being successful. For example, in \gamename{Beleaguered Castle}, one instance was solved at a maximum search depth of more than 190 million, with a total of 460 million nodes searched (in 1,270 secs).  That is, the first solution found would have required a player to make more than 190 million moves to win: this is certainly impractical for a human but is nevertheless a legal winning sequence. Another instance was proved unwinnable with a maximum depth of more than 27 million. The latter case involved a total search of just over 1 billion nodes (in 2,131 secs), meaning that the mean number of nodes per depth averaged across all depths visited is less than 40.
 This indicates an unusual search space, since normally one would expect nodes searched to be exponential with depth at a branching rate of at least 2. 
We have not further investigated the nature of the search space, but we can mention some possible factors. 
First, we counted depth as moves in the game, so  sometimes only one move might have been possible. 
Second, even where there are search choices, most of the branches may end rapidly, leading to a very tall but thin search tree.
Third, almost all configurations may be achievable by continuing to move rather than backtracking to the root, with transposition tables (\Cref{sec:transposition}) preventing states being revisited.  
Considering the absurd depths reached, search could accurately be described as going down a very deep rabbit hole. But, given that exhaustive searches were completed, we can say that search was able to completely explore the entire rabbit warren.

The choice of depth-first search has been very successful, as shown by results in this paper, but it  does result in some tradeoffs.   We would mention two in particular.  First, we do not even approximate getting the shortest possible solution: in the example above there may well be a solution at depth 190 instead of 190 million. Second, search can spend a long time in an area where there is no solution after an early incorrect choice, leading to very long search times for instances that might be easily solvable by more flexible methods.   We did consider the use of iterative deepening to avoid the first problem and also possibly the second.  However, preliminary experiments suggested that the overhead of iterative deepening did not pay off for our primary goal of determining winnability.

\subsubsection{Configurable Rule-Sets}

\label{sec:configurable-rule-sets}
An important feature of our solver is that games are not hard-wired into the solver. That is, the input to the solver is a description of the rules of the game in a textual format in JSON, \cite{crockford2006application} 
specifying  values for different aspects of the game. 
As an example,  
the rules of \gamename{Klondike} in this format are shown in  Listing~\ref{rules-klondike}, page~\pageref{rules-klondike}.  
While games like \gamename{Klondike} are provided by name for convenience to the user, this simply means that the JSON is included in the executable rather than preprocessed in any way.
Configurable rule-sets also enables us to alter the rules of existing games to test how they affect the winnability of the game, as we showed in \Cref{sec:variants}.
 However, our rules language does not cater to every possible patience: we were concerned that a much richer language might have made search less efficient. 
 Our chosen tradeoff between expressiveness and complexity enabled us to obtain many new and improved results.
In Appendix \ref{app:rule-description-language} we give the full JSON schema for the rules language, as well as the default values which are used unless overridden.

\lstset{
  basicstyle=\ttfamily\footnotesize,
  columns=fullflexible,
  keepspaces=true,
}

\begin{lstlisting}[float,label=rules-klondike, caption={Rules of \gamename{Klondike} in our JSON format. Note the specification of a dominance in moving built groups to limit the available moves, as discussed in~\Cref{sec:dominance-tableau}.  }]
 "tableau piles": {
      "count": 7,
      "build policy": "red-black",
      "spaces policy": "kings",
      "move built group": "partial-if-card-above-buildable",
      "diagonal deal": true,
      "face up cards": "top" },
 "foundations": {
      "removable": true },
 "stock": {
      "size": 24,
      "deal count": 3,
      "redeal": true }
\end{lstlisting}

The use of a flexible rule description language gives us two huge advantages over all previous work in the area, which has allowed at most a limited flexibility of game definition within a relatively small family.  The most obvious advantage is the wide range of games that can be experimented on without any adaptation at all of the underlying search engine. This can be seen throughout this paper, where we experimented on dozens of very different games, as well as many minor variants of some.  Games that we had not considered at all can be tested just by constructing appropriate JSON input: \citet{masten_carpet} did this to use Solvitaire to find the winnability of two variants of (thoughtful) \gamename{Carpet}. 
The second advantage is that, when we fixed bugs or introduced optimisations for a particular rule, all games using that rule gained the advantage of improved results. For example, the dominance we prove in Appendix \ref{app:dominance-proof-partial-pile}  has previously been used only in special-purpose solvers for \gamename{Canfield} and \gamename{Klondike}  but could be applied without change to 
\gamename{Northwest Territory}, where it massively improved our ability to solve this game.  As well as greater efficiency this enhances robustness of our results, since any remaining bugs for a given rule will have had to escape detection in any game they applied to.

We use a very naive approach to create the list of possible legal moves at each node in search.  Apart from processing the JSON rules for a game into an internal data structure, we do not optimise checking which rules apply.  Solvitaire exhaustively checks possible game rules to find which are being used in the current game, and then whether any lead to possible legal moves in the current position.   This naive approach does have potential inefficiencies. For example if a game does not contain free cells this fact is checked at each node in search instead of just once at the root.
We  do not perform any preprocessing to optimise finding legal moves during search.
At each state, having computed the legal moves we retain the list for possible backtracking. Apart from this, we do not preserve legal moves between states. At each new node, we simply compute this list from scratch. 
 It was a surprise to us that this very straightforward approach was still so effective in practice, but  it is certainly possible that it could be optimised to give even better results. 

{ 
\begin{table}[htbp] \centering 
  \caption{\textbf{Results of variants of Solvitaire on the same 10,000 instances of \gamename{Klondike}.}\newline
Number incomplete in one hour are shown first, and these are not included in other statistics.  Mean cpu time (seconds),  mean number of nodes searched (in kilonodes, i.e. thousands of nodes), and  mean and maximum RAM used (in MB) are given for all determined instances.  Number winnable/unwinnable is also given, with mean  nodes taken for each category.  The column marked $\times$ is only relevant to streamliners: it indicates how many winnable problems the streamliner incorrectly reported as unwinnable.
  All results are to 4 significant figures. \newline
  In each family results for the following base setting is repeated and indicated in bold:  a cache limited to 100,000,000 entries; the use of both the dominance which force moves to foundations when safe and the dominance which limits moves of partial built piles; the use of full symmetry in considering cached states; and the use of no streamliner. }
\begin{tabular}{ |l | r  r   r| r D{.}{.}{2.3} D{.}{.}{4.1} D{.}{.}{5.1}| r D{.}{.}{5.1} r r|  } 
\hline 
  & \multicolumn{3}{c|}{Number} &  \multicolumn{4}{c|}{All Determined}  & \multicolumn{2}{c}{Winnable} & \multicolumn{2}{|c|}{Unwinnable} \\ 
& $>$1hr &\multicolumn{2}{c|}{Determined}  & Knodes & \multicolumn{1}{c}{cpu(s)} & 
\multicolumn{1}{c}{RAM} & \multicolumn{1}{c|}{RAM} & \multicolumn{1}{c}{num} & \multicolumn{1}{l}{Knodes} & \multicolumn{1}{|c}{num} & \multicolumn{1}{c|}{Knodes}  \\ 
&  && & \multicolumn{1}{c}{mean} & \multicolumn{1}{c}{mean}& 
\multicolumn{1}{c}{mean} &\multicolumn{1}{c|}{max}& \multicolumn{1}{c}{} & \multicolumn{1}{c}{mean} & \multicolumn{1}{|c}{} & \multicolumn{1}{c|}{mean}  \\
\hline
{\textbf{Cache Size}}&&&&&&&&&&&\\
\multicolumn{1}{|r|}{1,000,000} & 434 & 9,566 &  & 8,967 & 19.12 & 29.5 & 354.0 & 7,992 & 3,992 & 1,574 & 34,230 \\ 
\multicolumn{1}{|r|}{2,000,000} & 337 & 9,663 & & 8,046 & 17.12 & 47.2 & 706.6 & 8,035 & 3,728 & 1,628 & 29,360 \\ 
\multicolumn{1}{|r|}{5,000,000} & 241 & 9,759 &  & 6,623 & 14.50 & 84.4 & 1,740 & 8,060 & 2,095 & 1,699 & 28,100 \\ 
\multicolumn{1}{|r|}{10,000,000} & 169 & 9,831 &  & 6,442 & 14.25 & 131.3 & 3,474 & 8,086 & 3,093 & 1,745 & 21,960 \\ 
\multicolumn{1}{|r|}{20,000,000} & 113 & 9,887 &  & 6,310 & 14.14 & 200.9 & 6,933 & 8,104 & 2,822 & 1,783 & 22,160 \\ 
\multicolumn{1}{|r|}{50,000,000} & 65 & 9,935 &  & 6,331 & 14.68 & 324.1 & 17,180 & 8,118 & 2,928 & 1,817 & 21,530 \\ 
\multicolumn{1}{|r|}{\textbf{100,000,000}} & 38 & 9,962 & & 7,764 & 18.42 & 440.6 & 34,010 & 8,123 & 3,141 & 1,839 & 28,180 \\ 
\multicolumn{1}{|r|}{200,000,000} & 21 & 9,979 & & 9,628 & 22.86 & 558.0 & 67,500 & 8,127 & 3,618 & 1,852 & 36,000 \\ 
\hline
{\textbf{Symmetry}}&&&&&&&&&&&\\
\multicolumn{1}{|r|}{None} & 195 & 9,805 & & 16,880 & 29.19 & 786.1 & 34,410 & 8,068 & 10,460 & 1,737 & 46,720 \\ \multicolumn{1}{|r|}{\textbf{Full}} & 38 & 9,962 &  & 7,764 & 18.42 & 440.6 & 34,010 & 8,123 & 3,141 & 1,839 & 28,180 \\ 
\hline
{\textbf{Dominance}}&&&&&&&&&&&\\
\multicolumn{1}{|r|}{None} & 481 & 9,519 &  & 33,240 & 63.11 & 1,331 & 34,450 & 7,908 & 26,920 & 1,611 & 64,260 \\ 
\multicolumn{1}{|r|}{Safe 
moves} & 294 & 9,706 &  & 28,710 & 57.79 & 1,066 & 34,360 & 8,004 & 20,830 & 1,702 & 65,760 \\ 
\multicolumn{1}{|r|}{Partial pile} 
& 92 & 9,908 &  & 10,620 & 25.30 & 652.9 & 34,280 & 8,109 & 4,685 & 1,799 & 37,360 \\ 
\multicolumn{1}{|r|}{\textbf{Both}} & 38 & 9,962 & & 7,764 & 18.42 & 440.6 & 34,010 & 8,123 & 3,141 & 1,839 & 28,180 \\ 

\hline
{\textbf{Streamliner}}&&&{$\times$}&&&&&&&&\\
\multicolumn{1}{|r|}{Foundations} & 13 & 9,987 & 120 & 6,416 & 14.22 & 346.7 & 33,840 & 8,007 & 2,723 & 1,980 & 21,350 \\ 
\multicolumn{1}{|r|}{Suit} & 0 & 10,000 & 1 & 3,161 & 7.691 & 179.2 & 33,910 & 8,130 & 980.2 & 1,870 & 12,640 \\ 
\multicolumn{1}{|r|}{Found.+Suit} & 0 & 10,000 & 126 & 2,355 & 5.280 & 133.5 & 33,070 & 8,005 & 871.2 & 1,995 & 8,309 \\ 
\multicolumn{1}{|r|}{Smart} & 33 & 9,967 & 0 & 6,883 & 15.84 & 368.6 & 37,660 & 8,128 & 939.4 & 1,839 & 33,150 \\ 
\multicolumn{1}{|r|}{\textbf{None}} & 38 & 9,962 &  0 & 7,764 & 18.42 & 440.6 & 34,010 & 8,123 & 3,141 & 1,839 & 28,180 \\ 

\hline 
\end{tabular} 

    \label{table:optimisations} 
\end{table} 
}


\subsection{Transposition Tables}

\label{sec:transposition}
We use transposition tables  \cite{Greenblatt_transposition,smith-caching} to avoid trying the same position twice. To do this we record every attempted position in a cache. Any position we might consider which is already in the cache can be ignored: its existence in the cache means that it would be potentially explored twice.
\citet{DBLP:conf/socs/AkagiKF10} show that the use of transposition tables can lead to suboptimal solutions, but this is not an issue for us as our design goal was simply to find any solution rather than optimal ones.

Some care is needed to ensure that a cache hit correctly links to a previously explored position, so it is important to ensure that a complete game state is stored in the cache. For example, if the cache does not record whether cards in the layout are face-down or not, then obscure bugs can result.
 We never need to retrieve any data from the cache except the existence of the state, so to save space we store a compressed representation of the state. For each component of the layout (stock, tableau pile, etc.) the cards in that component are listed in order.  Also, we need to take care in recording points such as which cards are hidden and face-up.
 This is not a highly optimised representation but is much smaller than the representation used for the active state.  
   A secondary use of transposition tables is to avoid loops, i.e. a sequence of moves which arrives in a state previously visited as a parent of the current node. This actually reduces to the same case as the general one. If the transposition table becomes full, we discard elements on a least-recently-used basis. The exception is that we never discard any ancestor of the current state, as otherwise loops can occur. If the transposition table is entirely full and all states in it are ancestors, then we give up on search and report that a memory-out has occurred. In extreme cases very large amounts of RAM are necessary, up to hundreds of gigabytes of RAM in some of the hardest problems we solved.

The first set of experiments in \Cref{table:optimisations} shows  how performance varies with size of transposition table.  Increasing  cache sizes give better results, and we see no point of diminishing returns in our experiments.   With a one million sized cache more than 4\% of instances remained unresolved, while with the largest size of 200 million, this reduced to 0.2\%.   However, this improvement does come at considerable space cost.  As would be expected, we see the RAM usage increase linearly with the size of cache.  While the mean usage remains reasonable, the worst case with the largest cache was a requirement of 67GB.  In our experiments, this limited the number that could be run simultaneously on a single machine.  The conclusion seems clear, that one should use the largest cache that is consistent with the resources available.  It also suggests that a more highly optimised cache representation could lead to better results by using less memory.

\subsection{Symmetry}
\label{sec:symmetry} 
Symmetry in search problems has often been pointed out as an issue which can lead to much redundant search \cite{gent2006symmetry}. That is the case in patience games where we can have equivalent but non-identical positions. A common example in patience games is that all spaces in the tableau are equivalent. We should not waste time trying a card in a second space if it did not work in the first. 
The use of symmetry is also related to transposition tables, because it means that a single cached state can represent many future states in the game.  This is because layouts which differ only in the order of piles are considered identical.    More subtly, if a sequence of moves precisely swaps two complete piles from an original position, then we should stop search as we have just returned to an equivalent position. We take a simple but effective approach to avoid this problem. Before storing states in a cache we reduce them to a canonical form, maintaining each group of indistinguishable locations such as tableau piles and free cells in a sorted order. For efficiency this order is maintained incrementally during search.
Additionally, where a game does not use suits in any way (for example \gamename{Black Hole}) the canonical form can discard suit information for greater reduction. 

Choice of when to use symmetry breaking techniques can be handled automatically.  None of our rules allow distinction between different tableau piles, free cells etc, so these can be safely assumed to be indistinguishable.   On the other hand, whether or not suits are indistinguishable depends on the rules of the game.  But the rules language (Appendix \ref{app:rule-description-language}) can be checked for components which depend on suit, such as building to foundation or building within the tableau. If no rules do have this dependency then suit symmetry can be added to the use of the transposition table.  
\Cref{table:optimisations} clearly shows that switching symmetry off increases the number of unresolved instances five-fold. In this case there is no RAM penalty compared to use of transposition tables alone, so this is an unambiguous win.

\subsection{Dominances}
\label{sec:dominance-description}  
\label{sec:dominance}

The use of `dominances' has proven to be important in AI search \cite{Chu2015}. A dominance occurs when we can commit to not considering some legal transition in the search space, having detected that a solution in which we make that transition is `dominated' by an alternative sequence in which we do not make the transition.   
As an example from 1962, the `pure literal' rule in the classic DPLL algorithm is a dominance \cite{DBLP:journals/cacm/DavisLL62}.  
The general idea has been widely used in search problems in many areas of AI, for example as stubborn sets in verification \cite{10.5555/647736.735461}, partial order reduction in planning \cite{10.5555/3038546.3038581}, and automatic move pruning in single-player games \cite{DBLP:conf/socs/BurchH11}.

There are two key types of dominances in searching patience games. The first is a move which we can commit to making in a given situation and therefore avoid backtracking from the choice, because we know that if any solution exists, there is a solution where this move is made next. The second is a possible move which we can decide not to attempt at all, because we know that if that move leads to a solution, there is another way of winning the game without making that move next.\footnote{Because our focus is on winnability of games, we do not insist that the safe sequences be the same length or shorter, so the dominances we  use might not be appropriate in searches for the shortest winning sequence.}

Although not under that name, previous workers on patience have recognised the importance of dominances since they can greatly reduce the search space to explore \cite{Keller-Dominance,masten_wormhole,wolter_canfield,shootme_klondike}. 
However, there are some issues with the use of dominances.  First, it can be easy to be misled into thinking some proposed dominance is correct when it can actually lead to bugs.  We discuss bugs we found related to dominances in our own and other solvers in \Cref{canfield-correction}.  Second, and closely related, dominances have been used without being proven correct including the most widely used.  In this paper we therefore  give proofs of the two dominances we use, in Appendix \ref{app:dominance-proofs}.   
In \Cref{sec:safemoves}, we discuss  the most commonly used dominance in playing patience games, allowing moves to foundations to be committed to. 
In \Cref{sec:dominance-tableau}, we discuss an important dominance which applies to key games like \gamename{Klondike} and \gamename{Canfield}, and which can greatly improve Solvitaire's performance.

\subsubsection{Safe Moves To Foundations}
\label{sec:safemoves}

In many patience games, the goal is to move cards to the foundations.  Beginners often make such moves whenever possible, but this is not always safe. However, an important family of dominances make these moves when it is genuinely safe to do so, and can thus be used to reduce search.  

The most typical games build up by suit on the foundations but build down in alternating colour on the tableau.   In such games we can automatically move a card to the foundation if it is at most \textit{two} more than the current card on foundations of the opposite colour and at most \textit{three} more than the current card on foundation of the other suit of the same colour \cite{SolLab,Keller-Dominance}.
For example, if the foundations have been built to $8\clubsuit$, $7\diamondsuit$, $9\heartsuit$, $8\spadesuit$, it is safe to build the $10\heartsuit$ from tableau to foundation unconditionally. The only use we could have for the $10\heartsuit$ is to put a black 9 on it, which in turn can only be used to put the $8 \diamondsuit$ on.  But all these cards could instead - and preferably - go to the foundation immediately, so there is no need for them on the tableau and therefore not for the $10\heartsuit$.  Following this, it would not be safe to put up the $J \heartsuit$ to foundation, because we might want to keep it to build down $10\spadesuit$ and $9\diamondsuit$.\footnote{It is unclear where this dominance originated, perhaps being invented independently multiple times. \citet{Keller-Dominance} described it as `a clear and obvious rule' and states it was implemented in some of the earliest FreeCell
programs.} 

If a game does not allow worrying back, then we can use a slightly stronger rule.  The rule as above applies but we can also move to foundation unconditionally if the card is no more than one higher ranked than the foundations of the opposite colour \cite{Keller-Dominance}. The reason is that there are no cards of the opposite colour that can possibly be built in the foundation onto this card.
On the other hand, when a game does allow worrying back, then we can add a related dominance.  We can ban worrying back from foundations to tableau if the card replaced on the tableau would be eligible for automatic movement to the tableau under the first dominance: such a move would lead to a pointless loop. While seemingly minor, this is important as it ensures that progress is not reversed unnecessarily.   This is a slightly stronger and generalised version of a dominance proposed by \citet{rollout1} for 
\gamename{Klondike}.

Similar, but less complex, dominances are available with other building rules than the standard red-black.  If the build policy is by suit, then we can always require cards to be moved from the tableau to foundations if they can be, since no other card can be built onto them.  If the build policy is that building is regardless of suit, then we can move a card to foundation if it is no more than two higher than the lowest card yet built to foundation.

These dominances apply to moving cards from the tableau, as well as from a free cell or the reserve.  However, it is not safe to enforce this dominance from the stock - as we discuss in \Cref{canfield-correction}. The exception is when the stock draw size is 1 and infinite redeals of stock are allowed: in this case the stock can be treated as if it were a reserve.   

Solvitaire implements all the preceding dominances. While well known, these dominances have not been proven correct.  Accordingly we prove them correct in Appendix \ref{app:dominance-proof-safe-moves}. 
All of the preceding discussion concerns single-deck games, since that is what our proof covers: some adjustment to the dominance would be necessary for multiple-deck games.

As well as reduction in search space, when a safe move is available we can save space in the transposition table.  If some move would be made by the dominance there is no need to enter the state into the transposition table.  We can make all available safe moves and then only store the state when no more are available.  If any state reoccurs then the safe moves will be made a second time and the final state at the end of the sequence will be found again.

\subsubsection{Tableau Moves of Incomplete Piles}
\label{sec:dominance-tableau}
In studying code by \citet{wolter-code} for \gamename{Canfield} 
and \citet{shootme_klondike} for \gamename{Klondike}, we noticed an interesting dominance in both.  This is that moves of built piles on the tableau are only allowed if \textbf{either} the entire pile is being moved \textbf{or} only a part of a pile is being moved and it is possible to build to the foundation the card above\footnote{For clarity, in the built pile $10\clubsuit9\heartsuit8\spadesuit$ we  say  the $10\clubsuit$ is \textit{above} the $9\heartsuit$  while the $8\spadesuit$
is \textit{below} the $9\heartsuit$. 
The possible confusion is that $10\clubsuit$ is placed physically underneath the $9\heartsuit$ when played on a table.
} 
the top card in the built pile being moved.
Our experiments failed to show any case where this optimisation changed results. 
Wolter has died and \citet{birrell-pc} did not have a correctness proof. We have not found this optimisation documented in the literature, and its correctness is not obvious, so we give what we believe to be the first correctness proof of this dominance.    

In Appendix \ref{app:dominance-proof-partial-pile} we  generalise the dominance to make it apply more widely, and then give the detailed proof of correctness.   
We actually prove a slightly stronger version of the dominance, that the card above must not only be buildable to foundation in principle, but must actually be built to foundation immediately.   However, as we had not yet noticed this potential improvement, the weaker restriction as suggested by Wolter and Birrell is what we implemented in  Solvitaire's code and experiments we report in this paper.
An analogue of the stronger restriction for the game of Worm Hole has proved to be important in effective search 
\cite{masten_wormhole}. The importance of similar techniques in different circumstances illustrates the need to be able to reason more effectively about dominances in general, to allow them to be used when it is correct to do so, without the need for the detailed kind of proof we presented here.

We can give the intuition behind the dominance which also plays a key role in the proof.  
Suppose we make a partial pile move but do not immediately build the card above it to foundation.  This means the partial pile move was not really urgent so we can delay it until later, or even not do it at all.  This is straightforward in all but one case.  The exception is where the very next move is to build a different card on the card that we just vacated.  
We can illustrate by example in the case of a red-black build policy: consider the move of a three card pile $10 \clubsuit9\heartsuit8\spadesuit$ from the J$\diamondsuit$ to J$\heartsuit$, followed immediately by a move of the 10$\spadesuit$ to the J$\diamondsuit$.  This makes simply delaying the first move impossible  as it would invalidate the second move, so we have to take another approach. 
In this case we cancel the move of $10 \clubsuit$ and change the following move by moving the move 10$\spadesuit$ to the J$\heartsuit$ instead of J$\diamondsuit$.   This causes no significant problem until we  want to build the J$\diamondsuit$ to foundation, but it might now be covered by the $10 \clubsuit$ when it was previously free.  But if this happens, the J$\heartsuit$ must be free itself by a symmetry argument that the J$\diamondsuit$ was originally free to move to foundation.  So we can  \textit{now} move  $10 \clubsuit$ from J$\diamondsuit$ to J$\heartsuit$.  As required by the dominance, the next move will be of the J$\diamondsuit$ to foundation. So we have shown that a game won without complying with the dominance rule can also be won complying with it.  Full details of all cases are given in Appendix \ref{app:dominance-proof-partial-pile}.

Importantly, we also prove, in \Cref{thm:dominance-combined}, page~\pageref{thm:dominance-combined}, that the  above two dominances  are also compatible: i.e. if both apply separately then their combined use cannot lead to incorrect results.

\subsubsection{Implementation in Solvitaire}

To exploit dominances we adapt the search process when finding and making legal moves.   For a safe-tableau move, if any is available then one is made immediately and no alternative moves are stored for backtracking. Also, the state does not need to be recorded in the cache because if the state was revisited then dominance moves would be repeated so it is enough to store the endpoint of a sequence of dominance moves. 
However we do still record the move for purposes of reversing it later during backtracking.  For 
the incomplete pile dominance, if it applies we do not consider a move legal if it moves a partial pile where the card above cannot be built to foundation.  However, we do not force the next move to be of the card above to foundation though \Cref{thm:dominance} would allow that: this is simply because we did not realise the stronger rule was valid when we implemented Solvitaire.

Dominances that seem correct can easily turn out to be unsafe, as we discuss in \Cref{canfield-correction}.  This is a particular problem when using the general rule language such as provided by Solvitaire.   Unusual combinations of rules may invalidate a dominance which is valid in very similar games.
For the dominance of \Cref{sec:dominance-tableau}, we require it to be specified in the JSON statement of the rules of the game, avoiding automatically applying it incorrectly.  In experiments for this paper we only add the dominance when the game meets the conditions of \Cref{thm:dominance}. 
We do automate the use of the dominance of \Cref{sec:safemoves} but only under strict conditions.  First, it is disabled completely for games of more than one deck, for games with \gamename{Spider}-type building rules, or games like \gamename{Gaps} without either foundations or a hole.
Second, the move has to be from tableau, free cell or reserve, with one exception: the exception is that moves from the stock are allowed if there are unlimited redeals and the draw size is 1, since in this case the stock is actually equivalent to a reserve. 
Finally, when the dominance is allowable, the game rules are checked for what the build policy is and whether worrying back is allowed. The relevant dominance from \Cref{sec:safemoves} is then applied.

\Cref{table:optimisations} shows performance with all combinations of the two key dominances that we use.  These both prove to be very important: if neither is used we fail to resolve more than ten times as many instances as when both are.  While the partial pile restriction is more critical in \gamename{Klondike}, both dominances should clearly be used.

\subsection{Streamliners}
\label{sec:streamliners}
The final AI technique that we use is `streamliners' \cite{streamliner-gomes,wetter2015automatically}.  A streamliner imposes an additional property which does not necessarily hold in all solutions. A good streamliner is a property that greatly reduces the search space while also having a good chance of leaving at least one solution. While not under the name `streamliner', the general idea of interleaving incomplete and complete searches  has been used in other contexts within AI Search.\footnote{For example, \cite{Lipovetzky_Geffner_2017} describe  the process in their own and the FF planner \cite{10.5555/1622394.1622404} as being `dual', meaning a ``slow but incomplete search, the planner front-end, is followed if not successful, by a slower and complete search, the planner back-end."}
Past patience researchers have  used the idea of running a solver which might produce false negatives, thereby speeding up cases where a solution can be found \cite{fish_simplesimon}, 
but our implementation generalises this across games. 

We use two general streamliners. First, in a game in which cards are moved to foundations, always make such a move when it is possible to do so.  This is a very common technique of human players and  massively reduces the search space while typically allowing most (but not all) winnable instances to be won. When used, this is implemented by treating moves to foundation in a similar way to dominances,  making the move immediately when available and not backtracking on this choice.
Second, we pretend that cards have more symmetry than they do to increase the chance of cache hits. This is very relevant to games which build down in red-black order on the tableau, but up in suits on foundation.  If we have a position that differs from a previously visited state only in suits (but not in colours) in the tableau, it is very unlikely to succeed if the first one does not.  Exceptions do occur because of the differences between suits, but again the tradeoff is good for this streamliner.    This is implemented in the same way as if the symmetry did in fact apply, by discarding suit information when storing and checking states in the transposition table as discussed in \Cref{sec:symmetry}.

In Solvitaire, the user chooses via command-line option whether to use one, both or neither streamliner.  This is a run-time option since there are games where streamliners cannot possibly help. If there is a solution found with a streamliner then we have proved the instance is winnable, but if not then we have to start search again without that property holding.
To facilitate this, we provide a command-line option to do this automatically under the name `smart streamliner'. 
When this option is used we allocate 10\% of the original time-limit for a streamlined search and if that fails to prove the game winnable, we allocate the original time-limit for a search with no streamliner.  For many games, the streamlined search very commonly finds a solution very much faster than the full search would do, leading to greatly improved performance over a large set of instances.

\Cref{table:optimisations} shows performance of streamliners on \gamename{Klondike}. Notice that both streamliners  can yield false negatives, both individually and together, but they do greatly reduce runtime.  For this reason the best combination is the `smart' streamliner which first runs for 10\% time with both streamliners: unlike a pure streamliner, this overall process cannot give a wrong result.  Indeed, we see that 
 smart streamliner gives a slight increase in number resolved but also a more significant  improvement in CPU time. It reduces time by a mean of 2.5s per instance, equivalent to about a month of CPU time on our main experiment on a million instances of \gamename{Klondike}.   In other games we see much more dramatic improvements through streamliners,  as shown for example by a more than 40-fold speedup in \gamename{FreeCell} (see \Cref{tab:comparative-freecell-winnable} in Appendix \ref{app:comparative}, page~\pageref{app:comparative}).

 \FloatBarrier

\section{Relationships between Games}
\label{sec:relations}

In some cases one ruleset is \textbf{stronger} than another, in that any legal move in the stronger game would also be legal in the weaker one.  For example, \gamename{Worm Hole}
 is identical to \gamename{Black Hole} but with the addition of a free cell.  Any instance that can be won as the stronger game of 
 \gamename{Black Hole} must automatically be winnable in \gamename{Worm Hole} via the same sequence of moves:
Some examples from the rules of \gamename{Klondike} further illustrate the concept.
\begin{itemize}
    \item Allowing worrying back makes a game strictly weaker.  If we can win the game without worrying back then we can make the same sequence of moves in the game that allows worrying back. 
    \item Allowing nothing to be put into empty spaces is strictly stronger than allowing only kings to put in spaces, which in turn is strictly stronger than allowing any card to be put in spaces.
    \item Allowing building down in any suit is strictly weaker than building down in red/black ordering.  It is also strictly weaker than building down in same-suit ordering. However, red/black and same-suit orderings are incomparable.
    \item Any draw size from the stock is strictly weaker than any multiple of it.  For example, draw size 2 is strictly weaker than draw size 4 and 6. While draw sizes 4 and 6 are incomparable, they are each strictly weaker  than draw size 12.
\end{itemize}

Where one game is stronger than another,   winnable instances of the stronger game must be winnable in the weaker, and unwinnable instances of the weaker game must be unwinnable in the stronger. We can use this to reduce greatly the set of instances that must be tested to obtain results between games.   

Despite the extreme amounts of time we spent on the hardest \gamename{Klondike} instances, we still found some that were proved unwinnable in weaker games or winnable in stronger ones.  
Of 396 instances that were not solved directly, 7 were found winnable in games where the stock is drawn in units of 6 instead of 3, and one more where the stock is drawn in units of 9.  While valid playthroughs for the original game, the reduced search space in the stronger game allowed the winning moves to be found faster.    A further 231 instances were shown to be unwinnable when cards are drawn from the stock in ones.   It might seem surprising that it is easier to prove the instance unwinnable in a weaker game. The reason is that drawing cards by one allows for a dominance that is not valid when drawing by 3. Drawing cards by one with unlimited redeals makes all cards in the stock available at any time, so we can apply the  dominance described in \Cref{sec:safemoves} to the stock as well as to the tableau: this is invalid with other draw sizes. 
When the dominance is applied it reduces the size of the search space and thus allows all possibilities to be exhausted.  This is an example of \textit{relaxation} in a search problem \cite{10.1287/ijoc.14.4.295.2828}.  
We were able to use these approaches to resolve 239 of the 396 unknown instances, leaving only 157.

When computing results on two games which were strictly stronger/weaker than each other, we used an identical set of instances in each case.  This gives us two significant advantages.  First, it acts to reduce the statistical variance in computing the difference in winnability between the games.  Second, we were able to exploit the linkage between games to avoid recomputing winnability results we already knew. For example, since draw size 5 is strictly weaker than draw size 10, we did not need to test draw size 10 on any of the 465,656 instances proven unwinnable at size 5.  Having found the 42,372 winnable instances with worrying back at draw size 10, these were the only ones we needed to test for winnability without worrying back.  We used this to greatly reduce the time taken to compute accurate winnability percentages across a range of related games. This can be seen in Appendix \ref{app:summary-stats}, which shows for example that we could determine the result of one variant of \gamename{Klondike} on one million instances while testing only 5,997 instances for that specific game.  
\label{sec:stronger-weaker}
This could be used as an additional form of streamliner when searching for solutions for individual patiences - e.g. when trying to win a game with worrying back one could first try it without, which greatly reduces the search space while often not greatly reducing the chances of winning.   This is an interesting area for future research that we have not yet investigated.

 \Cref{tab-variants} shows up a weakness in Solvitaire's ability to solve patience games.  Of a million instances, more than 97\% could not be determined when combining building in any suit with spaces not being fillable.   We discuss this weakness further in \Cref{sec:evaluation}.

\section{Implementation, Testing and Debugging}

Solvitaire is implemented in the C++ programming language.  During development, code was profiled to identify hotspots in code which needed optimisation.  Some areas which did not turn out to be critical were surprising; for example the code to find available moves is barely optimised despite being used at each node in search. 

We used a number of strategies to test our code and reduce bugs to a minimum.
First, we used unit, integration and performance tests to guard against regressions in the code. As we introduced new game features we created bespoke, simplified games to target the added functionality. Our tests were build upon these game types, using hand-crafted instances with known solutions.
We also had a performance benchmark script, which
measured the performance of the solver on a number of benchmark instances to
let us know if our latest code changes had slowed it down.

Second, we ran strict and loose versions of particular games over identical instances. Where Solvitaire reported a looser versions of a game as unwinnable but the stricter version as winnable, a bug was indicated which we then fixed. This can be seen as a form of metamorphic testing, which has also been used in testing constraint solvers which have a similar problem of vast search trees  without knowing results in advance \cite{DBLP:conf/cp/AkgunGJMN18}.

Third, we could test our work on the macroscopic scale, by comparing overall results obtained using Solvitaire on games also estimated by previous researchers.
For example, we discuss in  \Cref{canfield-correction} that this allowed us to identify and fix a bug in our pseudorandom instance generator. Table~\ref{tab:solpercs-comparison} shows that, where we were able to compute confidence intervals for related work, all our 95\% confidence intervals now overlap with the best existing estimate.   
Given the complete independence of our implementations with those of many different past researchers, this strongly suggests that bugs that significantly affect winnability percentages are unlikely.  

Finally, at the microscopic level, for the games \gamename{FreeCell}, \gamename{Canfield}, and \gamename{Klondike}, we tested individual instances to make sure our solver gave consistent results with independent solvers.   For \gamename{FreeCell}, we  ran Solvitaire on each of the 102,075 unsolvable instances of \gamename{FreeCell} found by \citet{fish_fc_billiard}: all were correctly identified as unsolvable except for two that could not be determined. 
We tested Solvitaire against the best existing solvers for each of \gamename{Canfield} \cite{wolter-code} and \gamename{Klondike} \cite{shootme_klondike} on 50,000  individual instances each.  Detailed study of individual inconsistent results allowed us to determine  which solver was correct. If the bug was in Solvitaire, we corrected it.
As discussed in  \Cref{canfield-correction}, we also  found problems in both existing solvers. Although this happened in very rare cases, it indicates the detailed work that allowed us to discover such rare bugs in existing solvers. 

We cannot rule out that bugs remain in our code that might affect winnability of some games, especially using unusual combinations of rules we have not tested exhaustively. Availability of our codebase will enable future researchers to identify any remaining bugs in our code \cite{solvitaire-v0.10.2}. The code includes random generation of instances which is portable across different machines, so other researchers should be able to recreate the same test instances to check our results against theirs.

\subsection{Incorrect Optimisations in Existing Solvers for Klondike and Canfield}
\label{canfield-correction}
\label{sec:canfield-correction}

We tested Solvitaire's results on 50,000 instances each against the best existing solvers for \gamename{Klondike} \cite{shootme_klondike} and \gamename{Canfield} \cite{wolter-code}. Where both the existing solver and Solvitaire determine the answer, they should  both agree that a given instance is winnable or unwinnable. Where there was disagreement on a specific instance, we looked at the solution produced by whichever solver claimed the game was winnable, which we could check by hand for correctness. In some cases the inconsistency was due to a different understanding of the rules, in which case we always revised our rules to match those of the existing solver. Some bugs remained, and where the bug was in Solvitaire we corrected it, but some bugs were found in existing solvers.

We discovered the same incorrect dominance in both an earlier version of Solvitaire and in \citeauthor{shootme_klondike}'s  
Klondike Solver. This concerned worrying back, i.e. returning a card from foundation to the tableau. It might seem that it would be unnecessary ever to do this immediately after placing the same card from tableau to foundation, but we can construct instances in which it is necessary. In one such example, we move the 3$\clubsuit$ to foundation, revealing the previously hidden 4$\heartsuit$: the only winning continuation is to reverse this immediately, then move the 2$\heartsuit$ onto the 3$\clubsuit$, uncovering the 5$\clubsuit$ onto which we can now move the pile under 4$\heartsuit$. We believe Klondike Solver incorrectly reports six of its first 50,000 random instances  to be unwinnable, due to this or other bugs. Given the much smaller sample of 1,000, we do not know if the results reported by \citet{shootme_klondike} are affected.

\citet{wolter_canfield}, provided the best previous analysis of \gamename{Canfield}  giving statistics over 50,000 tests of 35,606 solved, 13,730 proved unsolvable, and 664 indeterminate.  
The code for his solver is available \cite{wolter-code}. As published, 
the code gives different results because it implements a rule that   only entire columns or the bottom card alone can be moved. This is different from the game rules that  \citet{wolter_canfield_rules} gives himself, where partial built piles may be moved instead of just whole columns.\footnote{Curiously, the implemented rule is precisely that given by \citet{parlett}, but we do not know whether this was intentional. We cannot check this since Jan Wolter died on 1 January, 2015. We are happy to have this opportunity to pay tribute to him both for his excellent work on solitaire solving programs, and for his openness in making his code publicly available, allowing us to build on his work.} 
A minor change to the published code restores the game to \citeS{wolter_canfield_rules} rules and after doing this we obtained \textit{identical} results to those \citet{wolter_canfield} reported.\footnote{Perhaps Wolter corrected the code but never pushed to Google code, or alternatively computed the results before some later code change.}  
After making this change, we compared results between Wolter's solver and ours for \gamename{Canfield}. There remained discrepancies which revealed Solvitaire to have both an unintended rule and a separate bug. When these were corrected we still found a small number of different results, which led to the discovery of two obscure bugs in Wolter's code, arising from an incorrect dominance rule.  This was a dominance which forced moves to the foundation be made when an appropriate card was in the last two cards in the stock, because playing these cards could (apparently) never prevent another card being played. Unfortunately, if the number of cards in waste is not a multiple of the number of cards played from stock (typically 3), then immediately playing the last card in stock prevents access to the card at the top of the waste pile, and possibly others. For much more subtle reasons, it is not safe to allow the penultimate card in the stock to be played.  
 While rare, we did see examples of random instances where Wolter's code incorrectly reported winnable instances as impossible.  For example, in one game in which the base card was 5$\diamondsuit$, the stock started 3$\clubsuit$ 6$\clubsuit$ 6$\diamondsuit$ and ended K$\clubsuit$ 7$\diamondsuit$  5$\heartsuit$ Q$\spadesuit$.
There was no solution if the 5$\heartsuit$ (the second last card in the stock) was played immediately.  To win, the player has to wait until the 6$\diamondsuit$ and 6$\clubsuit$ are both played consecutively. Having delayed the play of 5$\heartsuit$ allows it to be played now, uncovering the 7$\diamondsuit$ which can be put on the 6$\diamondsuit$.  The situation is the curious one that if we have already played 5$\heartsuit$ earlier, then after 6$\diamondsuit$ we are able to play either the 7$\diamondsuit$ or the 6$\clubsuit$ but not both.  
We believe a very weak version of Wolter's dominance is correct: when the last card of stock (not the last two)  meets the conditions of   \Cref{sec:safemoves} \textit{and} the stock is currently at a multiple of the draw size, the last card can be moved to foundation.  We did not implement this in our code.

To correct these two bugs we rewrote Wolter's code to allow dominance moves only for the last card in stock and only when the number of cards in the waste pile is a multiple of the number of cards played from stock. With these corrections our code does not disagree on any of 50,000 instances we tested.  Using this corrected code with the parameters \citet{wolter_canfield} previously used, we obtained 35,605 solved, 13,671 proved unsolvable, and 724 indeterminate instances: if those results had been reported, Table~\ref{tab:solpercs-comparison} would have a confidence interval of 71.929\% $\pm$ 1.118\% for Wolter's results.
This confidence interval did not, at that time, overlap with our results, leading us to investigate closely the pseudorandom generators for both programs. We found flaws in both generators, with Wolter's code producing identical instances on repeated seeds, e.g. the same results for seeds 12 and 1212, and ours a slightly biased sample. We corrected our generator appropriately and our results are now consistent with Wolter's, as shown in Table~\ref{tab:solpercs-comparison}.  

The general point we make in this section is not a criticism of other programmers, but to emphasise the ease with which apparently correct optimisations can in fact be wrong, and to show the difficulty that can arise in locating the errors.   Additionally, it shows the power of Solvitaire in being able to run such extensive comparisons with other solvers that it is able to find very rare inconsistencies, and the benefit to other games of  fixing bugs found while investigating one game.

\section{Experimental Methods}

\subsection{Statistics}
\label{sec:stats}

Each random instance is necessarily either solvable or unsolvable, and therefore the true picture for any given game is it behaves as a binomial with probability $p$ of success. As discussed in \Cref{sec:stronger-weaker}, in some cases we used winnability facts from stronger or weaker games where this was guaranteed correct, saving considerable time.  We used the following consistent protocol for measuring a confidence interval on the estimate of winnability percentage. From a sample, if we know the number of winnable and unwinnable instances, we  calculate a 95\% confidence interval for the true value of $p$ using Wilson's method \cite{agresti1998approximate}.  When some instances' winnability are unknown, e.g. due to timeouts, we form the most conservative possible interval by calculating the interval both on the assumption that every unknown instance is unwinnable and on the assumption that every unknown instance is winnable.   
We then report the range from the lower bound of the first interval to the upper bound of the second. While it would be nice to be less conservative and get a smaller interval, no other totally general approach seems valid: for example, in \gamename{Spider} it is very likely that almost all unresolved instances are winnable, while in \gamename{Klondike} most long-running instances turn out to be unwinnable. We normally report percentage winnability to 3 decimal places, but give more places where winnability is very close to either 0 or 100\%. 
   We  use the most conservative possible rounding: given the number of digits we are reporting, we round the lower bound down and the upper bound up. Given the range calculated, we report it from the centre, plus or minus half the range (with the centre chosen arbitrarily from the two choices where the range is odd in the last digit). For calculating equivalent intervals for comparison with previous work, in most cases we could deduce the raw numbers of solved, unsolvable and indeterminate cases from past publications, and calculate the confidence interval that would result from the same protocol.    While a confidence interval we compare against may not be the same as that reported in a previous paper, our comparisons with previous results are on a like-for-like basis without being dependent on varying methodologies for estimating the range of winnability used by different authors. 

For computing  necessity of worrying back in \Cref{tab-worryback}, we used the same set of instances in both games to reduce variance. For calculating bounds we used a similar protocol to the above,   but had to carefully allow for cases where the result for an instance was unknown either with or without worrying back. The highest possible necessity of worrying back would be if all such instances were unwinnable without worrying back but winnable with it.   The lowest possible necessity would be if all unknown instances gave the same result with or without worrying back, and all instances where the result without worrying back is unknown are winnable.  In  \Cref{tab-worryback} the upper bound of necessity is the high end of the 95\% confidence interval in the first case and the lower bound is the lower end of the interval in the second case.

Statistics were calculated using R \cite{Rmanual}.

\subsection{Experimental Setup}

Monte Carlo methods using pseudo-random generation were used to create instances of each game. We  used the Mersenne twister generator \cite{Matsumoto:1998:MTE:272991.272995} mt19337 provided by the C++ standard library to generate a stream of pseudorandom numbers. The stream of numbers passed all tests for randomness in the 
 Dieharder test suite, v3.31.1 \cite{dieharder}, simulating the way it was used in our code with the initial seed incremented after every 52 random numbers. 
 We wrote our own code to create instances from the stream of numbers: this generator is portable so should produce identical results for the same seed, and is included in our code for Solvitaire. 
In running experiments, a critical point is that runs which were unresolved are included in our statistics. In many cases we re-ran failed seeds with larger computational resources, but where we could never resolve the instance, they are included in our data  as unknown. It would be improper to ignore them and rerun with a new seed as hard instances can have a different likelihood of being winnable to a new random seed.  
Having decided on a sample size for an experiment we used a consecutive sequence of seeds for that experiment.   Seeds for each instance are recorded in our data. As well as winnability, we recorded many other features of search such as run-time, memory usage, cache usage, and search depth. We do not report those statistics in detail but they are available in full in our data files: Table~\ref{tab:times} gives an overview by game of run-time and nodes searched.  

Experiments mostly used the Cirrus UK National Tier-2 HPC Service at EPCC (see acknowledgements). 
Additionally, a small number of our results presented here were obtained on  local compute-servers at the University of St Andrews. 
In selecting experimental parameters such as sample size, number of cores used per machine, timeout limits, and cache sizes, we made choices intended to optimise the computing resources and time available. For example, for some games it was critical to run with very large amounts of RAM, reducing the number that could be run in parallel on one machine. 
In some cases, we accepted a small number of timeouts in order to get a very large sample size (e.g. \gamename{American Canister}). In others where there were many timeouts, we focussed on a smaller sample size but very long runtimes to minimise the number of unknowns (e.g. \gamename{Gaps One Deal}). All results were obtained using Solvitaire, but to save CPU time we sometimes reused results for one game for a related game, as described in \Cref{sec:relations}. 
During experiments, minor changes were made to Solvitaire: our data files indicate the version used for each experiment.  As a consistency check, we compared results of the current version (0.10.1) against our reported results by testing 1000 instances (or 100 in two very hard games). In all cases where both versions completed, all features of search including precise numbers of nodes were identical.

\section{Evaluation of Solvitaire}
\label{sec:evaluation}

We have achieved considerable success using Solvitaire, as we have reported throughout this paper. In this section we reflect on the strengths and weaknesses of our program, Solvitaire, and how future work can build on our success.

A significant feature of our work is that most predecessors have written a special program for each main game, while our single program Solvitaire  can solve games from a simple textual description.  This gives two significant advantages over previous work. First, the uniform approach enables us to implement advanced AI techniques just once but apply them to many games. Second, we can gain improved confidence in correctness through bugfixes from one game automatically applying to all others.  
 Therefore our work has significant value even in games where previous studies have been done.

 We regard it as remarkable that we have been able to obtain so many new and improved results using a general purpose patience playing program. Normally, we would expect a general purpose program to be significantly outperformed by specially written programs.  In some cases we have obtained very much improved results over previous work, but this may be due to the availability of significant computer time on modern CPUs rather than an improved solver.
 Nevertheless, it remains clear that Solvitaire is an outstanding solver in most games we have evaluated it on. 
 
In Appendix \ref{app:comparative} we  compare run times of Solvitaire with the best previous specialised solvers for \gamename{Canfield}, \gamename{Klondike}, and \gamename{FreeCell}.  For \gamename{Canfield}, we found that Solvitaire did not perform quite as well as \citeS{wolter-code} solver.
For \gamename{Klondike}, we found Solvitaire performed slightly better than \citet{shootme_klondike}'s solver.   For \gamename{FreeCell}, Solvitaire was much worse than \citeS{fish_fcsolve}'s solver when streamliners were not used, but with the use of our  smart streamliner did in fact perform slightly better than Fish's, though Fish's solver remains better on the rare unwinnable instances. 
 These comparisons should not be taken as a proper  scientific comparison of our solvers with competing ones, since for example other solvers may have options or settings which would improve their performance.    It is clear that performance of Solvitaire approaches the performance of state-of-the-art solvers while being much more general.
 One limitation in our design of Solvitaire is that we do not attempt to find shortest solutions to instances, or even solutions with some maximum number of moves.

 While giving us many advantages, our configurable rule set has some limitations. For example, it does not allow for games with a fixed number of redeals of stock.  We also do not allow for some key rules such as pairing,  eliminating games like \gamename{Doublets} and many others.
These limitations were conscious in the sense that in the expressivity/speed tradeoff, we prioritised getting good performance on games we could express rather than total generality.   

There are some games where we could not improve on previous results, as seen in \Cref{tab:solpercs-comparison}. 
Some examples of this are very near to 100\%  winnability, such as \gamename{FreeCell}, \gamename{Spider}, and \gamename{Accordion}.  We may have been less effective for these games due to our entirely general approach of prioritising proving whether a game was winnable or not, rather than fastest possible finding of a winning sequence when it existed.    A particularly interesting example where another solver outperforms Solvitaire is \gamename{Worm Hole}.   We invented this game to show the flexibility of our ruleset and got reasonable results, but since doing so \citet{masten_wormhole} has reported a dominance we were not aware of which allows for improved performance.  This again shows the value of dominances in patience solving, as we discussed in \Cref{sec:dominance}, and how much more remains to be done in this area.
 
 One important weakness we have identified is that in some games, many instances are unwinnable but Solvitaire is unable to prove them so. A particularly clear example of this is seen in \Cref{tab-variants}, page~\pageref{tab-variants}.   
  We believe the problem is the combination of a very liberal rule for moving cards (in this case any suit) with a very strict restriction in another area (in this case unfillable spaces). The liberal rule makes the search space very large, while the restriction means that the location of a small number of cards can make the game unwinnable.  In this game, imagine a King covering a Queen of the same suit in the tableau.  The Queen can never be reached because the King cannot be moved to a space, while the King cannot be built to the foundation because the Queen is not available.  Yet there can be literally billions of potential paths that Solvitaire might have to explore, leading to timing out.  
 We have seen similar problems in  other games where a game with many possible moves can be unwinnable for small local reasons, and thrashing occurs.  While we did not report results here, we saw this with the game `\gamename{Alina}' \cite{250sol_rules}, 
 where Solvitaire has never proved one to be unwinnable even though our human examination shows  that many cannot be won. 
 It should be possible to create solvers which combine the exploratory search strength of Solvitaire while also adding more reasoning ability to exclude possibilities.  For example, one might use an approach for detecting inconsistency in planning problems such as suggested by \citet{DBLP:conf/socs/BackstromJS13}.   Some initial work shows that constraint solvers can be used to prove instances unwinnable in \gamename{Klondike}, but much remains to be done \cite{DBLP:conf/cp/0001GNUW25}.
 
It is remarkable that Solvitaire has been so successful on so many games despite this weakness.

\section{Conclusions} 
 
\label{sec:conclusion}

We have shown that a single depth-first search based solver, Solvitaire, is able to produce state-of-the-art results across  a very wide variety of patience games.  We achieved this by combining a variety of general AI search techniques. In doing so, we have obtained many entirely new results across a wide variety of games.  We have also greatly improved the state-of-the-art winnability estimates on many games 
including some of the most famous games such as \gamename{Klondike} (often just called `Solitaire') and 
\gamename{Canfield}.  
In a pleasing callback to their origin, we have now used Monte Carlo methods to answer the question that  caused Stanislaw Ulam to invent Monte Carlo methods.

Despite the level of interest we described in Section~\ref{history}, we are surprised that this study is the first of its kind, i.e. an academic study of the winnability of many different patience games.   Previous studies within academia have tended to focus one game, possibly with some variants.  There have been more wide ranging studies done outside the academic literature, for example by \citet{masten_winrates}, \citet{wolter_analysis}, and others.  Showing that general AI methods can be applied across patience games, we hope very much that other researchers will build on what we have done and no doubt greatly improve on it.

The importance of dominances for patience solving is very high, but a number of significant problems remain with their application, which should be addressed in future work.   
First, one has to discover dominances in the first place.   They can be difficult to notice and are often not well publicised in the literature.   There is also the overhead of implementation as they can be quite specialised: for example we have not implemented possible dominances to forbid making pointless moves of a card on the tableau which clears a space that cannot be usefully used. 
Apart from discovering and implementing dominances  in the first place, ensuring their correctness can be very difficult.   
There is a very close link between streamliners and dominances, since a streamliner is just an incorrect dominance, so unifying their treatment would be interesting. 
Ideally we would like to be able to apply dominances and 
streamliners automatically, correctly, and generally.  Achieving this remains a key challenge for future work in patience solving.

 While we believe we have made a significant contribution to the study of games that have occupied humans for uncounted hours, much remains to be done.   
 Without doubt, the most interesting question we leave open is one we have not attempted to tackle at all.  In games with hidden cards like \gamename{Klondike}, what is the true probability of winning from a starting position?  We have always solved the `thoughtful version'.  When faced with a game of the classic Solitaire, \gamename{Klondike}, with no peeking on physical cards or electronic undo button, what is the best attainable probability of winning, and how does one obtain this?  
 For many games, this remains a very hard problem, with an answer that is not currently known for \gamename{Klondike} even within a factor of two. While the thoughtful winnability gives an upper bound, it gives us no direct information about a lower bound.

 \section*{Data and Code Availability}
Full experimental results reported in this paper  are available at figshare.com
with DOI \href{https://doi.org/10.6084/m9.figshare.8311070}{10.6084/m9.figshare.8311070} \cite{figshare-data}. This dataset includes
all runs used to report data in this paper, together with other material such as details
of testing and analysis scripts used to compute winnability estimates reported here. 
The code for Solvitaire is open-source under the GNU GPL Version 2 licence. The code used for this version of the paper is available in Zenodo at identifier doi:10.5281/zenodo.3529524 \cite{solvitaire-v0.10.2}. 
Development history of Solvitaire is also available on Github at URL \url{https://github.com/thecharlesblake/Solvitaire}. 

 \section*{Author Contributions} 
 IPG proposed and supervised the project. 
 CB and IPG jointly made high-level design decisions.  
 CB made all low-level design decisions, implemented Solvitaire, and named it. 
 CB and IPG debugged Solvitaire, and ran exploratory experiments.  
IPG ran the full experiments reported here and analysed them.
IPG constructed the proof of the Theorems.
IPG drafted the paper, with CB and IPG  revising it.
 
\section*{Acknowledgements}
This work was in part supported by EPSRC (EP/P015638/1). 
This work used the Cirrus UK National Tier-2 HPC Service at EPCC (http://www.cirrus.ac.uk) funded by the University of Edinburgh and EPSRC (EP/P020267/1).

We thank reviewers of earlier versions of this paper for valuable suggestions for improvement.  
We thank others who have helped us in our work on patience, including Matt Birrell, Dawn Black, Laura Brewis, Arthur W. Cabral, Gal Cohensius, Nguyen Dang, Joan Espasa Arxer, Shlomi Fish, Jordina Franc\`es de Mas, Alan Frisch, Patrik Haslum, Chris Jefferson, Michael Keller, Donald Knuth, Dana Mackenzie, Mark Masten, Ian Miguel, Peter Nightingale, Theodore Pringle,  Bill Roscoe, Andr\'as Salamon, Felix Ulrich-Oltean, Judith Underwood, Jack Waller, and (posthumously) Jan Wolter. 

Ian Gent thanks his mother Margaret Gent (1923-2021) for her patience in teaching him love for the game of patience. 

\printbibliography

@misc{wiki:SimpleMonteCarlo,
    author  = {{Wikipedia Contributors}},
    title       = {{Simple Monte Carlo}},
    month        = {10},
    day          = {24},
    url          = {https://web.archive.org/web/20251024103921/https://en.wikipedia.org/wiki/Monte_Carlo_method#Simple_Monte_Carlo},
    year         = {2025},
    howpublished = {In \textit{Wikipedia}},
      addendum={Archive of 24 Oct 2025}
}

@InProceedings{10.1007/978-3-540-75538-8_7,
author="Coulom, R{\'e}mi",
editor="van den Herik, H. Jaap
and Ciancarini, Paolo
and Donkers, H. H. L. M.",
title="Efficient Selectivity and Backup Operators in Monte-Carlo Tree Search",
booktitle="Computers and Games",
year="2006",
publisher="Springer Berlin Heidelberg",
address="Berlin, Heidelberg",
pages="72--83",
isbn="978-3-540-75538-8"
}

@inproceedings{DBLP:conf/cp/0001GNUW25,
  author       = {Nguyen Dang and
                  Ian P. Gent and
                  Peter Nightingale and
                  Felix Ulrich{-}Oltean and
                  Jack Waller},
  editor       = {Maria Garcia de la Banda},
  title        = {Constraint Models for Klondike},
  booktitle    = {31st International Conference on Principles and Practice of Constraint
                  Programming, {CP} 2025, August 10-15, 2025, Glasgow, Scotland},
  series       = {LIPIcs},
  volume       = {340},
  pages        = {9:1--9:20},
  publisher    = {Schloss Dagstuhl - Leibniz-Zentrum f{\"{u}}r Informatik},
  year         = {2025},
  url          = {https://doi.org/10.4230/LIPIcs.CP.2025.9},
  doi          = {10.4230/LIPICS.CP.2025.9},
  bibsource    = {dblp computer science bibliography, https://dblp.org}
}

@article{KORF198535,
title = {Macro-operators: A weak method for learning},
journal = {Artificial Intelligence},
volume = {26},
number = {1},
pages = {35-77},
year = {1985},
issn = {0004-3702},
doi = {https://doi.org/10.1016/0004-3702(85)90012-8},
url = {https://www.sciencedirect.com/science/article/pii/0004370285900128},
author = {Richard E. Korf}
}

@article{JUNGHANNS2001219,
title = {Sokoban: Enhancing general single-agent search methods using domain knowledge},
journal = {Artificial Intelligence},
volume = {129},
number = {1},
pages = {219-251},
year = {2001},
issn = {0004-3702},
doi = {https://doi.org/10.1016/S0004-3702(01)00109-6},
url = {https://www.sciencedirect.com/science/article/pii/S0004370201001096},
author = {Andreas Junghanns and Jonathan Schaeffer}
}

@article{10.5555/1622519.1622534,
author = {Botea, Adi and Enzenberger, Markus and M\"{u}ller, Martin and Schaeffer, Jonathan},
title = {Macro-FF: improving AI planning with automatically learned macro-operators},
year = {2005},
issue_date = {July 2005},
publisher = {AI Access Foundation},
address = {El Segundo, CA, USA},
volume = {24},
number = {1},
issn = {1076-9757},
journal = {J. Artif. Int. Res.},
month = oct,
pages = {581–621},
numpages = {41}
}

@misc{howe-crowdsourcing,
author = {Jeff Howe}, 
title = {The Rise Of Crowdsourcing.},
howpublished = {Wired},
note={June 2006},
url={https://web.archive.org/web/20151028232825/https://www.wired.com/2006/06/crowds/},
year = 2006, 
month=6}

@inproceedings{10.5555/647736.735461,
    author = {Valmari, Antti},
    title = {Stubborn sets for reduced state space generation},
    year = {1991},
    isbn = {3540538631},
    publisher = {Springer-Verlag},
    address = {Berlin, Heidelberg},
    booktitle = {Proceedings of the 10th International Conference on Applications and Theory of Petri Nets: Advances in Petri Nets 1990},
    pages = {491–515},
    numpages = {25}
    }

@inproceedings{10.5555/3038546.3038581,
author = {Wehrle, Martin and Helmert, Malte},
title = {About partial order reduction in planning and computer aided verification},
year = {2012},
publisher = {AAAI Press},
booktitle = {Proceedings of the Twenty-Second International Conference on International Conference on Automated Planning and Scheduling},
pages = {297–305},
numpages = {9},
location = {Atibaia, S\~{a}o Paulo, Brazil},
series = {ICAPS'12}
}

@article{10.5555/1622394.1622404,
author = {Hoffmann, J\"{o}rg and Nebel, Bernhard},
title = {The FF planning system: fast plan generation through heuristic search},
year = {2001},
issue_date = {January 2001},
publisher = {AI Access Foundation},
address = {El Segundo, CA, USA},
volume = {14},
number = {1},
issn = {1076-9757},
journal = {J. Artif. Int. Res.},
month = {5},
pages = {253–302},
numpages = {50}
}

@inproceedings{DBLP:conf/socs/BackstromJS13,
  author       = {Christer B{\"{a}}ckstr{\"{o}}m and
                  Peter Jonsson and
                  Simon St{\aa}hlberg},
  editor       = {Malte Helmert and
                  Gabriele R{\"{o}}ger},
  title        = {Fast Detection of Unsolvable Planning Instances Using Local Consistency},
  booktitle    = {Proceedings of the Sixth Annual Symposium on Combinatorial Search,
                  {SOCS} 2013, Leavenworth, Washington, USA, July 11-13, 2013},
  pages        = {29--37},
  publisher    = {{AAAI} Press},
  year         = {2013},
  url          = {https://doi.org/10.1609/socs.v4i1.18294},
  doi          = {10.1609/SOCS.V4I1.18294},
  bibsource    = {dblp computer science bibliography, https://dblp.org}
}

@misc{figshare-data,
author = "Ian P. Gent and Charles Blake",
title = "{Patience Experimental Results}",
year = "2024",
month = "8",
url = "https://figshare.com/articles/Patience_Experimental_Results/8311070",
doi = "10.6084/m9.figshare.8311070"
}

@book{cavendish,
author = "Cavendish",
title = "Patience Games",
year = 1890,
publisher = "De La Rue"}

@book{parlett,
author = "David Parlett",
title = "The Penguin Book of Patience",
year = 1980,
publisher = "Penguin"}

@Manual{Rmanual,
    title = {R: A Language and Environment for Statistical Computing},
    author = {{R Core Team}},
    organization = {R Foundation for Statistical Computing},
    address = {Vienna, Austria},
    year = {2016},
    url = {https://www.R-project.org/},
  }

@misc{DBLP:journals/corr/abs-0907-1955,
  author    = {Matthew C. Clarke},
  title     = {On the Chances of Completing the Game of ``{Perpetual Motion}".},
  howpublished   = {arXiv.cs},
  year      = {2009},
  doi = {10.48550/ARXIV.0907.1955},
  archivePrefix = {arXiv},
  eprint    = {0907.1955},
  bibsource = {dblp computer science bibliography, https://dblp.org}
}

@techreport{crockford2006application,
   author={Crockford, Douglas},
   TITLE = {The application/json media type for {JavasSript} {Object} {Notation} ({JSON}).},
   HOWPUBLISHED = {Internet Requests for Comments},
   TYPE="{RFC}",
   NUMBER=4627,
   YEAR = {2006},
   MONTH = {7},
   ISSN = {2070-1721},
   PUBLISHER = "{RFC Editor}",
   INSTITUTION = "{RFC Editor}",
   URL={https://www.rfc-editor.org/rfc/rfc4627.txt}
   }

@misc{DBLP:journals/corr/Roscoe16,
  author    = {A. W. Roscoe},
  title     = {Card games as pointer structures: case studies in mobile {CSP} modelling.},
  howpublished  = {arXiv.cs},
  doi = {10.48550/ARXIV.1611.08418},
  year      = {2016},
  month = {11},
  day = {25},
  archivePrefix = {arXiv},
  eprint    = {1611.08418},
  bibsource = {dblp computer science bibliography, https://dblp.org}
}

@techreport{accordion-stanford,
 author = {Ross, Kenneth A. and Knuth, Donald E.},
 title = {A Programming and Problem Solving Seminar.},
 year = {1989},
 number = {STAN-CS-89-1269},
 url = {https://web.archive.org/web/20180409232321/http://i.stanford.edu/pub/cstr/reports/cs/tr/89/1269/CS-TR-89-1269.pdf},
 institution = {Stanford University},
 address = {Stanford, CA, USA}
}

@article{Lipovetzky_Geffner_2017, 
title={Best-First Width Search: Exploration and Exploitation in Classical Planning}, 
volume={31}, 
url={https://ojs.aaai.org/index.php/AAAI/article/view/11027}, 
DOI={10.1609/aaai.v31i1.11027}, 
abstractNote={ &lt;p&gt; It has been shown recently that the performance of greedy best-first search (GBFS) for computing plans that are not necessarily optimal can be improved by adding forms of exploration when reaching heuristic plateaus: from random walks to local GBFS searches. In this work, we address this problem but using structural exploration methods resulting from the ideas of width-based search. Width-based methodsseek novel states, are not goal oriented, and their power has been shown recently in the Atari and GVG-AI video-games. We show first that width-based exploration in GBFS is more effective than GBFS with local GBFS search (GBFS-LS), and then proceed to formulate a simple and general computational framework where standard goal-oriented search (exploitation) and width-based search (structural exploration) are combined to yield a search scheme, best-first width search, that is better than both and which results in classical planning algorithms that outperform the state-of-the-art planners. &lt;/p&gt; }, 
number={1}, 
journal={Proceedings of the AAAI Conference on Artificial Intelligence}, 
author={Lipovetzky, Nir and Geffner, Hector}, 
year={2017}, month={2} }

@inproceedings{DBLP:conf/socs/AkagiKF10,
  author       = {Yuima Akagi and
                  Akihiro Kishimoto and
                  Alex Fukunaga},
  editor       = {Ariel Felner and
                  Nathan R. Sturtevant},
  title        = {On Transposition Tables for Single-Agent Search and Planning: Summary
                  of Results},
  booktitle    = {Proceedings of the Third Annual Symposium on Combinatorial Search,
                  {SOCS} 2010, Stone Mountain, Atlanta, Georgia, USA, July 8-10, 2010},
  pages        = {2--9},
  publisher    = {{AAAI} Press},
  year         = {2010},
  url          = {https://doi.org/10.1609/socs.v1i1.18164},
  doi          = {10.1609/SOCS.V1I1.18164},
  bibsource    = {dblp computer science bibliography, https://dblp.org}
}

@inproceedings{DBLP:conf/cp/AkgunGJMN18,
  author       = {{\"{O}}zg{\"{u}}r Akg{\"{u}}n and
                  Ian P. Gent and
                  Christopher Jefferson and
                  Ian Miguel and
                  Peter Nightingale},
  editor       = {John N. Hooker},
  title        = {Metamorphic Testing of Constraint Solvers},
  booktitle    = {Principles and Practice of Constraint Programming - 24th International
                  Conference, {CP} 2018, Lille, France, August 27-31, 2018, Proceedings},
  series       = {Lecture Notes in Computer Science},
  volume       = {11008},
  pages        = {727--736},
  publisher    = {Springer},
  year         = {2018},
  url          = {https://doi.org/10.1007/978-3-319-98334-9_46},
  doi          = {10.1007/978-3-319-98334-9_46},
  bibsource    = {dblp computer science bibliography, https://dblp.org}
}

@inproceedings{trailingVsCopying,
doi={10.5555/341176.341217},
author = {Schulte, Christian},
title = {Comparing trailing and copying for constraint programming},
year = {1999},
isbn = {0262541041},
publisher = {Massachusetts Institute of Technology},
address = {USA},
booktitle = {Proceedings of the 1999 International Conference on Logic Programming},
pages = {275–289},
numpages = {15},
location = {Las Cruces, New Mexico, USA}
}

@inproceedings{fcnn,
  title={``{Freecell}" neural network heuristics.},
  author={Dunphy, A and Heywood, MI},
    booktitle={Proceedings of the International Joint Conference on Neural Networks, 2003},
  volume={3},
  pages={2288--2293},
  year={2003},
  organization={IEEE},
  doi={10.1109/IJCNN.2003.1223768}
}

@incollection{gent2006symmetry,
  title={Symmetry in constraint programming.},
  author={Gent, Ian P and Petrie, Karen E and Puget, Jean-Fran{\c{c}}ois},
  booktitle={Foundations of Artificial Intelligence},
  volume={2},
  pages={329--376},
  year={2006},
  publisher={Elsevier},
  doi = {10.1016/S1574-6526(06)80014-3}
}

@inproceedings{Greenblatt_transposition,
author = {Greenblatt, Richard D. and Eastlake, Donald E. and Crocker, Stephen D.},
title = {The {Greenblatt} Chess Program.},
year = {1967},
isbn = {9781450378963},
publisher = {Association for Computing Machinery},
address = {New York, NY, USA},
doi = {10.1145/1465611.1465715},
booktitle = {Proceedings of the November 14-16, 1967, Fall Joint Computer Conference},
pages = {801–810},
numpages = {10},
location = {Anaheim, California},
series = {AFIPS '67 (Fall)}
}

@InProceedings{streamliner-gomes,
author="Gomes, Carla
and Sellmann, Meinolf",
editor="Wallace, Mark",
title="Streamlined Constraint Reasoning.",
booktitle="Principles and Practice of Constraint Programming -- CP 2004",
year="2004",
publisher="Springer Berlin Heidelberg",
address="Berlin, Heidelberg",
pages="274--289",
doi="10.1007/978-3-540-30201-8_22",
isbn="978-3-540-30201-8"
}

@inproceedings{helmstetter2004searching,
author="Helmstetter, B.
and Cazenave, T.",
editor="Van Den Herik, H. Jaap
and Iida, Hiroyuki
and Heinz, Ernst A.",
title="Searching with Analysis of Dependencies in a Solitaire Card Game.",
bookTitle="Advances in Computer Games: Many Games, Many Challenges",
year="2004",
publisher="Springer US",
address="Boston, MA",
pages="343--360",
isbn="978-0-387-35706-5",
doi="10.1007/978-0-387-35706-5_22"
}

@InProceedings{smith-caching,
author="Smith, Barbara M.",
editor="van Beek, Peter",
title="Caching Search States in Permutation Problems.",
booktitle="Principles and Practice of Constraint Programming - CP 2005",
year="2005",
publisher="Springer Berlin Heidelberg",
address="Berlin, Heidelberg",
pages="637--651",
isbn="978-3-540-32050-0",
doi="10.1007/11564751_47"
}

@inproceedings{wetter2015automatically,
  title={Automatically Generating Streamlined Constraint Models with {ESSENCE} and {CONJURE}.},
  author={Wetter, James and Akg{\"u}n, {\"O}zg{\"u}r and Miguel, Ian},
 booktitle="Principles and Practice of Constraint Programming",
 editor="Pesant, Gilles",
  pages={480--496},
  year={2015},
publisher="Springer International Publishing",
doi="10.1007/978-3-319-23219-5_34"
}

@inproceedings{DBLP:conf/socs/BurchH11,
  author       = {Neil Burch and
                  Robert C. Holte},
  editor       = {Daniel Borrajo and
                  Maxim Likhachev and
                  Carlos Linares L{\'{o}}pez},
  title        = {Automatic Move Pruning in General Single-Player Games},
  booktitle    = {Proceedings of the Fourth Annual Symposium on Combinatorial Search,
                  {SOCS} 2011, Castell de Cardona, Barcelona, Spain, July 15.16, 2011},
  pages        = {31--38},
  publisher    = {{AAAI} Press},
  year         = {2011},
  url          = {https://doi.org/10.1609/socs.v2i1.18187},
  doi          = {10.1609/SOCS.V2I1.18187},
  bibsource    = {dblp computer science bibliography, https://dblp.org}
}

@inproceedings{klondikemcts,
  title={Lower Bounding {Klondike Solitaire} with {Monte-Carlo} Planning.},
  author={Bjarnason, Ronald and Fern, Alan and Tadepalli, Prasad},
  booktitle={ICAPS'09: Proceedings of the Nineteenth International Conference on International Conference on Automated Planning and Scheduling},
  pages={26--33},
  year={2009},
  url={https://dl.acm.org/doi/10.5555/3037223.3037228}
}

@article{gent2007search,
  title={Search in the patience game '{Black Hole}'.},
  author={Gent, Ian P and Jefferson, Chris and Kelsey, Tom and Lynce, Ines and Miguel, Ian and Nightingale, Peter and Smith, Barbara M and Tarim, S Armagan},
  journal={AI Communications},
  volume={20},
  number={3},
  pages={211--226},
  year={2007},
  publisher={IOS Press},
  URL={https://dl.acm.org/doi/10.5555/1365527.1365533}
}

@article{patience-history,
author = {Ross, A.S.C. and Healey, F.G.},
title = "Patience {Napol\'eon}.",
year = 1963,
month = 9,
pages = "137-190",
journal = "Proceedings of the Leeds Philosophical and Literary Society (Literary and Historical Section)",
volume = "10"}

@inproceedings{paul2016optimal,
  title={Optimal solitaire game solutions using A* search and deadlock analysis.},
  author={Paul, Gerald and Helmert, Malte},
  booktitle={Ninth Annual Symposium on Combinatorial Search},
  year={2016},
  ISBN={978-1-57735-769-8},
  pages={135--136},
  URL={https://web.archive.org/web/20220805135304/https://ojs.aaai.org/index.php/SOCS/article/view/18405}
}

@inproceedings{rollout2,
  title={Solitaire: Man versus machine.},
  author={Yan, Xiang and Diaconis, Persi and Rusmevichientong, Paat and Roy, Benjamin V},
  booktitle={Advances in Neural Information Processing Systems},
  pages={1553--1560},
  URL={https://dl.acm.org/doi/10.5555/2976040.2976235},
  year={2005}
}

@article{10.1287/ijoc.14.4.295.2828,
title = {Logic, Optimization, and Constraint Programming},
author = {John N. Hooker},
journal = {"INFORMS" Journal on Computing},
year = {2002},
issue_date = {November 2002},
publisher = {INFORMS},
address = {Linthicum, MD, USA},
volume = {14},
number = {4},
issn = {1526-5528},
doi={10.1287/ijoc.14.4.295.2828},
month = {11},
numpages = {27}
}

@article{DBLP:journals/cacm/DavisLL62,
  author    = {Martin Davis and
               George Logemann and
               Donald W. Loveland},
  title     = {A machine program for theorem-proving},
  journal   = {Commun. {ACM}},
  volume    = {5},
  number    = {7},
  pages     = {394--397},
  year      = {1962},
  url       = {https://doi.org/10.1145/368273.368557},
  doi       = {10.1145/368273.368557},
  bibsource = {dblp computer science bibliography, https://dblp.org}
}

@article{agresti1998approximate,
  title={Approximate is better than `exact' for interval estimation of binomial proportions.},
  author={Agresti, Alan and Coull, Brent A},
  journal={The American Statistician},
  volume={52},
  number={2},
  pages={119--126},
  year={1998},
  publisher={Taylor \& Francis},
  doi          = {10.2307/2685469}
}

@article{rollout1,
  title={Searching Solitaire in Real Time.},
  author={Bjarnason, Ronald and Tadepalli, Prasad and Fern, Alan},
  journal={ICGA Journal},
  volume={30},
  number={3},
  pages={131--142},
  year={2007},
  doi={10.3233/ICG-2007-30302}
}

@Article{Chu2015,
author="Chu, Geoffrey
and Stuckey, Peter J.",
title="Dominance breaking constraints.",
journal="Constraints",
year="2015",
month="4",
day="01",
volume="20",
number="2",
pages="155--182",
issn="1572-9354",
doi="10.1007/s10601-014-9173-7"
}

@article{ulam-solitaire,
title={{S}tan {U}lam, {J}ohn von {N}eumann, and the {M}onte {C}arlo method.},
author = {Roger Eckhardt},
year = 1987,
journal = {Los Alamos Science},
volume= {15},
pages = {131--141},
doi= {10.2172/1054744},
url = {https://web.archive.org/web/20161129030550/https://permalink.lanl.gov/object/tr?what=info:lanl-repo/lareport/LA-UR-88-9068}
}

@article{genetic1,
  title={Evolutionary design of {FreeCell} solvers.},
  author={Elyasaf, Achiya and Hauptman, Ami and Sipper, Moshe},
  journal={IEEE Transactions on Computational Intelligence and AI in Games},
  volume={4},
  number={4},
  pages={270--281},
  year={2012},
  publisher={IEEE},
  doi={https://doi.org/10.1109/TCIAIG.2012.2210423}
}

@article{doi:10.1080/0025570X.1981.11976927,
author = {T. A. Jenkyns and E. R. Muller},
title = {A Probabilistic Analysis of Clock Solitaire.},
journal = {Mathematics Magazine},
volume = {54},
number = {4},
pages = {202-208},
year  = {1981},
publisher = {Taylor & Francis},
doi = {10.1080/0025570X.1981.11976927}, 
eprint = {https://doi.org/10.1080/0025570X.1981.11976927}
}

@article{Matsumoto:1998:MTE:272991.272995,
 author = {Matsumoto, Makoto and Nishimura, Takuji},
 title = {Mersenne Twister: A 623-dimensionally Equidistributed Uniform Pseudo-random Number Generator.},
 journal = {ACM Trans. Model. Comput. Simul.},
 issue_date = {Jan. 1998},
 volume = {8},
 number = {1},
 month = jan,
 year = {1998},
 issn = {1049-3301},
 pages = {3--30},
 numpages = {28},
 doi = {10.1145/272991.272995},
 acmid = {272995},
 publisher = {ACM},
 address = {New York, NY, USA},
}

@misc{dieharder,
title={Dieharder: A Random Number Test Suite},
	author={Robert G. Brown and Dirk Eddelbuettel and David Bauer},
	year= 2019,
	Note={Duke.edu},
	addendum={Archive of 4 Aug 2019},
	url="https://web.archive.org/web/20190804063819/https://webhome.phy.duke.edu/~rgb/General/dieharder.php"
	}

@misc{accordion-rules,
  title={Accordion Solitaire},
  year={2003},
  url={https://web.archive.org/web/20030714125217/http://www.bvssolitaire.com:80/rules/Accordion.htm},
  author={{BVS Development Corporation}},
  addendum={Archive of 15 July 2003, original date unknown}
}

@misc{250sol_rules,
Title = {Rules of Some Patience Games from {``250+ Solitaire Collection"}},
author = {Ian P. Gent},
Note = {Ian Gent's Blog},
year = {2022},
month = {8},
day = {5},
url={https://web.archive.org/web/20220805214221/https://blog.ian.gent/2022/08/rules-of-some-patience-games-from-250.html}

}

@misc{fish_bhstats,
title={Solving Statistics for the First 1 Million PySolFC Black Hole Solitaire Deals},
url={https://web.archive.org/web/20220805135149/https://www.shlomifish.org/fc-solve-temp/mail-lists/fc-solve-discuss/archive/1034.html},
author={Fish, Shlomi},
year = 2010,
  addendum={Archive of 5 Aug 2022}
}

@misc{fish_simplesimon,
  title={Updated {Simple Simon} Statistics},
  url={https://web.archive.org/web/20220428151919/https://fc-solve.shlomifish.org/mail-lists/fc-solve-discuss/archive/0974.html},
  Note={Yahoo! Groups},
  author={Fish, Shlomi},
  year={2009},
  month=7,
  day=9,
  addendum={Archive of 28 Apr 2022}
}

@misc{fish_freecell_2cell_stats,
title={Two Freecell Solvability Report for the First 400,000 Deals },
url={https://web.archive.org/web/20130719010443/http://fc-solve.blogspot.com/2012/09/two-freecell-solvability-report-for.html},
author={Fish, Shlomi},
Note={fc-solve.blogspot.com},
year = 2012,
  addendum={Archive of 19 Jul 2013}
}

@misc{fish_fc_billiard,
  title={Report: The solvability statistics of the {Freecell Pro} 4-Freecells Deals},
  url={https://web.archive.org/web/20180815201227/https://fc-solve.shlomifish.org/charts/fc-pro--4fc-deals-solvability--report/},
  author={Fish, Shlomi},
  Note={ShlomiFish.org},
  year ={2018},
  addendum={Archive of 15 Aug 2018}
}

@misc{fish_freecell_0cell_stats,
title={freecell-pro-0fc-deals},
Note ={Github Repository},
url={https://web.archive.org/web/20220419155553/https://github.com/shlomif/freecell-pro-0fc-deals/blob/master/README.md},
author={Fish, Shlomi},
year = 2021,
addendum={Archive of 19 Apr 2022}
}

@misc{SolLab,
  author = "Michael Keller",
  title = "{FreeCell} -- Frequently Asked Questions ({FAQ})",
  year = "2015",
  Note = "Solitaire Laboratory",
  url = "https://web.archive.org/web/20181215222456/http://solitairelaboratory.com/fcfaq.html",
  addendum = "Archive of 15 Dec 2018"
}

@misc{Keller-Dominance,
  author = "Michael Keller",
  title = "When can {I} play the six of clubs? {O}bservations on the autoplay controversy",
  year = "2012",
  month = "11",
  day = "29",
  Note = "ShlomiFish.org",
  url = "https://web.archive.org/web/20220823080948/https://fc-solve.shlomifish.org/mail-lists/fc-solve-discuss/archive/1214.html",
  addendum = "Archive of 23 Aug 2022"
}

@misc{masten_winrates,
  title={Solitaire Win Rates and Analysis},
  url={https://web.archive.org/web/20221208175740/https://solitairewinrates.com/},
  year={2022},
  author={Masten, Mark},
  addendum={Archive of 8 Dec 2022}
}

@misc{masten_carpet,
  title={Carpet},
  url={https://web.archive.org/web/20221208175917/https://solitairewinrates.com/Carpet.html},
  year={2022},
  author={Masten, Mark},
  addendum={Archive of 8 Dec 2022}
}

@misc{masten_wormhole,
  title={Technical Details of {Mark} {Masten's} Worm Hole Solver},
  url={https://web.archive.org/web/20221208180007/https://solitairewinrates.com/WormHoleTechnicalDetails.html},
  year={2022},
  author={Masten, Mark},
  addendum={Archive of 8 Dec 2022}
}

@misc{masten_perpetual,
  title={Perpetual Motion (Narcotic)},
  url={https://web.archive.org/web/20221208175940/https://solitairewinrates.com/PerpetualMotion.html},
  year={2022},
  author={Masten, Mark},
  addendum={Archive of 8 Dec 2022}
}

@misc{internet-freecell,
author = "Chris Plante",
year = "2012",
Note = "The Gameological Society",
title = "Unbeatable",
url= "https://web.archive.org/web/20190516164853/http://gameological.com/2012/04/unbeatable/index.html",
Addendum="Archive of 16 May 2019"}

@misc{pringle_bakers,
  title={BakersGame-10million},
  url={https://bitbucket.org/theodorepringle/bakersgame-10million/},
  author={Pringle, Theodore},
  year={2017},
  month={5},
  day={12},
  Note={Bitbucket Repository. Accessed 19 September 2018, not archivable}
}

@misc{plspider,
	title={Winnable Spider Solitaire Games},
	author={Alex Robinson},
	Note = {Tranzoa.net},
	year= 2020,
	addendum={Archive of 5 Mar 2021},
	url="https://web.archive.org/web/20210305230500/https://www.tranzoa.net/~alex/plspider.htm"}

@misc{solitaire-100million,
title={Celebrating 30 Years of {Microsoft} {Solitaire} with Those Oh-So-Familiar Bouncing Cards},
author={Paul Jensen},
year=2020,
month={5},
day={22},
addendum={Archive of 22 May 2022},
Note={Xbox.com},
url="https://web.archive.org/web/20200522210053/https://news.xbox.com/en-us/2020/05/22/celebrating-30-years-microsoft-solitaire/"
}

@misc{american-canister-goodsol,
  title={American Canister},
  url={https://web.archive.org/web/20160324031437/https://www.goodsol.com/games/americancanister.html},
  author={Warfield, Thomas},
  year = 2016,
  Note={Goodsol.com},
  addendum={Archive of 24 Mar 2016, original date unknown}
}

@misc{easthaven_goodsol,
  title={EastHaven},
  year=2016,
  Note={Goodsol.com},
  url={https://web.archive.org/web/20160621140912/https://www.goodsol.com/games/easthaven.html},
  author={Warfield, Thomas},
  addendum={Archive of 21 Jun 2016, original date unknown}
}

@misc{northwest-territory-goodsol,
  title={Northwest Territory},
  year=2017,
  Note={Goodsol.com},
  url={https://web.archive.org/web/20171010061224/https://www.goodsol.com/games/northwestterritory.html},
  author={Warfield, Thomas},
  addendum={Archive of 10 Oct 2017}
}

@misc{seahaven_goodsol,
  title={Sea Towers ({Seahaven Towers})},
  year=2018,
  Note={Goodsol.com},
  url={https://web.archive.org/web/20180520082726/https://www.goodsol.com/games/seatowers.html},
  author={Warfield, Thomas},
  addendum={Archive of 20 May 2018}
}

@misc{spanish_goodsol,
  title={Spanish Patience},
  year=2017,
  Note={Goodsol.com},
  url={https://web.archive.org/web/20170629170622/https://www.goodsol.com/games/spanishpatience.html},
  author={Warfield, Thomas},
  addendum={Archive of 29 Jun 2017}
}

@misc{stronghold_goodsol,
  title={Stronghold},
  year=2019,
  Note={Goodsol.com},
  url={https://web.archive.org/web/20190122171602/https://www.goodsol.com/pgshelp/index.html?stronghold.htm},
  author={Warfield, Thomas},
  addendum={Archive of 22 Jan 2019}
}

@misc{beleaguered-castle-wikipedia,
  title={Beleaguered Castle},
  year=2017,
  url={https://web.archive.org/web/20170205150411/https://en.wikipedia.org/wiki/Beleaguered_Castle},
  publisher={Wikipedia},
  author={{Wikipedia Contributors}},
  addendum={Archive of 5 Feb 2017}
}

@misc{wolter_analysis,
  title={Experimental Analysis of Various Solitaire Games},
  url={https://web.archive.org/web/20170724143426/https://politaire.com/article/intro.html},
  journal={Politaire},
  Note="Politaire.com",
  author={Wolter, Jan},
  year={2013},
  month={4},
  day={11},
  addendum={Archive of 24 Jul 2017}
}

@misc{wolter_canfield,
  title={Experimental Analysis of Canfield Solitaire},
  url={https://web.archive.org/web/20180429220704/https://politaire.com/article/canfield.html},
  journal={Politaire},
  Note="Politaire.com",
  author={Wolter, Jan},
  year={2013},
  month={4},
  day={11},
  addendum={Archive of 29 Apr 2018}
}

@misc{wolter_canfield_rules,
  title={Rules for {Canfield}  Solitaire},
  url={https://web.archive.org/web/20150218063842/http://politaire.com/help/canfield},
  journal={Politaire},
  author={Wolter, Jan},
    Note="Politaire.com",
  year={2014},
  month={12},
  day={23},
  addendum={Archive of 18 Feb 2015}
}

@misc{wolter_thirtysix_rules,
  title={Rules for {Thirty Six} Solitaire},
  url={https://web.archive.org/web/20170619054612/http://politaire.com/help/thirtysix},
  author={Wolter, Jan},
  year={2014},
  Note="Politaire.com",
  month={12},
  day={23},
  addendum={Archive of 19 Jun 2017}
}

@misc{wolter_trigon_rules,
  title={Rules for {Trigon} Solitaire},
  url={https://web.archive.org/web/20170618222612/http://politaire.com/help/trigon},
  author={Wolter, Jan},
    Note="Politaire.com",
  year={2014},
  month={12},
  day={23},
  addendum={Archive of 18 Jun 2017}
}

@misc{droettboom2015understanding,
  title={Understanding {JSON} Schema},
  author={Droettboom, Michael},
  note={Space Telescope Science Institute},
  year={2023},
  month={1},
  day={11},
  url={https://json-schema.org/UnderstandingJSONSchema.pdf},
}

@misc{solvitaire-v0.10.2,
  author       = {Charlie Blake and
                  Ian P. Gent},
  title        = {{thecharlesblake/Solvitaire}: Release for {Zenodo DOI}-issuing (v0.10.2)},
  month        = Nov,
  year         = 2019,
  Note    = {Zenodo},
  version      = {v0.10.2},
  doi          = {10.5281/zenodo.3529524},
}

@misc{shootme_klondike,
  title={{Klondike-Solver}},
  url={https://web.archive.org/web/20180611030256/https://github.com/ShootMe/Klondike-Solver},
  Note={Github Repository},
  author={Matt Birrell},
  year={2017},
  addendum={Archive of 11 Jun 2018}
}

@misc{wolter-code,
  title={{solsolve:} A Solving Workbench for Various Solitaire Games},
  url={https://code.google.com/archive/p/solsolve/},
  Note={Google Code Archive},
  author={Wolter, Jan},
  year={2014},
  month={12},
  day={23},
  addendum={Accessed Sep 2018}
}

@misc{fish_fcsolve,
  title={Freecell Solver},
  url={http://fc-solve.shlomifish.org/},
  author={Fish, Shlomi},
  month=6,
  year=2024
}

@misc{birrell-pc,
  title={Re: {Solitaire Solver}},
  author={Matt Birrell},
  year={2018},
  month={11},
  day={18},
  Note = {Email to Ian Gent, 18 November}
  }

@misc{cabral-pc,
  title={Seahaven {Towers}},
  author={Arthur W. Cabral},
  month={9},
  day = {1},
  year={2019},
  Note={Email to Ian Gent, 1 September}
  }

@misc{pringle-pc,
author = {Pringle, Theodore},
year = 2018,
month = {7},
day={25},
title = "Re: {Query} about your {Baker's Game} results",
Note = "Email to Ian Gent, 25 July"
}

@misc{roscoe-pc,
  title={Re: {Patience}},
  author={A. W. Roscoe},
  month={8},
  day = {22},
  year={2019},
  Note = "Email to Ian Gent, 22 August"
  }

@misc{graham-mackenzie,
title="Email exchange reported by {M}ackenzie to {I}an {G}ent, 21 {N}ov 2019",
author="Dana Mackenzie and Ronald Graham",
year = 2019
}

\appendix

\section{Rules Of Patience Games}
\label{sec:rules}
\label{app:rules}

\Cref{rules-table} shows the rule for most main games in this paper. Exceptional games not in this table are Accordion, Late-Binding Solitaire and Gaps variants, for which detailed rules in JSON are shown in Listings~\ref{rules-accordion} and~\ref{rules-gaps}. 
\label{key-rules-table}

\lstset{
  basicstyle=\ttfamily\footnotesize,
  columns=fullflexible,
  keepspaces=true,
}
\begin{lstlisting}[float,label=rules-accordion,caption={{Rules of \gamename{Accordion}. The rules of \gamename{Late-Binding Solitaire} are the same except with size 18.}.}]
  "foundations": { 
    "present": false},
  "tableau piles": {
    "count": 0},
  "accordion": {
    "size": 52,
    "moves": ["L1", "L3"],
    "build policies": ["same-suit", "same-rank"]}
\end{lstlisting}

\begin{lstlisting}[float,label=rules-gaps,caption={Rules of \gamename{Gaps (One Deal)}. The rules of \gamename{Gaps (Basic Variant)} are the same except with fixed suit true.}]
  "foundations": { i
    "present": false},
  "tableau piles": {
    "count": 0},
  "sequences": {
    "count": 4,
    "direction": "L",
    "build policy": "same-suit",
    "fixed suit": false}
\end{lstlisting}

To present most games in a uniform format, \Cref{rules-table} uses a very concise notation which is explained below.  Where variants of a game are reported in this paper, the rules are as given here except for the stated change, e.g. Fore Cell (Same Suit) has the same rules as Fore Cell except with BP set to $=$.

\begin{description}
    \item[Game:] Name of game, plus citation which gives the name and rules we use (but is not necessarily the inventor of the game).
     \textit{Symbols Used:} \originalnote Game invented for this paper
    \item[Decks:] Number of complete decks used in the game, of 52 cards by default. 
        \textit{Symbols Used:}
        \beziquenote Deck of 32 cards, we use A+2-8 of each suit. 
    \item[Foundations:] Rules for Foundations or Hole. 
        Number of cards initially placed in foundations, $\bullet$ for hole being used, or S for \gamename{Spider}-type elimination of suits. \textit{Symbols Used:}
       $\checkmark$ Worrying back from foundations to tableau is allowed. 
       $\times$ Worrying back is not allowed.
         \randomcard Random base of foundation.
\golfnote King/Ace not considered adjacent in rank. 

\item[Tableau:] Number of tableau piles, plus shape of tableau.  
\textit{Symbols Used:}  $\square$ piles all of same length except possibly for some piles of one extra length; $\varbigtriangleup$ piles in triangular form;  solid shapes indicates that cards face-down except the top card, otherwise all cards face-up.  \fannote we use 16 piles of 3 and 2 piles of 2.
\item{\textbf{TC}:} Tableau cards - total number of  cards placed in the tableau.
	\item{\textbf{BP:}} Build Policy, rule by which one card may be placed on another in the tableau. Where allowed, cards must be one lower in rank (including from K on A if Foundations start on random base).
	\textit{Symbols Used:} $\times$ building not allowed; * card of any suit allowed; \rb~ card must be of opposite colour (red on black or black on red); = card of same suit. 

	\item{\textbf{MG}:} Move of Groups, whether or not a consecutive sequence of built cards may be moved as a unit in the tableau.
\textit{Symbols Used:} $\times$ not allowed; $\checkmark$ allowed with the same restriction as BP; = allowed for sequence of cards of the same suit. \wholepilemoves Only entire piles may be moved.
	
	\item{\textbf{SP:}} What card may be put in a free space in the tableau, or sequence if MG allows it.
	\textit{Symbols Used:}
	$\times$ Spaces may not be filled; $\checkmark$ Spaces may be filled by any card; K Spaces may be filled by a K only (or card one rank below foundation base if random). \canfieldspaces Space must be refilled immediately from stacked reserve until that is empty, then may be filled freely.
\fortunesspaces Space must be refilled immediately from waste (or stock if empty).

\item{\textbf{Stock}:} The first number indicates the number of cards in the stock; the second symbol indicates the number of cards drawn at a time from a stock, with $\square$ indicating that one card from stock is dealt to each tableau pile; the third number indicates whether no redeals are allowed when the stock is empty or an infinite number are.

	\item{\textbf{FC}:} The number of Free Cells in the game, if any, followed by the number of free cells that are filled at the start of the game.
 
\item{\textbf{Reserve:}} The size of any Reserve in the game. S indicates that the reserve is `stacked', i.e. only the top card of it is available for play. Otherwise all cards are available at any time. 

\end{description}

\begin{table}
    \caption{Detailed rules of  main games studied in this paper, excepting those in Listings~\ref{rules-accordion} and~\ref{rules-gaps}.  See page \pageref{key-rules-table} for key.  \originalnote Original game.
    }
\begin{footnotesize}
\begin{center}

\begin{tabular}{| l |c|c| rrclc | ccc |  l|l |}
\hline
 \multicolumn{1}{|c|}{\textit{Game} \hfill\textit{Rules Citation}}& \textit{Decks} & \multicolumn{1}{c|}{\textit{Foundations}}& \textit{Tableau}& TC& BP & MG  & SP  
    & \multicolumn{3}{c|}{\textit{Stock}} & 
 \textit{FC} & \textit{Reserve} \\ 
\hline
Alpha Star \hfill\cite{250sol_rules} & 1 & 4 $\times$ & 12 $\square$ & 48 & \samesuit &  $\checkmark$ & $\checkmark$ & \multicolumn{3}{l|}{} &  &   \\
American Canister \hfill\cite{american-canister-goodsol}  & 1 & 0 $\times$ & 
8 $\square$ & 52 
& \rb & 
$\checkmark$ & $\checkmark$ & &&&& \\
Baker's Game \hfill\cite{parlett}& 1 & 4 $\times$ & 8 $\square$ & 52 & = & $\times$ & $\checkmark$ & &&& 4 0 &  \\
Beleaguered Castle \hfill\cite{parlett}&1&4 $\times$&8 $\square$ & 48 & \anysuit & $\times$ & $\checkmark$ & \multicolumn{3}{l|}{} &  &   \\
Black Hole \hfill\cite{parlett}& 1 & $\bullet$	 $\times$ & 17 $\square$ & 51 & $\times$ & $\times$ & $\times$ & \multicolumn{3}{l|}{} &  &   \\
(British) Canister \hfill\cite{parlett} & 1 & 0 $\times$ & 8 $\square$ & 52 & \rb & $\times$ & K & &&&& \\
Canfield \hfill\cite{wolter_canfield_rules}\hfill & 1 & 1 $\times$ \randomcard & 4 $\square$  & 4 &  \rb & $\checkmark$ & \canfieldspaces & \multicolumn{3}{l|}{ 34 3 $\infty$ }&  & 13 S  \\

Delta Star \hfill\cite{250sol_rules} & 1 & 4 $\times$ & 12 $\square$ & 48 & = & $\times$ & $\checkmark$ & &&& & \\
East Haven \hfill\cite{easthaven_goodsol}& 1 & 0 $\checkmark$ & 7 $\blacksquare$ & 21 & \rb &
$\times$ &$\checkmark$& \multicolumn{3}{l|}{ 31 $\square$ 0 } & &  \\ 
Eight Off \hfill\cite{parlett} & 1 & 0 $\times$ & 8 $\square$ & 48 & = & $\times$ & K & &&& 8 4 &  \\
Fan \hfill\cite{parlett}  & 1 & 0 $\checkmark$ & 18 $\square$ \fannote& 52  & \samesuit & $\times$ & K & \multicolumn{3}{l|}{} &  &  \\
Fore Cell \hfill\cite{SolLab}&1&0 $\times$& 8 $\square$ & 48 & \rb & $\times$ & K & &&& 4 4 & \\
Fortune's Favor \hfill\cite{parlett} & 1 & 4 $\times$ & 12 $\square$ & 12 & = & $\times$ & \fortunesspaces & \multicolumn{3}{l|}{ 36 1 0 }& & \\
Freecell \hfill\cite{SolLab} & 1& 0 $\times$&  8 $\square$  & 52 & \rb & $\times$ & $\checkmark$ &  \multicolumn{3}{l|}{}& 4 0 &    \\ 
Golf \hfill\cite{parlett}& 1 & $\bullet$	 $\times$ \randomcard \golfnote& 7 $\square$ & 35 & $\times$ & $\times$ & $\times$ & \multicolumn{3}{l|}{16 1 0} &  &   \\
King Albert \hfill\cite{parlett}& 1&0 $\checkmark$&9 $\varbigtriangleup$  & 45 & \rb & $\times$ &   $\checkmark$ & \multicolumn{3}{l|}{} &  & 7  \\
Klondike \hfill\cite{rollout1} & 1&0 $\checkmark$&7 \filledtriangle& 28  & \rb  & $\checkmark$ & K & \multicolumn{3}{l|}{  24 3  $\infty$} & & 	  \\ 
Mrs Mop \hfill\cite{parlett}& 2 & \spidertype $\times$ & 13 $\square$ & 104 & * & = & $\checkmark$ & &&& & \\
Northwest Territory \hfill\cite{northwest-territory-goodsol} & 1 & 0 $\checkmark$ & 8 $\filledtriangle$ & 36 & \rb & $\checkmark$ & K & &&& & 16 
\\
Raglan \hfill\cite{parlett} & 1&4 $\checkmark$&9 $\varbigtriangleup$  & 42 & \rb & $\times$ &  $\checkmark$ & \multicolumn{3}{l|}{} &  & 6  \\
Seahaven Towers~\hfill\cite{seahaven_goodsol,cabral-pc} & 1 & 0 $\times$ & 10 $\square$ & 50 & = & $\times$& K & &&& 4 2 &  \\
Siegecraft \hfill\cite{beleaguered-castle-wikipedia} &1&4 $\times$&8 $\square$ & 48 & \anysuit & $\times$ & $\checkmark$ & & && 1 0 &   \\
Simple Simon \hfill\cite{parlett}&1&S $\times$& 10 $\varbigtriangleup$ & 52 & * & = & $\checkmark$ &&&&&\\
Somerset \hfill\cite{parlett} & 1&0 $\checkmark$&10 $\varbigtriangleup$  & 52 & \rb & $\times$ &  $\checkmark$ & \multicolumn{3}{l|}{} &  &   \\
Spanish Patience \hfill\cite{spanish_goodsol} & 1&0 $\checkmark$& 13 $\square$ & 52 & \anysuit & $\times$ & $\checkmark$ & & & & & \\
Spiderette \hfill\cite{parlett}&1&S $\times$& 7 $\filledtriangle$ & 28 & * & = & $\checkmark$ &\multicolumn{3}{l|}{ 24 $\square$ 0 }&&\\
Spider \hfill\cite{parlett}&2&S $\times$& 10 $\blacksquare$ & 54 & * & = & $\checkmark$ &\multicolumn{3}{l|}{ 50 $\square$ 0 }&&\\
Streets \& Alleys \hfill\cite{parlett}&1&0 $\times$& 8 $\square$ & 52 & \anysuit & $\times$ & $\checkmark$ & \multicolumn{3}{l|}{} &  &   \\
Stronghold \hfill\cite{stronghold_goodsol}&1&0 $\times$&8 $\square$ & 52 & \anysuit & $\times$ & $\checkmark$ & & && 1 0 &   \\
Thirty \hfill\cite{parlett}  &1\beziquenote &0 $\times$&5 $\square$&30&\anysuit & $\checkmark$&$\checkmark$&&&&&2\\
Thirty Six  \hfill\cite{wolter_thirtysix_rules}&1&0 $\times$&6 $\square$&36&*&$\checkmark$&$\checkmark$&\multicolumn{3}{l|}{16 1 0}&&\\
Trigon \hfill\cite{wolter_trigon_rules} &1&0 $\times$&7\;\filledtriangle& 28  & \samesuit  & $\checkmark$ & K & \multicolumn{3}{l|}{  24 3  $\infty$}&  	&\\
Will O' The Wisp \hfill\cite{parlett}&1&S $\times$& 7 $\blacksquare$ & 21 & * & = & $\checkmark$ &\multicolumn{3}{l|}{ 31 $\square$ 0 }&&\\ 
Worm Hole \hfill\originalnote & 1 & $\bullet$	 $\times$ & 17 $\square$ & 51 & $\times$ & $\times$ & $\times$ & \multicolumn{3}{l|}{} & 1 0 &   \\
    \hline
\end{tabular}
\label{rules-table}
\end{center}
\end{footnotesize}
\end{table}

\section{Proof of Correctness of Key Dominances}
\label{sec:dominance-proofs}
\label{app:dominance-proofs}

The proofs in this section will proceed by permuting the move order, e.g. swapping the order of  consecutive moves.  For this reason, we will require that moves are not made illegal by their position in the move sequence: so the proof would not apply to a game where moves to foundation could only be made if the position in the move sequence were divisible by 3.

We define a move $m$ as being a pair $[c,t]$ where $c$ is the card being moved and $t$ is the card or location it is moved to. (We assume some unambiguous notation where cards are moved to locations such as spaces which are currently empty.) We write  $C(m)$ for the card being moved by a move, and $T(m)$ for the card or location moved to, i.e. if $m=[c,t]$ then $C(m)=c$ and $T(m)=t$.  
We remark that, in games with multiple decks, there will be distinct cards with the same suit/rank, but these are still considered separate cards.

For the purposes of the proofs, we will call a move `\compliant' if it respects the constraints the dominance places on moves in either disallowing or requiring certain moves, and `\noncompliant' if it does not.

The general pattern of the proofs is to start from 
 any winning sequence of moves in the original rules, containing at least one \noncompliant move.  From this we will create a new sequence of moves which is both legal and a winning sequence. The new sequence will either have fewer \noncompliant moves than the original, or have the same number but with the last \noncompliant move closer to the end of the sequence than before. 
This means that by repeated application of the process we will eventually obtain a sequence of moves that wins the game and has zero \noncompliant moves.  Thus there is always a \compliant sequence available.

\subsection{Safe Moves To Foundations}
\label{sec:dominance-proof-safe-moves}
\label{app:dominance-proof-safe-moves}

In \Cref{sec:safemoves}, we described dominances which apply to moving cards from the tableau, as well as from a free cell or the reserve.  However, it is not safe to enforce this dominance from the stock - as we discuss in \Cref{canfield-correction}. The exception is when the stock draw size is 1 and infinite redeals of stock are allowed: in this case the stock can be treated as if it were a reserve.   While previously described \cite{Keller-Dominance}, they have not previously been proven correct formally.  We give the first such proof in this section.

\begin{definition}[Safely buildable] 
~
	\begin{itemize}
    \item A card $c$ is called `\textbf{potentially safely buildable}' to foundation  at time $i$ if: 
    \begin{itemize}
    \item
     the card of one rank lower than $c$ and the same suit is already on foundation or card $c$ is the first card to be played to its foundation (usually an Ace); 
    \item 
    \textbf{and,} if another card $d$ has $c$ as the target location of any possible future move $[d,c]$ (other than foundation build), 
    then $d$ is also potentially safely buildable at  time $i$.
	\end{itemize}
 \item A card $c$ is called `\textbf{safely buildable}' to foundation at time $i$ if it is potentially safely buildable and the move of $c$ to foundation is legal at time $i$.
 
\end{itemize}
\label{def:safe-buildable}
\end{definition}

We restrict consideration to games which involve building to foundation and moving all cards there to win.  We also assume that we do not have multiple decks: i.e. while the theorem applies to a single deck with eight different suits of two colours, it does not apply to a game with two copies of the standard deck.  The occurrence of duplicate cards leads to potential edge cases that we do not consider in this proof.   Finally, since our proof involves permuting moves, we assume that the game has no rules in which move order affects a moves legality.

\FloatBarrier

\begin{thm}
In a game of the type described above, a winnable instance 
is also winnable with the restriction that when any card currently  in the tableau/free cell/reserve is safely buildable, the next move must be the move of a safely buildable card to foundation.
\label{thm:safemoves}
\end{thm}

\paragraph{Proof.} 

Consider any winning sequence of moves in the original rules, $m_1, m_2, \ldots, m_n$, containing at least one \noncompliant move. Suppose that the move $m_i = (c_i,t_i)$ is the last move in the sequence which is \noncompliant. As the last \noncompliant move, we know that: $i<n$ since 
the last move in a winning game must be moving a card to the foundation pile; $m_i$ is not the move of a safely buildable card to foundation;  that there must be at least one safely buildable card at time $i$; and that move $m_{i+1}$ is a compliant move.  
There are two cases: either  card $c_i$ was safely buildable at time $i$ or it was not.  

\begin{itemize}
    \item The first case is that $c_i$ was safely buildable at time $i$.
    In the original sequence of moves, $c_i$ was safely buildable at time $i$ and so must have been at the bottom of a built group.  It will remain there and so must also be safely buildable at time $i+1$. 
    Since $m_{i+1}$ and all subsequent moves were compliant, there must have been a consecutive sequence of moves from $m_{i+1}$ of safe builds to foundation, and one of these - say $m_j$ - was  the first move of $c_i$ to foundation. 
    All these moves remain legal and compliant.
      We now delete the original move $m_i$ and replace it with $m_j$, giving a new 
   sequence of moves is $m_j, m_{i+1} \ldots m_{j-1}, m_{j+1} \ldots m_n$.  This subsequence has  no \noncompliant moves as we deleted $m_i$.
    
    \item The second case is that $c_i$ was not safely buildable at time $i$.  First note that move  $m_i$ cannot have moved $c_i$ either from or to \textit{any} safely buildable card $d$.
    Card $c_i$ cannot have moved \textit{from} $d$ because $d$ was playable to foundation and therefore uncovered.  Card $c_i$ cannot have moved \textit{to} $d$ by definition of $d$ being  safely buildable: any card movable to $d$ must itself be potentially safely buildable but card $c_i$ was not.  Together this means that any safely buildable card is still safely buildable  at time $i+1$.  Since it was \compliant, $m_{i+1}$ must be the safe build of a card to foundation.  Although we cannot delete $m_i$ as we did in the first case, we can swap $m_i$ and $m_{i+1}$ 
    to give the new sequence $m_{i+1}, m_i, m_{i+2} \ldots m_n$.  
    The move $m_i$ must remain legal as it did not involve the safely buildable card $c_{i+1}$.  
    Move $m_i$ may remain \noncompliant but even if it does, it is one move nearer the end of the sequence.  All other moves must remain legal and compliant.
\end{itemize}

The new sequence will either have fewer \noncompliant moves than the original, or have the same number but with the last \noncompliant move closer to the end of the sequence than before. 
This means that by repeated application of the process we will eventually obtain a sequence of moves that wins the game and has zero \noncompliant moves.  
$\blacksquare$

It might seem that the proof would potentially be invalidated by worrying back cards from foundation to the tableau.  However, the proof applies equally to them.  Note that cards which are already in the foundation can still be potentially safely buildable, so they still count as possible cards that could be built in the tableau.  Worrying back a card that is potentially safely buildable is certainly pointless, so we can add the following simple corollary.   

\begin{corollary}
We can correctly add a dominance that disallows worrying back a card from foundation if it would be immediately safely buildable after being worried back. 
\end{corollary}

\paragraph{Proof.} We can enforce the dominance of moving safely buildable cards, which means the card must be immediately moved back to foundation without any non-foundation move intervening.  Therefore the worry-back and rebuilding moves cancel out and can safely both be deleted.  $\blacksquare$

\begin{corollary}
The dominances we outlined in \Cref{sec:safemoves} for automatically building cards to foundation are correct. 
Specifically, for the following build policies a card is potentially safely buildable when the given condition holds: 
\begin{description}

\item{\normalfont{\textbf{Red-black building with worrying back:}}}
a card is at most \textit{two} more than the top card on all foundations of the opposite colour and at most \textit{three} more than the current card on foundation of the other suit of the same;

\item{\normalfont{\textbf{Red-black building without worrying back:}}}
either the previous condition holds, or the card is no more than one higher ranked than all the foundations of the opposite colour, or both;

\item{\normalfont{\textbf{Building by suit:}}}
a card is buildable to foundation, i.e. one higher than the highest card on the foundation of the same suit;  
 
\item{\normalfont{\textbf{Building regardless of suit:}}}
a card is no more than two higher than the lowest card yet built to foundation of any suit.
\end{description}

\end{corollary}

\paragraph{Proof.} 

For each case we prove the condition is sufficient to ensure potential safe buildability by \Cref{def:safe-buildable}.

\begin{description}

\item{\normalfont{\textbf{Red-black building with worrying back:}}}

For a card $c$ of rank $r$ which meets the conditions, then only a card $d$ of rank $r-1$ of the opposite colour can be built onto $c$, but $d$ is potentially buildable to foundation since each opposite colour foundation is of rank at least $r-2$.  It would still be possible to build another card $e$ of $c$'s colour onto $d$: $e$ must therefore be of rank $r-2$.  But $e$ is potentially buildable since all foundations of this colour are at least at rank $r-3$.  Similarly $d$ is potentially safely buildable  since the only cards that can be moved to $d$ are $e$ or equivalent cards.  All cards of lower rank than $e$ are already on foundation: though they could be worried back to the tableau, they are themselves potentially safely buildable.   Therefore $c$ is potentially safely buildable under this build policy.

\item{\normalfont{\textbf{Red-black building without worrying back:}}}

The argument above for card $c$ of rank $r$  holds when that condition applies.  For the additional condition, if all cards of opposite colour and rank $r-1$ are built to foundation, there are no cards which can possibly be built onto $c$ in the tableau.  In this case the second condition of being potentially safely buildable in \Cref{def:safe-buildable} is vacuous and trivially true.

\item{\normalfont{\textbf{Building by suit:}}}
Similar to the last case, if the card of rank $r-1$ and the same suit is on the foundation, then the second condition in \Cref{def:safe-buildable} is vacuous.

\item{\normalfont{\textbf{Building regardless of suit:}}}
For card $c$ of rank $r$, since all cards of rank $r-2$  are on foundation,  only cards of rank $r-1$ can be built on $c$ in the tableau.  But all these can be built to foundations themselves and are thus potentially safely-buildable.

\item$\blacksquare$
\end{description}

\subsection{Immediate Building After Tableau Moves}
\label{dominance-proof}
\label{sec:dominance-proof}
\label{sec:dominance-proof-partial-pile}
\label{app:dominance-proof-partial-pile}

For an introduction to this dominance, see \Cref{sec:dominance-tableau}.   To maximise the utility of proving correctness, we wish to generalise the dominance and also strengthen it slightly from its original form.   The strengthening of the dominance is to insist that after a partial-pile move, the card above must be built immediately to foundation (not just be buildable in principle).  

The generalisation over previous uses is to allow its application in cases which don't use a standard four-suit deck or the common red-black building policy.  To do so we assume that the build policy has the property that, given any two cards, the two sets of places those cards can move to by  the build policy are either identical or disjoint.  We call a build policy an \textbf{indistinguishable} building policy if it both has  no distinction between two cards which can move to the same place, and it has no distinction between the rule for building individual cards and for moving piles of cards in a block.  For example, in the classic games using red-black building by rank, any two cards of different colours or ranks have disjoint cards they can be built on, while two cards of the same colour and rank can be built on exactly the same set of cards. Another indistinguishable policy would be in a game using five identical decks with three suits in which cards must be built in the same suit only: here there would be five copies of e.g. 9$\spadesuit$, each of which could be built on 10$\spadesuit$ but not on $10\clubsuit$ or $10 \heartsuit$.
Despite its flexibility this generalisation still excludes some  build policies.  An example is `different-suit'. This does not meet the condition because both spades and diamonds can move to clubs, while spades but not diamonds can move to diamonds.    Counterexamples to the theorem would occur if we allowed this build policy.  For example, we might need  a group headed by 9$\spadesuit$ to move from 10$\clubsuit$ to 10$\diamondsuit$ to allow the 9$\diamondsuit$ to move under the 10$\clubsuit$, as it cannot be moved to the 10$\diamondsuit$.  
An  indistinguishable build policy  also requires that the same build policy controls moving groups and individual cards.  Some games, such as \gamename{Spider}, use a policy where individual cards  
can be moved in any suit but built groups can only be moved if they are all the same suit.  This means that moves of groups can be necessary to establish sequences of the same suit, even if the card above is not buildable.

\begin{thm}
We consider any patience or solitaire game which: 
has a tableau which builds down according to an `indistinguishable' build policy (as defined above);  
allows moves of complete or incomplete built piles as a single move according to the same policy as for individual cards; 
the only place a card can move from the tableau is to another tableau pile or to a foundation; 
is won by moving all cards to the foundations; 
and contains no rules invalidating moves by constraints on their order in the move sequence.
For any instance of such a game, if the instance is winnable with the original rules, then it is also winnable with the restriction that an incomplete built pile may only be moved if the card above the moved partial pile is then built immediately to foundation.
\label{thm:dominance}
\end{thm}

\paragraph{Proof}

Consider any legal winning sequence of moves in the original rules, $m_1, m_2, \ldots, m_n$, containing at least one \noncompliant move.   We will create a new winning sequence of moves with either  fewer \noncompliant moves than the original, or have the same number but with the last \noncompliant move closer to the end of the sequence than before. 

Suppose that the move $m_i$ is the last move in the sequence which is \noncompliant, i.e. is the move of a partial pile not immediately followed by building the card above it to foundation.
Note that $i<n$ since 
the last move in a winning game must be moving a card to the foundation pile.  For $m_{i+1}$ we do know: it exists since $i < n$; it is a legal move; it is not the move of the card of above the just-moved partial pile to foundation; and if $m_{i+1}$ is a partial pile move then move $m_{i+2}$ is building the card above it  to foundation. 
We now show by case analysis how to replace moves $m_i, m_{i+1}$ in the sequence.  In most cases the adjustment is straightforward.  Before describing the straightforward cases, we consider the most critical, difficult,  case. 

\paragraph{Case 1.}
The critical case is where move $m_{i+1}$ is moving a card or pile onto the pile just vacated by the original move $m_i$ (which was by hypothesis the last \noncompliant move). 
We can illustrate by example in the case of a red-black build policy: this might be a move of a three card pile $10 \clubsuit9\heartsuit8\spadesuit$ from the J$\diamondsuit$ to J$\heartsuit$, followed immediately by a move of the 10$\spadesuit$ to the J$\diamondsuit$, i.e. $m_i = [10\clubsuit,J\heartsuit]$ and 
$m_{i+1} = [10\spadesuit,J\diamondsuit]$.
We deal with this case first by omitting move $m_i$, thus reducing the number of \noncompliant moves by one, and then replacing $m_{i+1}$ by a move $m'_{i+1} = [C(m_{i+1}),T(m_i)]$.  In the example above, we would delete the move of 10$\clubsuit$ and change the move of the 10$\spadesuit$ to be to the J$\heartsuit$ instead of the J$\diamondsuit$, i.e. $m'_{i+1} = [10\spadesuit,J\heartsuit]$.  The move $m'_{i+1}$ must be a legal move because the build policy is indistinguishable so cannot allow $m_{i+1}$ and disallow $m'_{i+1}$. Move $m'_{i+1}$ must also be \compliant since move $m_{i+1}$ was: i.e. if the move was of a partial pile then $m_{i+2}$ must be building the card above to foundation.

We now have to consider the  remaining moves in $m_{i+2}, \ldots m_n$.  We create moves $m'_{i+2}, \ldots m'_j$ until we have identical layouts again in the original and new sequence of moves, after which we retain moves $m_{j+1} \ldots m_n$.  
Until then, we will maintain an invariant property, that the cards $T(m_i)$ and $T(m_{i+1})$ (J$\diamondsuit$ and J$\heartsuit$ in our example) remain in the tableau, that at least one of them has a card built below it,
that the piles under those cards are swapped in the new sequence compared to the original, and that all other cards in the layout are identical. 
This invariant certainly holds after the deletion of $m_i$ and the replacement of $m_{i+1}$ by $m'_{i+1}$. Now we assume the invariant is true up to move $m'_{k-1}$ and consider move $m_k$.
If this move  does not involve either of the affected piles, then it necessarily retains the invariant, so we simply set $m'_k=m_k$. However, when the move does involve at least one of the affected piles, it must fall into one of the following five subcases. In the first three we have to adapt the sequence of moves to a new one to retain the invariant, with the last two being simpler.

\begin{enumerate}[label=(\alph*)]
\item If move $m_k$ is of $C(m_i)$ to $T(m_{i+1})$ (e.g. of 10$\clubsuit$ to J$\diamondsuit$ in our example) then we simply delete the move completely. Because of the invariant the cards below $C(m_i)$ are already at the intended target location, so we need do nothing. 
\label{exactreverse} Included in this sub-case is where move $m_{i+1}$ is an exact reverse of $m_i$ and both are deleted. In general, the final move sequence is 
$$m_1, \ldots m_{i-1}, m'_{i+1}, \ldots m'_{k-1}, m_{k+1}, \ldots m_n$$

\item If move $m_k$ is of $T(m_{i+1})$  to foundation
\textit{[respectively $T(m_i)$]}, e.g. moves J$\diamondsuit$  to foundation in our example \textit{[respectively J\;$\heartsuit$]}, then it now has a card built below it (by the invariant) so the move is not currently possible.  By the invariant, the card $T(m_{i})$  must itself be clear, since the card $T(m_{i+1})$ was clear in the original  \textit{[respectively $T(m_{i+1})$ must be clear]}. This means that we can now insert the move $m_k'=[C(m_i),T(m_i)]$ immediately before $m_k$ \textit{[respectively set $m_k'=[C(m_{i+1}),T(m_{i+1})]$]}.  
The move $m_k'$ is legal by indistinguishability.  It is a partial pile move where the immediately following move $m_k$ will be of the card above it to foundation, so $m_k'$  is  a \compliant move.
The move $m_k$ is now legal, positions are identical, and the new sequence contains one less \noncompliant move. The final move sequence is 
$$m_1, \ldots m_{i-1}, m'_{i+1}, \ldots m'_{k-1}, m_k', m_k, m_{k+1}, \ldots m_n$$

\item If move $m_k=[C_k,T_k]$ removes the last card below $T(m_i)$ and $T(m_{i+1})$, e.g. moves either 10$\clubsuit$ or 10$\spadesuit$ in our example, leaving both J$\heartsuit$ or J$\diamondsuit$ uncovered, then we set $m'_k$ to the same move $[C_k,T_k]$.   The move $m'_k$ must be legal, by indistinguishability.    If the move is to foundation or the the  move $m_{k+1}$ is building the card above $C_k$  to foundation, then $m'_k$ is \compliant and the sequence has one less \noncompliant move. Even if  $m'_k$ is \noncompliant, then the sequence has the same number of \noncompliant moves as before (since we deleted $m_i$), but the last is nearer the end of the sequence (since $n-k > n-i$). 
 By the invariant the layouts are now identical so the final move sequence is 
$$m_1, \ldots m_{i-1}, m'_{i+1}, \ldots m'_{k-1}, m'_{k}, m_{k+1}, \ldots m_n$$

\item If move $m_k$ is from (or to) a card on a pile below either $T(m_i)$ or $T(m_{i+1})$, but is not covered by one of the above sub-cases (e.g. of $8\spadesuit$ from the 9$\heartsuit$ to the 9$\diamondsuit$ in our example) then we can make the unchanged move $m_k$ now. That is, we move from $C(m_k)$ to $T(m_k)$: by the  invariant the identical card that $m_k$ was originally moved from (or to) is below the other one in the revised sequence.
 The invariant is thus retained. 
\item Any move $m_k$ of a pile of cards starting from $T(m_i)$ or $T(m_{i+1})$ or any card above them can be retained unchanged with $m'_k = m_k$ and the invariant is retained. 

\end{enumerate}

 All remaining cases are essentially straightforward because we can simply swap consecutive moves $m_i$ and $m_{i+1}$, sometimes with  minor changes.  In each case the position after the second move is identical in each sequence, and we have either removed an \noncompliant move or moved it one move closer to the end of the sequence, as required. 

\paragraph{Case 2.}
If the moves $m_i$ and $m_{i+1}$ are entirely unrelated then  we can simply swap the order of  the moves as they do not affect each other.  I.e. we create a new move sequence $$m_1, \ldots m_{i-1}, m_{i+1}, m_i, m_{i+2}, \ldots m_n$$
Note that the swap cannot affect  whether move  $m_{i+1}$ is \compliant: by hypothesis it was a \compliant move and it remains so.  However, it is possible that, in its new position, the move $m_i$ is now a \compliant move. If that happens we have reduced the number of \noncompliant moves, but if not we have moved the last \noncompliant move one closer to the end of the move sequence.

\paragraph{Case 3.} We can make consecutive moves \textit{from} the same pile, i.e. have $C(m_{i+1})$ be either the card above $C(m_{i})$ or a card in sequence above it. Because move $m_i$ is \noncompliant, the move $m_{i+1}$ cannot be of the card above $C(m_{i})$ to foundation, so the only remaining possibility is a second consecutive move between tableau piles. Again we can swap the order of the moves. The new move sequence is  $$m_1, \ldots m_{i-1}, m_{i+1}, m_i, m_{i+2}, \ldots m_n$$
Notice that in this case the card $C(m_i)$ (and any partial pile below it) is moved twice instead of just once, but ends in an identical position. The move $m_{i+1}$ must still be \compliant 
in the earlier position, while $m'_i$ remains \noncompliant, but appears one move nearer the end of  the sequence.  

\paragraph{Case 4.} 
We can make consecutive moves \textit{to} the same pile. In this case again we simply swap moves $m_i$ and $m_{i+1}$. 
The analysis is the same as in the previous case, except that this time it is the card $C(m_{i+1})$  (and possibly partial pile below it) that is moved twice instead of once.  Again,  $m_i$ remains \noncompliant, but appears one move closer to the end of the sequence.

\paragraph{Case 5.} The final possibility is that the second move is from the pile the first move went to. This gives a number of possibilities depending on the card moved the second time: the second card moved can be the same as the first card, a card above it in the new pile, or a card below it.
\begin{itemize} 
\item If the same card is moved twice, then the moves cannot be an immediate reversal of moves since that was covered as case~\ref{exactreverse} in the Critical Case above.
Thus, we can replace the two moves with a single move bypassing the intermediate position, $m'_i = [C(m_i),T(m_i+1)]$.    This move $m'_i$ may still be \noncompliant but is one move closer to the end. In this case the final sequence of moves is 
$$m_1, \ldots m_{i-1}, m'_i, m_{i+2}, \ldots m_n$$
\item If the second card is either above or below the first moved card, then again we can just swap the order of the two moves, giving the sequence 
$$m_1, \ldots m_{i-1}, m_{i+1}, m_i, m_{i+2}, \ldots m_n$$
The result is the same, with the \noncompliant move being one nearer the end of the sequence. If the card  $C(m_{i+1})$ was above the first moved card in the second pile, then the card  $C_{m_i}$ and any pile below it is only are now only moved once instead of twice. If the card $C(m_{i+1})$ was below $C(m_i)$ then the card $C(m_i)$ and any cards below it are now moved only once.
\end{itemize}

 In all of the cases analysed above, we are able to do one of two things.  We either produce a new sequence with one less \noncompliant move, or produce a sequence with the same number of \noncompliant moves but the last one nearer the end of the sequence.  Iterating this procedure must inevitably lead to a solution with zero \noncompliant moves.  Therefore we can impose the restriction without making any instance unsolvable.  $\blacksquare$

We wish to use the two dominances together, and have to consider the possibility that they might be acceptable individually, but together make a winnable instance unwinnable.  Fortunately, it is straightforward to  prove that this is not the case. 

\begin{thm}
If the conditions of \Cref{thm:safemoves} and \Cref{thm:dominance} both apply, then any winnable instance has a winning sequence in which all moves are \compliant with  both dominances.
    \label{thm:dominance-combined}
\end{thm}

\paragraph{Proof.} 
If a game is winnable, by \Cref{thm:dominance} it is also winnable while always building immediately after the move of an incomplete pile in the tableau. We can take this winning sequence as the starting point in the proof of \Cref{thm:safemoves}.  Each step of that proof retains winnability and moves towards a solution where the safe-building dominance is respected.  
The sequence changes in the proof of \Cref{thm:safemoves} concern some \noncompliant move $m_i$ which moves card $c_i$ when some card was safely buildable.  It is enough to check that no such change can produce a sequence which is \noncompliant with the incomplete pile dominance.   
There were two cases, depending whether $c_i$ was safely buildable or not at time $i$.

\begin{itemize}
    \item 
If $c_i$ was safely buildable, the proof of \Cref{thm:safemoves} deleted move $m_i$ and replaced it with the first move  $m_j$ building $c_i$ to foundation, which occurred in a sequence of safe builds to foundation. The only affected move that could possibly have been of a partial pile is the first move $m_i$, which has now been deleted so the new sequence remains \compliant with the incomplete pile dominance.
\item
If $c_i$ was not safely buildable at time $i$, then the proof of \Cref{thm:safemoves} simply swapped $m_i$ and $m_{i+1}$. But move $m_i$ was compliant with the incomplete pile dominance, so if $m_i$ was a partial pile move then $m_{i+1}$ was the immediate build of the card above $c_i$ to foundation. 
But this is an impossible combination because the  earlier proof showed that the move $m_i$ cannot have involved any safely buildable card and that $m_{i+1}$ was the build of a safely buildable card. 
\end{itemize}

$\blacksquare$

In summary, we have proven the correctness of two key dominances, neither of which were previously proved correct.  We have also shown that they can be used together if both are applicable.

\section{Comparative Statistics With Alternative Solvers On Three Major Games} 

\label{app:comparative}

{
\begin{table}[htb!]
\caption{Comparative results between Solvitaire and other solvers on \gamename{Canfield}, \gamename{Klondike} and \gamename{FreeCell}. } 
\begin{footnotesize}
\begin{tabular}{|l|rr|rrrrr|rrrrr|}
\multicolumn{13}{c}{Comparative results for \textbf{Canfield}.  Time limit of 30s for each solver.}   
\label{tab:comparative-freecell-unwinnable}
\label{tab:comparative-canfield} 
\label{tab:comparative-klondike}
\label{tab:comparative-freecell-winnable}
 \\
  \hline
Algorithm & Sample & $<$ limit & \multicolumn{5}{c|}{CPU time (seconds)}  & \multicolumn{5}{c|}{Kilonodes Searched}
\\ 
&&
&Mean&Median&90\% &99\% &Max 
&Mean&Median&90\% &99\% &Max \\ \hline
Wolter & 50,000 &  48,979
& 0.6677
&   $< 0.01$   &  0.8900 &  16.41 & 29.98 
& 219.3 & 0.216 & 276.5 & 5,449 & 13,650
\\
Solvitaire  & 50,000 & 47,881
  & 0.9543
&   0.020   & 1.71 & 19.99 & 29.95
& 79.06 &   0.551    & 141.5 & 1,658  & 4,318 
 \\ 
   \hline
  \multicolumn{13}{c}{~}\\
 \multicolumn{13}{c}{Comparative results for \textbf{Klondike}.  Time limit of 1hr for each solver.}\\
 \multicolumn{13}{c}{Six instances were solved incorrectly by Klondike-solver.}
\\ \hline 
Algorithm & Sample & $<$ limit & \multicolumn{5}{c|}{CPU time (seconds)}  & \multicolumn{5}{c|}{Kilonodes Searched}
\\ 
&&
&Mean&Median&90\% &99\% &Max 
&Mean&Median&90\% &99\% &Max \\ \hline
Klondike-Solver & 50,000 &  49,054
& 83.08 
&    20.50 & 140.5 & 1433 &  3578 
&
\multicolumn{5}{c|}{\textit{not recorded}}
\\
 Solvitaire  & 50,000 & 49,656
  & 32.45 
&   0.020   & 14.93 & 905.7 & 3598  &  3,897 &   3.177   &   1,685 &   113,300 & 511,000 \\ 
   \hline
\multicolumn{13}{c}{~}\\
 \multicolumn{13}{c}{Results for \textbf{FreeCell} on the first 10,000 seeds (all winnable). Time limit of 5 minutes for each solver.}\\
  \multicolumn{13}{c}{For Smart, initial run with streamliners reported one seed incorrectly but correctly found solution without streamliners.}
\\ \hline 
Algorithm & Sample & $<$ limit & \multicolumn{5}{c|}{CPU time (s)}  & \multicolumn{5}{c|}{Kilonodes Searched}
\\ 
&&
&Mean&Median&90\% &99\% &Max 
&Mean&Median&90\% &99\% &Max \\ \hline
FC-Solve & 10,000 & 9,998 & 
0.1226 &  0.0500 &   0.0600 & 0.7503 & 53.99
& 28.59 &      0.302   &   8.096 & 300.3 &  19,460
\\
 Solvitaire (None) & 10,000 & 9,746
  &6.839
   & 0.1900 &   10.32  & 160.4&  297.8 
&  3,518  & 104.6 &  5,329 &  81,110 & 163,600 \\ 
Solvitaire (Smart) & 10,000 & 10,000 
&  0.1496 &  0.0400 & 0.2800  & 1.870 & 23.24
& 75.1 & 19.93 &   144.3  & 891.0 &  10,960
\\
   \hline
\multicolumn{13}{c}{~}\\
 \multicolumn{13}{c}{Results for \textbf{FreeCell} on the first 1,000 unwinnable seeds. Time limit of 5 minutes for each solver.}\\
 \hline
Algorithm & Sample &  $<$ limit & \multicolumn{5}{c|}{CPU time (s)}  & \multicolumn{5}{c|}{Nodes Searched (thousands)}
\\
&&
&Mean&Median&90\% &99\% &Max 
&Mean&Median&90\% &99\% &Max \\ \hline
FC-Solve & 1000 & 1000 
& 0.3586 &  0.0800 &  0.2600 &  1.601 & 150.5 
&  119.2 &  15.40 &  105.5 &    688.5 &  50,030
\\
 Solvitaire  & $ 1000 $ & 998 
  & 0.8135 & 0.1200  & 1.073 & 8.830 & 205.8 & 
  431.7 &   69.33 &  546.6 &   4,760 & 101,000 \\ 
   \hline
   \end{tabular}
   \end{footnotesize}
\end{table}

 }

Our focus in this paper has been on obtaining winnability statistics using Solvitaire on the widest possible range of games.  We have therefore not focussed on performance comparison of Solvitaire with existing solvers for games where good solvers exist.  
To do such a comparison to a high scientific standard to give meaningful results would itself require a major effort, even where the alternative solver is freely available.  However, we have been able to run Solvitaire on identical instances with existing solvers for the three major games \gamename{Klondike}, \gamename{Canfield}, and \gamename{FreeCell}.  
These comparison gives a general indication of performance of the general purpose solver Solvitaire with solvers which were more specifically targeted at the relevant games.  

The three solvers were: Jan Wolter's solver for \gamename{Canfield} \cite{wolter-code}; Matt Birrell's Klondike Solver for \gamename{Klondike} \cite{shootme_klondike}; and Shlomi Fish's  Freecell Solver (version 6.10) \cite{fish_fcsolve}.
For each game, both Solvitaire and the alternative solver were run on machines with identical specification (though the machines across different games were not identical).  Timeouts varied between solvers: this was 30 seconds for \gamename{Canfield}, 5 minutes for \gamename{FreeCell}, and 1 hour for \gamename{Klondike}.  
Table~\ref{tab:comparative-canfield}  shows the results.  The sample size of each experiment is given, together with how many instances each solver could determine correctly within the timeout.  Remaining statistics only apply to those which could be determined correctly, and give statistics of time taken and nodes searched (where available).  For \gamename{Canfield} and \gamename{Klondike} we had identified bugs in the original solvers as discussed in \Cref{canfield-correction}.  For \gamename{Canfield}, we ran a minimally-corrected version of the solver, while for \gamename{Klondike} we discounted the 6 instances it reported incorrect results for.

For \gamename{Canfield}, we can see that  the corrected version of Wolter's solver was able to solve about 2.2\% more instances within 30 seconds, and also had lower run times in each statistic.  So Solvitaire is not quite as good, but the performance penalty is relatively slight.   In contrast, for \gamename{Klondike} the situation is reversed.  Here, Solvitaire is able to solve more instances and many metrics of runtime are very much better.  
Finally, for \gamename{FreeCell} we report separate experiments on the first 10,000 seeds in \Cref{tab:comparative-freecell-winnable} and (because of the rarity of unwinnable instances) on the first 1,000 unwinnable seeds in \Cref{tab:comparative-freecell-unwinnable}. For the winnable instances we report both Solvitaire used without streamliners, and with the `smart' setting.  It is notable that Solvitaire without streamliners is very much worse than FC-Solve on winnable instances, and still slightly worse on unwinnable instances.  For winnable instances, the use of the smart streamliners is extremely effective and on some measures slightly outperforms FC-Solve.  FC-Solve, however, does perform better that Solvitaire on unwinnable instances. 
 
In summary, we can see that Solvitaire is able to perform well on each of these three classic games when compared to existing solvers for those games.  For \gamename{Klondike} it is significantly better than the alternative, while it does not give as good performance as the alternatives for \gamename{Canfield} and \gamename{FreeCell}.  It is also interesting to see the dramatic improvements given by the use of streamlining in \gamename{FreeCell}.

Full results of each solver on each instance are included in our online dataset \cite{figshare-data}.

\section{Summary Statistics of Experiments Reported Here}

\label{app:summary-stats}

Statistics in this section are intended to give a general idea of the ease or difficulty that Solvitaire had with any game, as well as the total resources we devoted to that game.   However, they are not well-suited for benchmarking Solvitaire against alternative solvers, because 
the focus of our experiments was to obtain high-quality estimates of winnability percentage.
Experiments were run  on a variety of machines with different specifications, sometimes varying within a single exeriment.   

The first four columns in \Cref{tab:solvitaireresults} report on the winnability statistics from which confidence intervals were calculated. The total sample is given as well as the number proven winnable, proven unwinnable, and unknown.  In some cases, results for particular instances were not run on this particular game but deduced from related games, as described in  \Cref{sec:stronger-weaker}.

The final five columns give summary search and CPU statistics for all our data provided in our auxiliary data.  
The number of runs is the number of times Solvitaire was executed on that specific game in our experimental set, and so therefore can be either higher or lower than the sample size in the first column.  Lower numbers than the sample arises if other games are used to deduce results while higher numbers result from rerunning instances that Solvitaire initially failed to resolve.  The final three columns give, to 2 significant figures, the mean number of nodes searched, the mean CPU time per run in CPU-seconds, and the total CPU time over all runs in CPU-days.  
CPU times are as recorded by our internal timing mechanism: while we did often record a slightly more accurate external timing mechanism, which was typically $\approx 10\%$ higher, this statistic was not available for all instances.
  The most commonly used machine was provided by the Cirrus HPC system:  CPU Nodes contained 2$\times$Intel Xeon “Broadwell” 18-core cpus, 2.1 Ghz, and 256 GB RAM.   Additionally we used two local compute servers at the University of St Andrews: each of these servers  held 4$\times$AMD ``Opteron 6376'' 16-core processors for a total of 64 cores, 2.3 Ghz,
 about 512GB RAM. 
The final column for CPU Type indicates the type of machine used for runs within a game: `B' indicates the Broadwell processors, `O' the Opteron, and `X' indicates that we do not have a record for at least one run. If applicable, multiple letters can occur for a single game.

With the exception of two games, full data for all instances we experimented on is provided 
in our online dataset \cite{figshare-data}. 
The exceptions are \gamename{British Canister} and \gamename{Fortune's Favor}, which  
\label{billion-comment} were so easy and had winnability so close to 0/1 that we used samples of size $10^9$. We only retained instances which are one of: in the first $10^7$ instances; or took more than 1 sec. to solve; or 
had the rare result (winnable for \gamename{British Canister} or unwinnable for \gamename{Fortune's Favor}).   While this means complete data is not available, it  seemed a reasonable compromise between ability to check our work and excessive storage requirements.  Note that, since all instances of both games were solved, the winnability of all $10^9$ individual instances can be checked from our data. 

 \FloatBarrier
    \def \SummaryColumns {\hline Game \hfill Variant& \multicolumn{4}{c|}{Winnability Statistics}  & \multicolumn{2}{c|}{Search Statistics} & \multicolumn{3}{c|}{CPU Usage} \\  & Sample & Winnable & Unwinnable & Unknown & Runs & Mean  & Mean & Total  & Type\\   & & & & & & Nodes&(secs)&(days)&Used\\ \hline}

{ 
\begin{footnotesize}
\begin{longtable}{|l|rrrr|rr|rrl|}
\caption{Summary statistics for winnability and search for experiments reported in this paper. For search and CPU statistics, number of runs is precise with other figures given to two significant figures.  
$\dagger$ Run times for the full set of $10^9$ instances were not recorded - see main text on page~\pageref{billion-comment}.  WB : Worrying back.  SP : spaces.  BP : Build Policy, FC: number of free cells. CPU Type Used - B : Broadwell. O : Opteron.
X : Not recorded}
   \label{tab:times} \label{tab:solvitaireresults} \\
\SummaryColumns
\endfirsthead
\SummaryColumns\endhead 
Accordion & $ 10^{6}$ & 999,996 & 0 & 4 & 1,000,116 & $5.1 \times 10^{6}$ & 9.8 & 110 & OB \\ 
  Alpha Star & $ 10^{7}$ & 4,779,474 & 5,220,526 & 0 & 10,000,000 & $7.7 \times 10^{2}$ & 0.0037 & 0.42 & X \\ 
  American Canister & $ 10^{7}$ & 560,567 & 9,439,428 & 5 & 10,000,179 & $9.1 \times 10^{4}$ & 0.61 &  71 & OX \\ 
  Baker's Game & $ 10^{7}$ & 7,505,266 & 2,494,734 & 0 & 10,000,000 & $7.8 \times 10^{4}$ & 0.42 &  49 & X \\ 
  Beleaguered Castle & $2 \times 10^{6}$ & 1,362,720 & 635,919 & 1,361 & 2,671,263 & $2.6 \times 10^{6}$ & 4.9 & 150 & BX \\ 
  Black Hole & $ 10^{7}$ & 8,694,457 & 1,305,543 & 0 & 10,000,000 & $4.3 \times 10^{5}$ & 2.8 & 330 & BX \\ 
  British Canister \hfill $\dagger$& $ 10^{9}$ & 1,290 & 999,998,710 & 0 & \textit{10,001,326}  &  \textit{74} &\textit{0.000095}& n/a & X \\ 
  Canfield & $ 10^{7}$ & 7,124,239 & 2,875,241 & 520 & 10,000,000 & $1.6 \times 10^{6}$ & 4.8 & 560 & B \\ 
  Canfield \hfill (Whole pile) & $ 10^{7}$ & 6,755,771 & 3,243,482 & 747 & 10,000,000 & $2.4 \times 10^{6}$ & 6.3 & 730 & B \\ 
  Delta Star & $ 10^{7}$ & 3,441,247 & 6,558,753 & 0 & 10,000,000 & $1.0 \times 10^{3}$ & 0.0035 & 0.4 & X \\ 
  East Haven & $2 \times 10^{6}$ & 1,655,944 & 342,169 & 1,887 & 2,075,274 & $1.7 \times 10^{6}$ & 5.4 & 130 & BX \\ 
  Eight Off & $ 10^{7}$ & 9,988,054 & 11,946 & 0 & 10,000,000 & $1.4 \times 10^{4}$ & 0.046 & 5.3 & X \\ 
  Fan &$ 10^{6}$ & 487,759 & 512,241 & 0 & 1,000,000 & $6.3 \times 10^{5}$ &   1 &  12 & B \\ 
  Fore Cell & $ 10^{7}$ & 8,561,569 & 1,438,082 & 349 & 10,000,000 & $3.6 \times 10^{5}$ & 0.71 &  82 & B \\ 
  Fore Cell \hfill (BP $=$) & $ 10^{7}$ & 1,056,397 & 8,943,603 & 0 & 10,000,000 & $4.8 \times 10^{3}$ & 0.015 & 1.7 & BX \\ 
  Fortunes Favor \hfill $\dagger$ & $ 10^{9}$ & 999,999,881 & 119 & 0 &
  \textit{10,294,763}  & $\textit{2.1} \times \textit{10}^{\textit{\;4}}$ & \textit{0.068} & n/a & X \\ 
  FreeCell&  $ 10^{7}$ & 9,999,890 & 110 & 0 & 10,000,016 & $7.6 \times 10^{4}$ & 0.37 &  43 & OBX \\ 
  FreeCell\hfill (FC 0) & $ 10^{7}$ & 21,354 & 9,978,617 & 29 & 10,000,111 & $2.8 \times 10^{4}$ & 0.057 & 6.7 & BX \\ 
  FreeCell\hfill (FC 1) & $ 10^{6}$ & 193,335 & 806,370 & 295 & 1,000,749 & $1.4 \times 10^{6}$ & 3.3 &  39 & BX \\ 
  FreeCell\hfill (FC 2) & $ 10^{6}$ & 795,341 & 204,449 & 210 & 1,000,440 & $8.3 \times 10^{5}$ & 2.3 &  26 & B \\ 
  FreeCell\hfill (FC 3) & $ 10^{6}$ & 993,580 & 6,410 & 10 & 1,000,021 & $1.4 \times 10^{5}$ & 0.32 & 3.7 & B \\ 
  FreeCell\hfill (4 Piles) & $ 10^{7}$ & 864 & 9,999,136 & 0 & 10,000,000 & $1.5 \times 10^{3}$ & 0.0021 & 0.25 & X \\ 
  FreeCell\hfill (5 Piles) & $ 10^{6}$ & 38,577 & 961,392 & 31 & 1,000,173 & $5.5 \times 10^{5}$ & 1.1 &  12 & B \\ 
  FreeCell\hfill (6 Piles) & $2 \times 10^{6}$ & 1,227,828 & 770,982 & 1,190 & 2,003,743 & $5.6 \times 10^{6}$ &  13 & 290 & B \\ 
  FreeCell\hfill (7 Piles)& $ 10^{6}$ & 988,556 & 11,417 & 27 & 1,000,061 & $2.7 \times 10^{5}$ & 0.62 & 7.2 & B \\ 
  Gaps \hfill (Basic Variant) & $ 10^{7}$ & 2,490,171 & 7,509,829 & 0 & 10,000,000 & $7.2 \times 10^{5}$ & 3.4 & 400 & B \\ 
  Gaps \hfill (One Deal) & $ 10^{4}$ & 8,285 & 1,107 & 608 &  11,416 & $7.2 \times 10^{8}$ & 2,000 & 260 & B \\ 
  Golf & $ 10^{7}$ & 4,510,859 & 5,489,141 & 0 & 10,000,000 & $6.8 \times 10^{5}$ & 1.6 & 180 & B \\ 
  King Albert & $2 \times 10^{6}$ & 1,370,321 & 628,618 & 1,061 & 2,011,590 & $4.8 \times 10^{6}$ &  15 & 360 & OBX \\ 
  Klondike & $ 10^{6}$ & 819,371 & 180,472 & 157 & 1,005,717 & $2.9 \times 10^{7}$ &  75 & 870 & B \\ 
  Klondike\hfill (WB $\times$) & $ 10^{6}$ & 815,114 & 184,637 & 249 & 819,759 & $3.5 \times 10^{6}$ &  10 &  95 & B \\ 

  Klondike\hfill (SP $\checkmark$, BP $*$) & $ 10^{6}$ & 999,233 & 763 & 4 &   3,366 & $1.1 \times 10^{7}$ &  26 &   1.0 & B \\ 
    Klondike\hfill (SP $\checkmark$)& $ 10^{6}$ & 949,577 & 50,406 & 17 & 180,629 & $7.5 \times 10^{5}$ & 2.1 & 4.4 & B \\ 
    Klondike\hfill (SP $\checkmark$,BP $=$) & $ 10^{6}$ & 407,620 & 592,380 & 0 & 931,055 & $1.4 \times 10^{4}$ & 0.032 & 0.34 & B \\ 
      
  Klondike\hfill (BP $*$)& $ 10^{6}$ & 998,155 & 1,033 & 812 & 180,629 & $4.2 \times 10^{7}$ &  75 & 160 & B \\ 
 
  Klondike\hfill (BP $=$) & $ 10^{6}$ & 68,945 & 931,055 & 0 & 1,000,000 & $4.0 \times 10^{4}$ & 0.063 & 0.73 & B \\ 
  Klondike\hfill (SP $\times$, BP $*$) & $ 10^{6}$ & 24,068 & 1,134 & 974,798 &   2,645 & $2.8 \times 10^{8}$ & 600 &  18 & B \\ 
  Klondike\hfill (SP $\times$) & $ 10^{6}$ & 20,757 & 977,411 & 1,832 & 819,528 & $3.9 \times 10^{7}$ &  99 & 940 & B \\ 

  Klondike\hfill (SP $\times$,BP $=$) & $ 10^{6}$ & 1,772 & 998,228 & 0 &  68,945 & $3.3 \times 10^{4}$ & 0.051 & 0.040 & B \\ 
  Klondike\hfill (Draw 1) & $ 10^{6}$ & 904,226 & 94,629 & 1,145 & 180,837 & $6.4 \times 10^{7}$ & 190 & 400 & B \\ 
  Klondike\hfill (Draw 1,WB $\times$) & $ 10^{6}$ & 901,702 & 97,622 & 676 &  90,168 & $4.9 \times 10^{7}$ & 150 & 150 & B \\ 
  Klondike\hfill (Draw 2) & $ 10^{6}$ & 885,476 & 113,084 & 1,440 & 511,234 & $2.3 \times 10^{7}$ &  60 & 360 & B \\ 
  Klondike\hfill (Draw 2,WB $\times$) & $ 10^{6}$ & 882,409 & 116,624 & 967 & 167,750 & $4.4 \times 10^{7}$ & 110 & 220 & B \\ 
  Klondike\hfill (Draw 4) & $ 10^{6}$ & 693,296 & 306,564 & 140 & 779,013 & $2.9 \times 10^{6}$ & 8.5 &  77 & B \\ 
  Klondike\hfill (Draw 4,WB $\times$) & $ 10^{6}$ & 687,198 & 312,729 & 73 & 572,605 & $1.6 \times 10^{6}$ & 4.2 &  28 & B \\ 
  Klondike\hfill (Draw 5) & $ 10^{6}$ & 534,329 & 465,656 & 15 & 999,774 & $1.0 \times 10^{6}$ & 3.3 &  39 & B \\ 
  Klondike\hfill (Draw 5,WB $\times$) & $ 10^{6}$ & 526,376 & 473,621 & 3 & 494,742 & $3.4 \times 10^{5}$ & 0.94 & 5.4 & B \\ 
  Klondike\hfill (Draw 6) & $ 10^{6}$ & 358,539 & 641,460 & 1 & 819,759 & $3.0 \times 10^{5}$ &   1 & 9.6 & B \\ 
  Klondike\hfill (Draw 6,WB $\times$) & $ 10^{6}$ & 349,817 & 650,183 & 0 & 350,494 & $1.7 \times 10^{5}$ & 0.45 & 1.8 & B \\ 
  Klondike\hfill (Draw 7) & $ 10^{6}$ & 237,786 & 762,214 & 0 & 1,000,000 & $1.6 \times 10^{5}$ & 0.45 & 5.2 & B \\ 
  Klondike\hfill (Draw 7,WB $\times$) & $ 10^{6}$ & 229,522 & 770,478 & 0 & 237,755 & $1.2 \times 10^{5}$ & 0.31 & 0.86 & B \\ 
  Klondike\hfill (Draw 8) & $ 10^{6}$ & 122,759 & 877,241 & 0 & 1,000,000 & $6.9 \times 10^{4}$ & 0.18 & 2.1 & B \\ 
  Klondike\hfill (Draw 8,WB $\times$) & $ 10^{6}$ & 117,024 & 882,976 & 0 & 122,753 & $9.0 \times 10^{4}$ & 0.24 & 0.34 & B \\ 
  Klondike\hfill (Draw 9) & $ 10^{6}$ & 76,699 & 923,301 & 0 & 819,759 & $5.1 \times 10^{4}$ & 0.13 & 1.3 & B \\ 
  Klondike\hfill (Draw 9,WB $\times$) & $ 10^{6}$ & 72,140 & 927,860 & 0 &  76,676 & $7.6 \times 10^{4}$ & 0.2 & 0.18 & B \\ 
  Klondike\hfill ~~(Draw 10)& $ 10^{6}$ & 42,372 & 957,628 & 0 & 534,352 & $4.1 \times 10^{4}$ & 0.11 & 0.66 & B \\ 
  Klondike\hfill ~~(Draw 10,WB $\times$) & $ 10^{6}$ & 39,392 & 960,608 & 0 &  42,371 & $6.1 \times 10^{4}$ & 0.16 & 0.08 & B \\ 
  Klondike\hfill ~~(Draw 11) & $ 10^{6}$ & 20,655 & 979,345 & 0 & 905,431 & $1.3 \times 10^{4}$ & 0.034 & 0.36 & B \\ 
  Klondike\hfill ~~(Draw 11,WB $\times$) & $ 10^{6}$ & 19,037 & 980,963 & 0 &  20,654 & $4.5 \times 10^{4}$ & 0.12 & 0.029 & B \\ 
  Klondike\hfill ~~(Draw 12) & $ 10^{6}$ & 8,489 & 991,511 & 0 & 358,540 & $1.1 \times 10^{4}$ & 0.03 & 0.12 & B \\ 
  Klondike\hfill ~~(Draw 12,WB $\times$) & $ 10^{6}$ & 7,788 & 992,212 & 0 &   8,488 & $3.3 \times 10^{4}$ & 0.088 & 0.0087 & B \\ 
  Klondike\hfill ~~(Draw 13) & $ 10^{6}$ & 5,998 & 994,002 & 0 & 905,431 & $4.3 \times 10^{3}$ & 0.011 & 0.12 & B \\ 
  Klondike\hfill ~~(Draw 13,WB $\times$) & $ 10^{6}$ & 5,444 & 994,556 & 0 &   5,997 & $3.4 \times 10^{4}$ & 0.092 & 0.0064 & B \\ 
  Late-Binding Solitaire & $ 10^{7}$ & 4,702,154 & 5,297,846 & 0 & 10,000,000 & $6.7 \times 10^{4}$ & 0.054 & 6.3 & B \\ 
  Mrs Mop & $2 \times 10^{6}$ & 1,958,661 & 38,969 & 2,370 & 2,004,805 & $1.3 \times 10^{7}$ &  38 & 880 & OB \\ 
  Northwest Territory & $ 10^{6}$ & 683,669 & 316,287 & 44 & 1,001,297 & $4.9 \times 10^{6}$ &  21 & 240 & B \\ 
  Raglan & $ 10^{6}$ & 812,184 & 187,650 & 166 & 1,000,009 & $4.1 \times 10^{5}$ & 0.98 &  11 & B \\ 
  Seahaven Towers & $ 10^{7}$ & 8,933,178 & 1,066,822 & 0 & 10,000,000 & $8.4 \times 10^{4}$ & 0.12 &  14 & B \\ 
  Siegecraft & $ 10^{6}$ & 991,378 & 8,595 & 27 & 1,000,054 & $1.8 \times 10^{5}$ & 0.31 & 3.6 & B \\ 
  Simple Simon & $ 10^{6}$ & 974,476 & 25,467 & 57 & 1,000,000 & $6.0 \times 10^{4}$ & 0.15 & 1.7 & B \\ 
  Somerset & $2 \times 10^{6}$ & 1,073,962 & 924,968 & 1,070 & 2,004,561 & $5.5 \times 10^{5}$ & 2.9 &  68 & OBX \\ 
  Spanish Patience & $ 10^{7}$ & 9,986,239 & 13,746 & 15 & 10,000,028 & $2.0 \times 10^{4}$ & 0.090 &  10 & OBX \\ 
  Spider & $ 10^{4}$ & 9,731 & 0 & 269 &  11,494 & $2.6 \times 10^{8}$ & 810 & 110 & OB \\ 
  Spiderette & $ 10^{6}$ & 996,153 & 3,751 & 96 & 1,000,000 & $1.0 \times 10^{6}$ & 1.8 &  21 & B \\ 
  Streets and Alleys & $2 \times 10^{6}$ & 1,021,425 & 973,933 & 4,642 & 2,012,134 & $1.6 \times 10^{7}$ &  29 & 670 & OBX \\ 
  Stronghold & $ 10^{6}$ & 973,689 & 26,106 & 205 & 1,000,320 & $1.7 \times 10^{6}$ & 3.4 &  40 & B \\ 
  Thirty & $ 10^{7}$ & 6,745,425 & 3,254,508 & 67 & 10,000,000 & $1.4 \times 10^{5}$ & 0.22 &  26 & B \\ 
  Thirtysix & $ 10^{6}$ & 946,196 & 52,704 & 1,100 & 1,001,085 & $9.2 \times 10^{6}$ &  18 & 200 & B \\ 
  Trigon & $ 10^{7}$ & 1,599,605 & 8,400,395 & 0 & 10,000,000 & $2.7 \times 10^{3}$ & 0.015 & 1.7 & X \\ 
  Will o' the Wisp & $ 10^{7}$ & 9,992,300 & 7,487 & 213 & 10,000,906 & $2.9 \times 10^{5}$ & 1.3 & 150 & BX \\ 
  Worm Hole & $ 10^{6}$ & 998,881 & 1,104 & 15 & 1,000,662 & $2.3 \times 10^{7}$ &  41 & 470 & B \\ 
   \hline
\end{longtable}

\end{footnotesize}
}

\section{Summary of Data from the Literature}
\label{app:literature-data}
Results from previous researchers on winnability of patience games is widely spread, and presented in numerous different forms. Here we present the raw data used to generate confidence intervals for existing results throughout this paper. \Cref{tab:literature-data} shows the raw data for the best results we could find for each game, while \Cref{tab:literature-archiveref} gives  an archival URL for pages giving the reported data.   Archival URLs are particularly important: for example, many results were originally discussed in Yahoo groups, which were deleted in 2020. 
This list only includes games we have compared with Solvitaire. 
For other games not included here, useful starting points are the summaries of \citet{SolLab}, \citet{masten_winrates} and \citet{wolter_analysis}.

\begin{table}[ht]
\caption{Totals from the literature used in this paper. Numbers in italics indicate issue discussed in  accompanying note. 
For archival URLs giving source of datas in this table,   see \Cref{tab:literature-archiveref}. 
}
\label{tab:literature-data}
\begin{footnotesize}
\begin{tabular}{|l r r r r r l|}
\hline
\multicolumn{2}{|l}{Game}& \multicolumn{1}{r}{Sample} & \multicolumn{1}{r}{Winnable} & \multicolumn{1}{r}{Unwinnable} & \multicolumn{1}{r}{Unknown} &\multicolumn{1}{l|}{Notes}
\\
\hline
\tableGame{Accordion}&
$3\times 10^7$&  30,000,000 & 0 & 0 & \\
\hline
\tableGame{Baker's Game}&
$10^7$&  7,501,119 &  2,498,881& 0 & \textit{Fish reported 7,431,962 solvable} \\
&&&&&& 
\textit{using a solver configuration} \\
&&&&&&\textit{allowing false negatives \cite{pringle-pc}}\\
\hline

  \tableGame{Black Hole}  &  $1.6 \times 10^6$ & 1,391,771 & 208,229 & 0 &\\ \hline

\multicolumn{2}{|l}{Canfield \thoughtful}  & 50,000 & \textit{35,606} & \textit{13,730} & \textit{664}& \textit{See \Cref{canfield-correction}}

\\ \hline 
\multicolumn{2}{|l}{Carpet \thoughtful}  &$10^6$& {8,755,758} & {1,244,242} & {0} & \textit{Results obtained by}\\
\multicolumn{2}{|l}{--"--\hfill Pre-founded Aces}  &$10^6$& 9,518,603 & 481,397 & 0 &  \textit{Mark Masten using Solvitaire}
\\ \hline 

\tableGame{Eight Off}  & 
 $5\times 10^7$ & 49,940,034 & 59,966 & 0  &\\ \hline

\multicolumn{2}{|l}{Fore Cell}   &32,000 & 27,395 & 4,605 & 0 & \\ 

{\dittostraight} & {Same Suit}  &$10^6$ & 105,560 & 894,440 & 0 & \\
\hline

\tableGame{FreeCell}
& 8,589,934,591& 8,589,832,516 & 102,075& 0 &\\ 
\tableGame{\dittostraight\hfill 0 Cells}& 8,589,934,591 & 18,577,014 & 8,571,181,674 & 175,903 
&\\
\tableGame{\dittostraight\hfill 1 Cell~~} & 100,000 & 19,473 & 80438& 89 &
\\
\tableGame{\dittostraight\hfill 2 Cells}& 400,000 & 317,873 &      82126 &      1 &
\\
\tableGame{\dittostraight\hfill 3 Cells} &$10^6$ & 993,600  & 6,380 & 20 &
\\
\tableGame{\dittostraight\hfill 4 Piles} &32,000& 5 &31995& 0  &
\\
\tableGame{\dittostraight\hfill 5 Piles} &32,000& 1,266 &30,713& 0  &
\\
\tableGame{\dittostraight\hfill 6 Piles} &32,000& 19,685 &12,184& 131&
\\
\tableGame{\dittostraight\hfill 7 Piles} &32,000&31,641&357&2  &
\\

\hline 

 {Gaps}& Basic Variant &
10,000 &  
\textit{2,480} & \textit{7,520} & \textit{0} & \textit{Paper states sample and 24.8\% success} \\
\dittostraight &
 One Deal
 &100 & 88 & 4& 8 & \textit{not these precise numbers}\\

\hline
\tableGame{Golf}
& 100,000& 45,077 & 54,923& 0  & \\\hline
 
 \tableGame{King Albert}  & 100 & 72 & 28 & 0 & \\ 
 \hline

  \tableGame{\hfill Draw 1}  & 1,000 & \textit{919} & \textit{62} &\textit{19} & \textit{See \Cref{canfield-correction}}
\\
   \tableGame{\hfill Draw 2}  & 1,000 & \textit{801} & \textit{71} &\textit{28} & 
\\

   \tableGame{\hfill Draw 3}  & 1,000 & \textit{836} & \textit{149} &\textit{15} & 
\\
  \tableGame{Klondike \hfill Draw 4} & 1,000 & \textit{709} & \textit{285} &\textit{6}  &
\\
   \tableGame{\thoughtful \hfill Draw 5}  & 1,000 & \textit{526} & \textit{473} &\textit{1} & 
\\
   \tableGame{\hfill Draw 6}  & 1,000 & \textit{345} & \textit{655} &\textit{0} & 
\\
   \tableGame{\hfill Draw 7}  & 1,000 & \textit{233} & \textit{767} &\textit{0} & 
 \\

 \hline

 \tableGame{Late-Binding Solitaire}
 &1,000&  454 & 546 & 0 & \\ \hline
 
\tableGame{Seahaven Towers} 
& $1.5\times 10^7$ & 13,397,816 & 1,602,184& 0 & \\ \hline

\tableGame{Simple Simon} 
& 5,000 & 4,533 & \textit{0}& \textit{467}  & \textit{Solver can report false negatives} 
\\ 
&&&&&& \textit{so unsolvable listed here as unknown}
\\ \hline

\tableGame{Spider \thoughtful }  
& 32,000 &  31,998 &0&2  & 

\\ 
\hline

\tableGame{Thirty Six \thoughtful}
& 100,000 &  94,327 &  5,343 & 330&\\\hline 
 
\tableGame{Trigon}
 & $10^6$& 160,076 & 839,924 & 0 &\\\hline 
 
 \tableGame{Worm Hole} 
 & $10^6$ & 998,908 & 1,092 & 0 &\\ \hline

 \end{tabular}

\end{footnotesize}

\end{table}

\begin{table}[ht]
\caption{Archival URLs for sources of data reported in \Cref{tab:literature-data}. Note that URLs are not necessarily those of  citations in Table~\ref{tab:solpercs-comparison}. 
URLs are relative to \nolinkurl{https://web.archive.org/web/}.   For the original URL, delete  the numerical prefix and first slash.  }
\begin{footnotesize}
\begin{tabular}{|l l |}
\hline
Game \hfill (Variant) & Archival URL \\
\hline
{Accordion}& \tableCite{masten_winrates} 
\tableURLC{20220425085012/https://masten.000webhostapp.com/Accordion.html} \\
\hline
{Baker's Game}& \tableCite{pringle_bakers} \tableURLC{20220425085149/https://masten.000webhostapp.com/BakersGame.html} 
\\\hline

{Black Hole}& \tableCite{masten_winrates}  
  \tableURLC{20220425085046/http://masten.000webhostapp.com/BlackHole.html} \\ \hline

{Canfield }  & \tableCite{wolter_analysis}  
\tableURLC{20180429220704/https://politaire.com/article/canfield.html} \\ \hline 

{Carpet }   &\tableCite{masten_carpet}  
\tableURLC{20220728095714/https://masten.000webhostapp.com/Carpet.html} \\\hline 

{Eight Off}    & \tableCite{masten_winrates}
 \tableURLC{20220426164250/https://masten.000webhostapp.com/EightOff.html} \\ \hline

{Fore Cell}   & \tableCite{SolLab} 
\tableURLC{20181215222456/http://solitairelaboratory.com/fcfaq.html}
\\ 
{\dittostraight} \hfill({Same Suit})& \tableCite{masten_winrates} 
\tableURLC{20220426164250/https://masten.000webhostapp.com/EightOff.html}
\\
\hline

{FreeCell} & 
\tableURLC{20180815201227/https://fc-solve.shlomifish.org/charts/fc-pro--4fc-deals-solvability--report/}\\ 
{\dittostraight} \hfill ({0 Cells})& 
\tableCite{fish_freecell_0cell_stats}
\tableURLC{20220419155553/https://github.com/shlomif/freecell-pro-0fc-deals/blob/master/README.md}\\

{\dittostraight} \hfill (2 Cells)
& \tableCite{fish_freecell_2cell_stats}
\tableURLC{20130719010443/http://fc-solve.blogspot.com/2012/09/two-freecell-solvability-report-for.html}\\
{\dittostraight} \hfill \textit{Others}& 
\tableURLC{20221221122054/https://ipg.host.cs.st-andrews.ac.uk/KellerMillion.htm}
\\ \hline

 {Gaps}& 
 \tableURLC{20180729133856/https://link.springer.com/content/pdf/10.1007/978-0-387-35706-5\_22.pdf}

\\

\hline
{Golf}
&  \tableURLC{20170625031422/https://politaire.com/article/golf.html}\\
\hline
 {King Albert} & \tableCite{DBLP:journals/corr/Roscoe16,roscoe-pc}
 \tableURLC{20220618044831/https://arxiv.org/pdf/1611.08418.pdf} \\ 
 \hline
 
  {Klondike}  &  

  \tableURLC{20160218015922/https://github.com/ShootMe/Klondike-Solver/blob/master/Statistics.txt}
  \\ \hline

{Late-Binding Solitaire} &\tableURLC{20180409232321/http://i.stanford.edu/pub/cstr/reports/cs/tr/89/1269/CS-TR-89-1269.pdf} \\ \hline
 
{Seahaven Towers} 
&\tableCite{masten_winrates}  
\tableURLC{20220426164824/http://masten.000webhostapp.com/SeahavenTowers.html}
 \\ \hline

{Simple Simon} 
&\tableCite{fish_simplesimon}  
\tableURLC{20220428151919/https://fc-solve.shlomifish.org/mail-lists/fc-solve-discuss/archive/0974.html} 
\\ \hline

{Spider }  
&  \tableCite{plspider} 
\tableURLD{20210305230500/https://www.tranzoa.net/~alex/plspider.htm}{20210305230500/https://www.tranzoa.net/\textasciitilde alex/plspider.htm}

\\ 
\hline

{Thirty Six }
&   {\tableCite{wolter_analysis}}\tableURLC{20170624201624/http://politaire.com/article/thirtysix.html} \\\hline 
 
{Trigon}
 &   {\tableCite{wolter_analysis}}\tableURLC{20170625011319/http://politaire.com/article/trigon.html} \\\hline 
 
 {Worm Hole} 
 & \tableCite{masten_winrates} 
 \tableURLC{20220426164749/https://masten.000webhostapp.com/WormHole.html} \\ \hline
 \end{tabular}
\label{tab:literature-archiveref}
\end{footnotesize}
\end{table}



\FloatBarrier

\section{Rule Description Language}

\label{app:rule-description-language}

Our rule description language is defined as a JSON schema \cite[Draft 4]{droettboom2015understanding}.   For convenience to the user in specifying games, each parameter has a default value which is used if it is not explicitly overridden. 
The default values for every parameter are shown in Listing~\ref{rules-streets}: they define the existing game
 \gamename{Streets and Alleys}.
The JSON schema we use to parse and validate a user-defined ruleset is shown in Listing~\ref{rules-schema}. As JSON schemas are not able to express every condition defining a valid game, Solvitaire also has a secondary post-schema validation step in code, outlined in the comments below. We cannot guarantee that every expressible game under this schema is handled correctly due to the large number of rule combinations, though many variants have been tested.

\lstset{
  basicstyle=\ttfamily\footnotesize,
  columns=fullflexible,
  keepspaces=true,
}
\begin{lstlisting}[label=rules-streets,caption={Rules of \gamename{Streets and Alleys} in our JSON format.  These are  also default values used for any game where that value is unspecified. The fields `accordion' and `sequences' are used for 
    \gamename{Accordion}-like and \gamename{Gaps}-like games respectively.}]
"tableau piles": {
    "count": 8,
    "build policy": "any-suit",
    "spaces policy": "any",
    "diagonal deal": false,
    "move built group": "no",
    "move built group policy": "same-as-build",
    "face up cards": "all" },
"foundations": {
    "present": true,
    "initial cards": "none",
    "base card": "A",
    "removable": false,
    "only complete pile moves": false },
"hole": {
    "present": false,
    "base card": "AS",
    "build loops": true },
"cells": {
    "count": 0
    "pre-filled": 0 },
"stock": {
    "size": 0,
    "deal type": "waste",
    "deal count": 1, 
    "redeal": false },
"reserve": {
    "size": 0,
    "stacked": false },
"accordion": {
    "size": 0,
    "moves": [],
    "build policies": [] },
"sequences": {
    "count": 0,
    "direction": "L",
    "build policy": "same-suit",
    "fixed suit": false },
"max rank": 13,
"two decks": false
\end{lstlisting}

\lstset{
  basicstyle=\ttfamily\footnotesize,
  columns=fullflexible,
  keepspaces=true,
}
\begin{lstlisting}[label=rules-schema, caption={A JSON schema defining the rule description language for games in Solvitaire. Comments specify additional constraints not covered by the schema itself.}]

  "$schema": "http://json-schema.org/draft-04/schema#",
  "description": "JSON Schema representing a generic solitaire game",
  "type": "object",
  "properties": {
    "tableau piles": {
      "type": "object",
      "properties": {
        "count": {
          "type": "integer",
          "minimum": 0},  // must be < deck size (= 4 * ["max rank"] (* 2 if ["two decks"]))
        "build policy": {
          "type": "string",
          "enum": [
            "any-suit",
            "red-black",
            "same-suit",
            "no-build"]},
        "spaces policy": {
          "type": "string",
          "enum": [
            "any",
            "no-build",
            "kings",  // ["max rank"] must be 13
            "auto-reserve-then-any",  // ["reserve"]["size"] must be > 0
            "auto-waste-then-stock",  // ["stock"]["size"] > 0 and ["stock"]["deal type"] is "waste"
            "auto-reserve-then-waste"]},  // both of the above conditions
        "diagonal deal": {
          "type": "boolean"},
        "move built group": {
          "type": "string",
          "enum": [
            "yes",
            "no",  // ["move built group policy"] ignored
            "whole-pile",
            "maximal-group",
            "partial-if-card-above-buildable"]},
        "move built group policy": {
          "type": "string",
          "enum": [
            "same-as-build",
            "any-suit",
            "red-black",
            "same-suit",
            "no-build"]},
        "face up cards": {
          "type": "string",
          "enum": [
            "all",
            "top"]}},
      "additionalProperties": false},
    // one and only one of [foundations], [hole], [accordion][size] > 0 and [sequences][count] > 0
    // must be present
    "foundations": {
      "type": "object",
      "properties": {
        "present": {
          "type": "boolean"},
        "initial cards": {
          "type": "string",
          "enum": [
            "none",
            "one",
            "all"]},
        "base card": {
          "type": "string",
          "oneOf":[
          	{"pattern": "^(([0-9]|1[0-3]|a|A|j|J|q|Q|k|K))$"},  // must respect ["max rank"]
          	{"enum": ["random"]}]},
        "removable": {
          "type": "boolean"},
        "only complete pile moves": {
          "type": "boolean"}},
      "additionalProperties": false},
    "hole": {
      "type": "object",
      "properties": {
        "present": {
          "type": "boolean"},
        "base card": {
          "type": "string",
          "oneOf":[
          	{"pattern": "^(([0-9]|1[0-3]|a|A|j|J|q|Q|k|K)(c|C|d|D|s|S|h|H))$"},
          	{"enum": ["random"]}]},
        "build loops": {
          "type": "boolean"}}},
    "cells": {
      "type": "object",
      "properties": {
        "count": {
          "type": "integer",
          "minimum": 0},
        "pre-filled": {  // must be less than deck size
          "type": "integer",
          "minimum": 0}},
      "additionalProperties": false},
    "stock": {
      "type": "object",
      "properties": {
        "size": {  // must be less than deck size
          "type": "integer",
          "minimum": 0},
        "deal type": {
          "type": "string",
          "enum": [  // waste / tableau / hole must be specified
            "waste",
            "tableau piles",
            "hole"]},
        "deal count": {  // // must be less than ["stock"]["size"]
          "type": "integer",
          "minimum": 1},
        "redeal": {
          "type": "boolean"}},
      "additionalProperties": false},
    "reserve": {
      "type": "object",
      "properties": {
        "size": {
          "type": "integer",
          "minimum": 0},  // must be less than deck size
        "stacked": {
          "type": "boolean"}},
      "additionalProperties": false},
    "accordion": {
      "type": "object",
      "properties": {
        "size": {  // must be less than deck size
          "type": "integer",
          "minimum": 0},
        "moves": {
          "items": {
            "type": "string",
            "pattern": "^((L|R)([1-9]|[1-4][0-9]|5[0-2]))$"}},
        "build policies": {
          "type": "array",
          "items": {
            "type": "string",
            "enum": [
              "same-suit",
              "red-black",
              "any-suit",
              "same-rank"]}}},
      "additionalProperties": false},
    "sequences": {
      "type": "object",
      "properties": {
        "count": {  // must be less than deck size
          "type": "integer",
          "minimum": 0},
        "direction": {
          "type": "string",
          "enum": [
            "L",
            "R",
            "LR"]},
        "fixed suit": {
          "type": "boolean"},
        "build policy": {
          "type": "string",
          "enum": [
            "any-suit",
            "red-black",
            "same-suit"]}},
      "additionalProperties": false},
    "max rank": {
      "type": "integer",
      "minimum": 1,
      "maximum": 13},
    "two decks": {
      "type": "boolean"}},
  "additionalProperties": false
\end{lstlisting}

{
\raggedright

~\hfill\\
~\hfill\\
~\hfill\\
~\hfill\\
}

Received 8 September 2024; accepted  1 February 2026.

\end{document}